\documentclass[unnumsec,webpdf,contemporary,large]{oup-authoring-template}
\graphicspath{{Fig/}}

\theoremstyle{thmstyleone}%
\theoremstyle{thmstyletwo}%
\theoremstyle{thmstylethree}%

\usepackage{amsmath}
\usepackage{subfig}
\usepackage{graphicx}
\usepackage[colorinlistoftodos]{todonotes}
\usepackage{bm}
\usepackage{amsfonts} % for use /mathbb
\usepackage{amssymb} % math symbol package
\usepackage{stmaryrd} %\llbracket \rrbracket

\usepackage{mathtools}

\usepackage{booktabs}
\usepackage{longtable}
\usepackage{array}
\usepackage{multirow}

\usepackage[linesnumbered,ruled,vlined]{algorithm2e}
\makeatletter
\renewcommand{\@algocf@capt@plain}{above}% formerly {bottom}
\makeatother

\usepackage{pifont}
\newcommand{\cmark}{\ding{51}}%
\newcommand{\xmark}{\ding{55}}%

\usepackage{setspace} % set the vertical space of algorithm

\usepackage[version=3]{mhchem} % Typesetting chemistry  https://www.johndcook.com/blog/2010/12/08/typesetting-chemistry-in-latex/

\usepackage{todonotes}
\usepackage{comment}
 
\usepackage{color}
\definecolor{Orange}{rgb}{1,0.5,0}
\definecolor{Red}{rgb}{1,0,0}
\definecolor{Blue}{rgb}{0,0,1}

\newcommand{\BL}[1]{\textsf{\textbf{\textcolor{Red}{\footnotesize [BL: #1]}}}}

\begin{document}

\journaltitle{Briefings in Bioinformatics}
\DOI{DOI HERE}
\copyrightyear{2022}
\pubyear{2019}
\access{Advance Access Publication Date: Day Month Year}
\appnotes{Problem Solving Protocol}

\firstpage{1}

%\subtitle{Subject Section}

\title[FGS for DTI Prediction]{Fine-Grained Selective Similarity Integration for Drug-Target Interaction Prediction
}

\author[1]{Bin~Liu}
\author[1$\ast$]{Jin~Wang}
\author[1]{Kaiwei~Sun}
\author[2]{Grigorios~Tsoumakas}
\authormark{Liu et al.}

\address[1]{Key Laboratory of Data Engineering and Visual Computing, Chongqing University of Posts and Telecommunications, Chongqing 400065, China}
\address[2]{School of Informatics, Aristotle University of Thessaloniki, 54124 Thessaloniki, Greece}

\corresp[$\ast$]{Corresponding author: Jin Wang, E-mail:   \href{wangjin@cqupt.edu.cn}{wangjin@cqupt.edu.cn} } 
%\href{4476505@qq.com}{4476505@qq.com} } 

%\corresp[$\ast$]{Corresponding author: Bin Liu and Jin Wang, E-mail: \href{liubin@cqupt.edu.cn}{liubin@cqupt.edu.cn} and  \href{wangjin@cqupt.edu.cn}{wangjin@cqupt.edu.cn} } 

\received{Date}{0}{Year}
\revised{Date}{0}{Year}
\accepted{Date}{0}{Year}

%\editor{Associate Editor: Name}

%\abstract{
%\textbf{Motivation:} .\\
%\textbf{Results:} .\\
%\textbf{Availability:} .\\
%\textbf{Contact:} \href{name@email.com}{name@email.com}\\
%\textbf{Supplementary information:} Supplementary data are available at \textit{Journal Name}
%online.}

%ABSTRACT should be less than 250 words for BiB
\abstract{
The discovery of drug-target interactions (DTIs) is a pivotal process in pharmaceutical development. Computational approaches are a promising and efficient alternative to tedious and costly wet-lab experiments for predicting novel DTIs from numerous candidates. Recently, with the availability of abundant heterogeneous biological information from diverse data sources, computational methods have been able to leverage multiple drug and target similarities to boost the performance of DTI prediction. Similarity integration is an effective and flexible strategy to extract crucial information across complementary similarity views, providing a compressed input for any similarity-based DTI prediction model. However, existing similarity integration methods filter and fuse similarities from a global perspective, neglecting the utility of similarity views for each drug and target. In this study, we propose a Fine-Grained Selective similarity integration approach, called FGS, which employs a local interaction consistency-based weight matrix to capture and exploit the importance of similarities at a finer granularity in both similarity selection and combination steps. We evaluate FGS on five DTI prediction datasets under various prediction settings. Experimental results show that our method not only outperforms similarity integration competitors with comparable computational costs, but also achieves better prediction performance than state-of-the-art DTI prediction approaches by collaborating with conventional base models. Furthermore, case studies on the analysis of similarity weights and on the verification of novel predictions confirm the practical ability of FGS.
}

% to handle multiple similarities, producing a fused similarity input that preserves crucial information across the complementary views for any similarity-based DTI prediction model.
%exploitation of multiple drug and target similarities derived from heterogeneous information sources can boost the predicting accuracy. 
%Similarity integration is an effective and flexible way to process multiple similarities, producing a fused similarity matrix that captures crucial information from complementary views ... 
% With the availability of heterogeneous biological information from diverse data sources, computational methods enable to leverage multiple drug and target similarties to boost the predicting accuacy. Similarity integration is 

\keywords{drug-target interaction prediction, fine-grained, similarity integration, similarity selection}

% \boxedtext{
% \begin{itemize}
% \item Key boxed text here.
% \item Key boxed text here.
% \item Key boxed text here.
% \end{itemize}}

\maketitle

\section{Introduction}
A crucial step in drug discovery is identifying drug-target interactions (DTIs), which can be reliably conducted via laborious and expensive \textit{in vitro} experiments. To reduce the expensive workload of the wet-lab based verification procedure, computational approaches are adopted to efficiently filter potential DTIs from a large amount of candidates for subsequent biological experiments~\cite{Bagherian2021MachinePaper}. 
Computational approaches rely on machine learning techniques, such as kernel machines~\cite{Nascimento2016APrediction}, matrix factorization~\cite{Ding2022_MKTCMF}, network mining~\cite{An2021AInteractions} and deep learning~\cite{Xuan2021IntegratingPrediction}, to build a prediction model, with the goal of accurately estimating undiscovered interactions based on the chemogenomic space that incorporates both drug and target information.

% since they employ machine learning techniques, such as neighborhood classier, kernel model, matrix factorization, network mining and deep learning, to estimate reliable undiscovered interactions based on both drug (chemical) and target (genomic) spaces. 

In the past, the primary information used by computational approaches was the chemical structure of drugs and the protein sequence of targets, from which a single drug and target similarity matrix could be obtained to describe the relationship between drugs and targets in the chemogenomic space~\cite{Ezzat2018ComputationalSurvey}. 
For instance, matrix factorization methods~\cite{Ezzat2017Drug-targetFactorization,Liu2016NeighborhoodPrediction} decompose the interaction matrix into latent drug and target feature matrices that preserve the local invariance of entities in the chemogenomic space.
% learn latent drug and target feature matrices, which can not only reconstruct the interaction matrix but also preserve the local invariance of drugs and targets in the chemogenomic space.
WkNNIR~\cite{Liu2022Drug-targetRecovery}, a neighborhood method, recovers possible missing interactions of known drugs (targets) and predicts interactions for new entities using proximity information characterized by chemical structure and protein sequence-based similarities.
% utilizes the recovered interactions inferred from proximity information, i.e., interactions of drugs having similar chemical structures and targets possessing similar protein sequences, to predict new DTIs. 
%Kron-SVM~\cite{Airola2018} trains a support vector machine using the Kronecker kernel, i.e., the Kronecker product of drug and target similarity matrices, and achieves the linear computational complexity via a generalized vec trick.
% BRDTI~\cite{Peska2017} is a Bayesian ranking method that learns an interaction value based ideal target sequence for each drug and utilizes the content alignment based regularization to capture structural drug and sequential target similarities. 
Apart from computing similarities, other solutions to describe characteristics of drug structures and target sequences include utilizing handcrafted molecular fingerprints and protein descriptors~\cite{Chu2021DTI-MLCD:Method,Pliakos2021PredictingPartitioning}, as well as learning more robust high-level drug and target representations by graph, recurrent, and transformer neural networks~\cite{zhang2022DeepMGT,li2022BACPI,wu2022BridgeDPI}.
% other computational approaches utilize manually designed molecular fingerprints and protein descriptors to describe characteristics of drug structures and target sequences, respectively. Recently, deep learning models employ graph, recurrent and transformer neural networks to learn more robust high-level representations from raw chemical structures of drugs and amino acid sequences of target.

%the chemical structure of drugs and structural and physicochemical properties of the genomic sequence of targets is to derive 
%molecular descriptors (fingerprints) describing the chemical structure of drugs   
%computational methods typically utilize information of drug and target from a single source, e.g., the chemical structure of drugs and the protein sequence of target similarity, to infer candidate DTIs. A drug similarity matrix and a target similarity matrix are 

With the advancement of biological databases, there is a renewed interest in utilizing various drug and target-related information retrieved from heterogeneous data sources to improve the effectiveness of DTI prediction models. 
Multiple similarities could be derived and calculated upon diverse types of drug (target) information, assessing the relation between drugs (targets) in complementary aspects. 
There are two main schemes used by DTI prediction methods to handle multiple similarities: i) building a prediction model capable of directly processing multiple similarities, and ii) combining multiple similarities into a fused one, and then learning a prediction model upon the integrated similarities~\cite{Olayan2018DDR:Approaches,liu2021optimizing,Ding2020IdentificationFusion}. Approaches following the first strategy jointly learn a linear combination of multiple similarities and train a prediction model, which is usually a matrix factorization or kernel-based model ~\cite{Zheng2013CollaborativeInteractions,Nascimento2016APrediction,Ding2022_MKTCMF}. On the other hand, the second strategy can shrink the dimension of input space via the similarity integration procedure, improving the computational efficiency of prediction models. Furthermore, flexibility is another merit of the second strategy, since any similarity integration method and DTI prediction model can be used.

Similarity integration is a crucial step in the second strategy, which determines the importance and reliability of the input similarities for the DTI prediction model, and further influences the accuracy of the predictions. \textit{Linear} similarity integration methods accumulate multiple similarity matrices, each of which is multiplied by a weight value to indicate the importance of the corresponding similarity view to the DTI prediction task~\cite{Nascimento2016APrediction,Qiu09_KA_KernelAlignment1,Ding2020IdentificationFusion,liu2021optimizing}. 
However, the linear weights are defined according to the whole similarity matrices, failing to distinguish the utility of each similarity view for different entities (drugs or targets). For example, a drug similarity view is useful for drugs sharing the same interactivity in proximity, but is detrimental to others surrounded by neighbors interacting with different targets.
%leading to the ignorance of the entity (drug or target)-wise utility of each similarity view.
% Therefore, a single weight value fails to distinguish the importance of a similarity view to different entities. 
Similarity Network Fusion (SNF)~\cite{Wang2014SimilarityScale}, a non-linear integration method, can enhance the relationships between drugs that are proximate in multiple views via inter-view information diffusion.
% Similarity Network Fusion (SNF), a non-linear integration method, could distinguish the importance of each similarity view for different entities via inter-view information distribution~\cite{Wang2014SimilarityScale}. 
Nevertheless, it completely discards the supervised interaction information. 
In DTI prediction, SNF usually collaborates with similarity selection to remove noisy and redundant similarity views, but it processes similarity matrices in a global way, neglecting the entity-wise utility~\cite{Olayan2018DDR:Approaches,Thafar2020DTiGEMS+:Techniques}.  

% iteratively distributes information across similarity views and obtain the fused similarity matrix when the convergence is reached~\cite{Wang2014SimilarityScale}. Although the non-linear fusion manner can distinguish the importance of each similarity view for different entities, it completely discards the interaction information in the similarity combination procedure. 

In addition to calculating multiple similarities, another way to deal with multi-source biological information is to construct a heterogeneous DTI network, composed of various types of entities (nodes) and edges. 
Network-based methods learn topology-preserving representations of drugs and targets via graph embedding approaches to facilitate DTI prediction. However, they usually conduct embedding generation and interaction prediction as two independent tasks, failing to fully exploit supervised interaction information in the embedding generation step~\cite{Luo2017AInformation,An2021AInteractions}.
Deep learning models, especially graph neural networks, have also been successfully applied to infer new DTIs from a heterogeneous network due to their capacity to capture complex underlying information~\cite{Wan2019NeoDTI:Interactions,Xuan2021IntegratingPrediction,chen2022DCFME,chen2022SupDTI}. Although deep learning models have achieved improved performance, they require larger amounts of training data and are computationally intensive. % lack interpretability,

This study focuses on the similarity integration approach, which preserves crucial information from heterogeneous sources, i.e., similarity views that can identify drugs (targets) sharing the same interactivity as similar ones, and provides more compact and informative drug and target similarity matrices as input for any similarity-based DTI prediction model.
To tackle the limitations of prevalent similarity fusion methods in DTI prediction, we propose a Fine-Grained Selective similarity integration approach (FGS), which uses an entity-wise weighting strategy to distinguish the utility of similarity views for each drug and target in both similarity selection and combination procedures.
It defines a local interaction consistency-based weight matrix to indicate the importance of each type of similarity to each entity, resolves zero-weight vectors of known entities with a global vector,and infers weights for all new entities. In addition, similarity selection is conducted to filter noisy information at a finer granularity. FGS follows the linear fusion manner, achieving comparable computational complexity with other efficient linear methods.
Extensive experimental results under various settings demonstrate the effectiveness and efficiency of FGS. Furthermore, a qualitative analysis of the similarity weights demonstrates the importance of entity-wise utility in similarity integration. The practical ability of FGS is also confirmed by verifying newly discovered DTIs.

The rest of this paper is organized as follows. Section \ref{sec:Formualtion_RelatedWork} defines the problem formulation and briefly reviews existing similarity integration methods used in DTI prediction. The proposed FGS is described in Section \ref{sec:method}. Experimental evaluation results and corresponding discussions are given in Section \ref{sec:experiment}. Finally, Section \ref{sec:conclusion} concludes this work.

\section{Preliminaries}
\label{sec:Formualtion_RelatedWork}

\subsection{Problem Formulation}

% Firstly, we formally introduce the DTI prediction problem and the similarity integration.
\subsubsection{DTI Prediction}
Let $D=\{d_i\}_{i=1}^{n_d}$ be a set of drugs and $T=\{t_i\}_{i=1}^{n_t}$ be a set of targets, where $n_d$ and $n_t$ are the number of drugs and targets, respectively. 
Let $\{\mathbf{S}^{d,h}\}_{h=1}^{m_d}$ be a drug similarity matrix set with a cardinality of $m_d$, where each element $\mathbf{S}^{d,h} \in \mathbb{R}^{n_d \times n_d}$. 
Likewise, $\{\mathbf{S}^{t,h}\}_{h=1}^{m_t}$ is a set containing $m_t$ target similarity matrices, with each entry $\mathbf{S}^{t,h} \in \mathbb{R}^{n_t \times n_t}$. 
The interactions between $D$ and $T$ are represented as a binary matrix $\mathbf{Y} \in \{0,1\}^{n_d \times n_t}$, where $Y_{ij}=1$ if $d_i$ and $t_j$ have been experimentally verified to interact with each other, and $Y_{ij} = 0$ if their interaction is unknown. 
The drugs (targets) that do not have any known interaction are called \textit{new} (\textit{unknown}) drugs (targets), e.g. $d_i$ is a new drug if $\mathbf{Y}_{i\cdot}=\mathbf{0}$, where $\mathbf{Y}_{i\cdot}$ is the $i$-th row of $\mathbf{Y}$. Let $D_n=\{d_i|\mathbf{Y}_{i\cdot}=\mathbf{0}, d_i \in D \}$ and $T_n=\{t_j|\mathbf{Y}^\top_{j\cdot}=\mathbf{0}, t_i \in T\}$ be the sets of new drugs and targets, respectively. Entities with at least one known interaction are considered as known drugs or targets. 

The DTI prediction model aims to estimate the interaction scores of unknown drug-target pairs, relying on the side information of drugs and targets as well as the known interactions. Those unknown pairs with higher prediction scores are considered to have potential interactions.

\subsubsection{Similarity Integration}
The goal of similarity integration is to fuse multiple drug (target) similarity matrices into one matrix that captures crucial proximity from diverse aspects and provides more concise and informative input for DTI prediction models. Formally, similarity integration defines or learns a mapping function $f: \{\mathbb{R}^{n \times n},..., \mathbb{R}^{n \times n}\} \to \mathbb{R}^{n \times n}$ to derive the fused similarity matrix, i.e., $f\left(\{\mathbf{S}^{d,h}\}_{h=1}^{m_d}\right) \to \mathbf{S}^{d}$ and $f\left(\{\mathbf{S}^{t,h}\}_{h=1}^{m_t}\right) \to \mathbf{S}^{t}$, where $\mathbf{S}^{d} \in \mathbb{R}^{n_d \times n_d}$ and $\mathbf{S}^{t} \in \mathbb{R}^{n_t \times n_t}$ are the fused drug and target similarity matrices, respectively.

\subsection{Related Work}
\label{sec:related_work}

In this part, we review similarity integration approaches used in the DTI prediction task, including four \textit{linear} and two \textit{non-linear} methods. A brief introduction of representative DTI prediction approaches can be found in Supplementary Section A1.

% Similarity integration, also referred to as similarity fusion, is an effective strategy to handle multiple input similarities retrieved from heterogeneous data sources. It is usually adopted in the DTI prediction model as a pre-processing step, generating a combined similarity matrix that preserves crucial information across the different data aspects. Similarity integration methods are model-agnostic, i.e. they could cooperate with any prediction model that handles a single type of similarity, and enable to improve the computational effciency of the prediction model via shrinking the dimension of the input space.

\subsubsection{Linear Methods}
Linear similarity integration combines multiple drug or target similarities as follows:
\begin{equation}
    \bm{S}^{\alpha} = \sum_{h=1}^{m_\alpha} w^{\alpha}_h \bm{S}^{\alpha,h}, \  \alpha \in \{d,t\}, 
\end{equation}
where the weight $w^{\alpha}_h$ denotes the importance of $\bm{S}^{\alpha,h}$, $\sum_{h=1}^{n_\alpha}w^{\alpha}_h=1$, and the superscript $\alpha$ indicates whether the similarity integration is conducted for drugs or targets. 
An essential component of linear similarity integration methods is the definition of similarity weights.
Considering that both linear and nonlinear methods usually integrate drug similarities and target similarities in the same way, we only introduce the drug similarity integration process for simplicity. 

% \BL{Example for describe process for durg and target are in the same way: "Note that, this interaction graph encoder is a unified structure that can be used to learn both the protein and the drug representations, while it takes different input interaction graph and initial features for drugs and proteins. In the following sections, we only describe the operations for learning protein representations with input graph Gp and initial node features Xp"}

AVE~\cite{Nascimento2016APrediction} is the most intuitive linear approach that simply averages multiple similarities by assigning the same weight to each view:
\begin{equation}
    w^d_h=\frac{1}{m_d},\  h=1,\cdots,m_d. 
\end{equation} 
AVE treats each similarity view equally and does not consider interaction information in the integration process. The complexity of computing the weights of AVE is $O(m_d)$. The complexity of linear similarity combination, which is identical to other linear methods, is $O(m_dn_d^2)$.

Kernel Alignment (KA)~\cite{Qiu09_KA_KernelAlignment1}
%~\cite{Nascimento2016APrediction}
evaluates the importance of each similarity matrix, also known as kernel, according to its alignment with the ideal similarity derived from the interaction information. Let $\mathbf{Z}^d=\mathbf{Y}\mathbf{Y}^\top$ be the ideal drug similarity matrix. The alignment between $\mathbf{S}^d_h$ and $\mathbf{Z}^d$ is computed as follows:
\begin{equation}
    %A(\mathbf{S}^{d,h},\mathbf{Z}^d) =\frac{\sum_{i=1}^{n_d}\sum_{j=1}^{n_d}S^{d,h}_{ij}Z_{ij}}{n_d\sqrt{\sum_{i=1}^{n_d}\sum_{j=1}^{n_d}\left(S^{d,h}_{ij}\right)^2}}.
    A(\mathbf{S}^{d,h},\mathbf{Z}^d) =\frac{\langle{\mathbf{S}^{d,h}},{\mathbf{Z}^d}\rangle_F}{\sqrt{\langle\mathbf{S}^{d,h},\mathbf{S}^{d,h}\rangle_{F}\langle\mathbf{Z}^d,\mathbf{Z}^d \rangle_{F}}}.
\end{equation}
where $\langle \mathbf{S}^{d,h},\mathbf{Z}^d \rangle_{F} = \sum_{i=1}\sum_{j=1}{S}^{d,h}_{ij}{Z}^d_{ij}$ is the Frobenius inner product. A higher alignment value indicates that $\mathbf{S}^{d,h}$ is closer to the ideal one. The weight of each drug similarity matrix in KA is defined as:
\begin{equation}
    w^d_h=\frac{A(\mathbf{S}^{d,h},\mathbf{Z}^d)}{\sum_{i=1}^{n_d}A(\mathbf{S}^{d,i},\mathbf{Z}^d)},\  h=1,\cdots,m_d. 
\end{equation} 
The fused similarity generated by KA tends to have the utmost alignment to the ideal similarity. The complexity of the weight computation in KA is $O(m_dn_d^2)$.

Hilbert–Schmidt Independence Criterion (HSIC)~\cite{Ding2020IdentificationFusion} based multiple kernel learning obtains an optimally combined similarity that maximizes the dependence of the ideal similarity via a HSIC-based measure, which is defined as: 
%The empirically estimated $HSIC$ value of drugs and interactions is defined as follows:
\begin{equation}
    HSIC(\mathbf{S}^{d},\mathbf{Z}^d)=\frac{1}{n_d^2}tr(\mathbf{S}^{d}\mathbf{H}\mathbf{Z}^d\mathbf{H}),
    \label{eq:HSIC}
\end{equation}
where $\mathbf{H}=\mathbf{I}-\mathbf{e}\mathbf{e}^\top/n_d$, $\mathbf{e}$ is the $n_d$-dimensional vector with all elements being 1, and $\mathbf{I} \in \mathbb{R}^{n_d \times n_d}$ is an identity matrix. A larger $HSIC$ value indicates a stronger dependence between the two input similarity matrices. Based on Eq.\eqref{eq:HSIC}, HSIC obtains the optimal integrated similarity by maximizing the following objective:
\begin{equation}
    \begin{aligned}
       \max_{\mathbf{w}^d, \mathbf{S}^{d}} & \    \frac{1}{n_d^2}\text{tr}(\mathbf{S}^{d}\mathbf{H}\mathbf{Z}^d\mathbf{H}) + \lambda_1{\mathbf{w}^d}^\top\mathbf{L}\mathbf{w}^d + \lambda_2||\mathbf{w}^d||_F^2 \\
       \text{s.t.} & \ \mathbf{S}^{d} = \sum_{h=1}^{m_d} w^d_h \mathbf{S}^{d,h},  \ \sum_{h=1}^{m_d} w^d_h=1,  \\
       & \ w^d_h \geq 0, h=1,\cdots,m_d,
       \label{eq:HSIC_MKL}
    \end{aligned}
\end{equation}
where $\mathbf{L}=\mathbf{\Lambda}-\mathbf{U}$, $\mathbf{U} \in \mathbb{R}^{m_d \times m_d}$ stores the alignment between each pair of drug similarity matrices, e.g., $U_{ij}=A({\mathbf{S}^{d,i}},{\mathbf{S}^{d,j}})$, and $\mathbf{\Lambda} = \text{diag}(\mathbf{U}\mathbf{e})$ is a diagonal matrix with row sum of $\bm{U}$. Eq.\eqref{eq:HSIC_MKL} is a quadratic optimization problem, where the first term rewards the dependence between $\mathbf{S}^d$ and $\mathbf{Z}^d$, and the last two terms favor smooth similarity weights. 
The complexity of solving the problem in Eq.\eqref{eq:HSIC_MKL} is $O(m_dn_d^3+rm_d^3)$ with $r$ being the number of iterations.
%\frac{\text{tr}({\mathbf{S}^{d,i}}^\top{\mathbf{S}^{d,j}})}{||\mathbf{S}^{d,i}||_{F}||\mathbf{S}^{d,j}||_{F}}$, 

Motivated by the \textit{guilt-by-association} principle, Local Interaction Consistency (LIC)~\cite{liu2021optimizing} emphasizes the similarity view in which more proximate drugs or targets have the same interactions. Let $\mathbf{C}^{d,h} \in \mathbb{R}^{n_d \times n_t}$ be the local interaction consistency matrix of $\mathbf{S}^{d,h}$. $C^{d,h}_{ij}$ denotes the local interaction consistency of $d_i$ for $t_j$ under the similarity view $\mathbf{S}^{d,h}$:
%the proportion of neighbors of $d_i$ having the same interactivity w.r.t. $Y_{ij}$:
\begin{equation}
    C^{d,h}_{ij} = \frac{1}{\sum_{d_l \in \mathcal{N}^{k,h}_{d_i}}S^{d,h}_{il}} \sum_{d_l \in {\mathcal{N}^{k,h}_{d_i}}}  S^{d,h}_{il} \llbracket Y_{il} = Y_{ij} \rrbracket,
\label{eq:Cdh_ij}
\end{equation}
where $\mathcal{N}^{k,h}_{d_i}$ denotes the $k$-nearest neighbors ($k$NNs) of $d_i$ retrieved according to $\bm{S}^{d,h}_{i\cdot}$ and $\llbracket \cdot \rrbracket$ is the indicator function. 
%that returns 1 if the input event is true and 0 otherwise. 
The local interaction consistency of a whole similarity matrix is obtained by averaging the $C^{d,h}_{ij}$ of all interacting pairs:
\begin{equation}
    c^d_h = \frac{1}{|P_1|} \sum_{(i,j) \in P_1} C^{d,h}_{ij}, \  h=1,\cdots,m_d,
\label{eq:LCd}
\end{equation} 
where $P_1 =\{(i,j)|Y_{ij}=1,\ i=1,\cdots,n_d,\ j=1,\cdots,n_t\}$. A higher $c^d_h$ indicates that similar drugs tend to have the same interactivity. In LIC, similarity weights are calculated by normalizing $\mathbf{c}^d$:
\begin{equation}
    w^d_h = \frac{c^d_h}{\sum_{i=1}^{m_d} c^d_i}.
\end{equation}
The computational complexity of computing LIC-based weights is $O(m_d(n_d^2+n_d k\log(k)+|P_1|))$.

\subsubsection{Nonlinear Methods}
Similarity Network Fusion (SNF)~\cite{Wang2014SimilarityScale} is a \textit{nonlinear} method that iteratively delivers information across diverse similarity views to converge to a single fused similarity matrix. Given a similarity matrix $\mathbf{S}^{d,h}$, its normalized matrix $\mathbf{P}^{d,h} \in \mathbb{R}^{n_d \times n_d}$ that avoids numerical instability and its $k$NN sparsified matrix $\mathbf{Q}^{d,h} \in \mathbb{R}^{n_d \times n_d}$ that preserves the local affinity are defined as follows:
\begin{equation}
P^{d,h}_{ij} = \left\{
\begin{aligned}
   & \frac{S^{d,h}_{ij}}{2\sum_{l=1,l\neq i}^{n_d}S^{d,h}_{lj}}, \text{ if } i\neq j \\
   & 0.5, \text{ otherwise}
\end{aligned}
\right. \  i,j=1,\cdots,n_d.
\end{equation}
\begin{equation}
Q^{d,h}_{ij} = \left\{
\begin{aligned}
   & \frac{S^{d,h}_{ij}}{\sum_{d_l \in \mathcal{N}^{k,h}_{d_i}}S^{d,h}_{il}}, \text{ if } d_j \in \mathcal{N}^{k,h}_{d_i} \\
   & 0, \text{ otherwise}
\end{aligned}
\right. \  i,j=1,\cdots,n_d.
\end{equation}
To merge the information of all similarities, SNF iteratively updates each similarity matrix using the following rule:
\begin{equation}
    \mathbf{P}^{d,h} = \mathbf{Q}^{d,h} \left( \frac{\sum_{i=1, i\neq h}^{m_d}  \mathbf{P}^{d,i}}{m_d-1} \right) {\mathbf{Q}^{d,h}}^\top, \  h=1,\cdots,m_d.   
\end{equation}
Each update of $\mathbf{P}^{d,h}$ produces $m_d$ parallel interchanging diffusion processes on $m_d$ similarity matrices. Once the fusion process converges or the terminal condition is reached, the final integrated similarity matrix is calculated as follows:
\begin{equation}
    \mathbf{S}^{d} = \frac{1}{m_d}\sum_{h=1}^{m_d} \mathbf{P}^{d,h}.
\end{equation}
If the two drugs are similar in all views, their similarity will be augmented in the final matrix and vice versa. 
%The advantage of SNF over other linear methods is the ability to distinguish the importance of similarity values for different drugs.
The complexity of SNF, which mainly depends on the iterative update, is $O(rm_dn_d^3)$, where $r$ is the number of iterations.  % $n^2+nk\log(k)$

In DTI prediction, SNF usually collaborates with similarity selection strategies to integrate only informative similarities. SNF with Heuristic similarity selection (SNF-H)~\cite{Olayan2018DDR:Approaches} applies the fusion process to a set of useful similarities, which are obtained by filtering less informative similarities with high entropy values and deleting redundant similarities based on the Euclidean distance between each pair of matrices. The major weakness of SNF-H is its ignorance of interaction information. As similarity selection is much faster than similarity fusion, the complexity of SNF-H is $O(rm'_dn_d^3)$, where $m'_d$ is the number of selected similarities.
 
SNF with Forward similarity selection (SNF-F)~\cite{Thafar2020DTiGEMS+:Techniques} determines the optimal similarity subset via sequentially adding similarity matrices in a greedy fashion until no predicting performance improvement is observed, and then fuses the selected similarities in a nonlinear manner. Although SNF-F implicitly utilizes interaction information by choosing similarities based on the DTI prediction results, this greedy selection fashion is more time-consuming, because it requires training one model for each similarity combination candidate. The complexity of SNF-F is $O(m_d^2\Theta(n_d,n_t)+r'm_dn_d^3)$, where $\Theta(n_d,n_t)$ is the complexity of training a DTI prediction model. % that provides the prediction performance.

\section{Methods}
\label{sec:method}
%fine-grained       finer granularity
% In this section, we first describe the proposed FGS, which  distinguishes the utility of similarities and linearly aggregates selected informative similarities from multiple views at a finer granularity. Then, the computational complexity of FGS is analyzed. Last, a summary comparison between FGS and other similarity integration method is given. 

\subsection{Fine-Grained Selective Similarity Integration}
\begin{figure*}[h]
\centering
\includegraphics[width=0.95\textwidth]{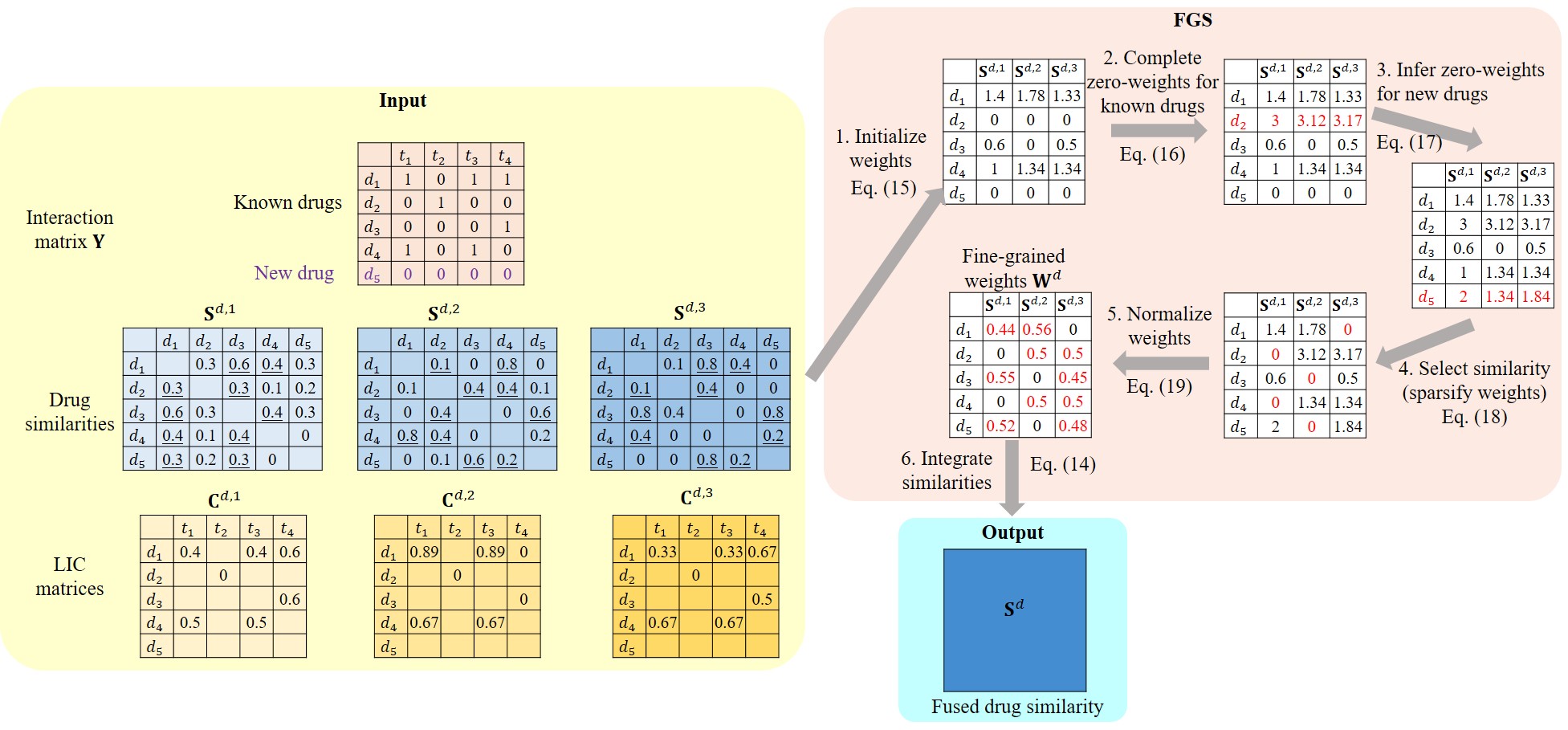}
\caption{The workflow of FGS. The input to FGS includes the interaction matrix ($\mathbf{Y}$), drug similarities ($\{\mathbf{S}^{d,h}\}_{h=1}^{3}$), and LIC matrices ($\{\mathbf{C}^{d,h}\}_{h=1}^{3}$) computed according to Eq.\eqref{eq:Cdh_ij}. Diagonals of similarity matrices and LIC matrix values corresponding to non-interacting pairs ($Y_{ij}=0$), which are not used in FGS, are not shown in the figure for simplicity. We use $k=2$ and $\rho=1/3$ in this example, and similarity values of each drug's 2-nearest neighbors (2NNs) are underlined. 
The whole procedure of FGS consists of six steps. 
1. Initialize weight matrix using $\mathbf{Y}$ and $\{\mathbf{C}^{d,h}\}_{h=1}^{3}$. 
2. Complete zero-weights for known drugs ($d_2$). 3. Infer weights for new drugs ($d_5$), where $W^d_{51}$ is the sum of $W^d_{11}$ and $W^d_{13}$ because $d_1$ and $d_3$ are the 2NNs of $d_5$ in the 1st similarity view, and $W^d_{52}$ ($W^d_{53}$) is the sum of $d_3$ and $d_4$'s weight for $\mathbf{S}^{d,2}$ ($\mathbf{S}^{d,3}$), i.e., $W^d_{32}$ and $W^d_{42}$ ($W^d_{33}$ and $W^d_{43}$).   
4. Select similarities by setting the smallest value in each row of the weight matrix to zero.
5. Normalize the weight matrix row-wise.
6. Integrate multiple similarities $\{\mathbf{S}^{d,h}\}_{h=1}^{3}$ using fine-grained weight ($\mathbf{W}^{d}$) to obtain the fused similarity $\mathbf{S}^{d}$.
}
% \textbf{Weight initialization} where $\mathbf{Y}$ and $\{\mathbf{C}^{d,h}\}_{h=1}^{3}$ are used. 2. \textbf{Zero-weight completion for known drugs}, 3.inferring weights for new drugs, 4.similarity selection, 5. row-wise normalization; 6. similarity integration.} 
\label{fig:FGS}
\end{figure*}

FGS distinguishes the utility of similarities per each drug (target) and linearly aggregates selected informative similarities from multiple views at a finer granularity.
To distinguish the different distribution and property of each drug, we define a fine-grained weight matrix $\mathbf{W}^{d} \in \mathbb{R}^{n_d \times m_d}$, with each element $W^{d}_{ih}$ denoting the reliability of similarities regarding $d_i$ in the $h$-th view, i.e., $\mathbf{S}^{d,h}_{i\cdot}$. Based on $\mathbf{W}^{d}$, we obtain the integrated drug similarity matrix $\mathbf{S}^{d}$, with each row linearly combining the corresponding weighted similarity vectors:
\begin{equation}
    \mathbf{S}^{d}_{i\cdot} = \sum_{h=1}^{m_d} W^{d}_{ih}\mathbf{S}^{d,h}_{i\cdot}, \  i=1,\cdots,n_d.   \label{eq:FGS_integration}
\end{equation}
%In \eqref{eq:FGS_integration}, the importance of similarities view to various drugs could be different.
Figure \ref{fig:FGS} schematically depicts the workflow of FGS. The procedure of target similarity fusion in FGS is the same as that of drugs. Therefore, we will mainly describe FGS for drug similarities. 
% The remainder of this section will first describe the computation of fine-grained drug similarity weights in detail, then illustrate how the proposed FGS works for arriving new drugs in the inductive learning setting, and analyze the computational complexity of FGS at last.

% Sparsifying weights to filter noisy similarities

% \subsection{Fine-Grained Weight}
\textbf{Initialization.}
In FGS, the local interaction consistency is employed to initialize the fine-grained weights due to its effectiveness in reflecting interaction distributions of proximate drugs. Specifically, given $\{\mathbf{C}^{d,h}\}_{h=1}^{m_d}$ defined in Eq.\eqref{eq:Cdh_ij}, $\mathbf{W}^{d}$ is initialized by aggregating all local interaction consistencies of interacting pairs for each drug similarity view:
\begin{equation}
    W^{d}_{ih} = \sum_{j=1}^{n_t}C^{d,h}_{ij}Y_{ij}, \  i=1,\cdots,n_d, \  h=1,\cdots,m_d.
\label{eq:Wd_ih}
\end{equation}

\textbf{Zero-Weights Completion for Known Drugs.}
There may be some known drugs whose similarity weights are all ``0"s, i.e., $\mathbf{W}^{d}_{i\cdot}=\mathbf{0}$. The zero-weight vector indicates that all types of similarities regarding the corresponding drug are totally useless, which would lead to an invalid integration. To avoid this issue, we define a global weight vector $\mathbf{v}^{d} = \sum_{i=1}^{n_d}\mathbf{W}^d_{i\cdot}$ which reflects the utility of each type of similarity over all drugs, and assign $\mathbf{v}^d$ to all drugs associated with zero-valued weight vectors:
\begin{equation}
    \mathbf{W}^{d}_{i\cdot} = \mathbf{v}^{d}, \text{ if }  \mathbf{W}^{d}_{i\cdot}=\mathbf{0}, \ i=1,\cdots,n_d.
\label{eq:Wd_zeroWight}
\end{equation}

\textbf{Inferring Weights for New Drugs.}
Since new drugs lack any known interacting target, their initialized weights, calculated according to Eq.\eqref{eq:Wd_ih}, are all zeros as well. 
However, the global weight vector based imputation is not suitable for new drugs. As different new drugs may have different potential interactions, assigning a unique weight vector to every new drug fails to reveal their distinctive properties.
In FGS, we extend the \textit{guilt-by-association} principle, assuming that similarity weights of a drug are also approximate to its neighbors. Specifically, given a new drug $d_x$, we infer its weights by aggregating the weights of its $k$ nearest known drugs in each similarity view:
\begin{equation}
    W^{d}_{xh} = \sum_{d_i \in {\mathcal{N}^{k,h}_{d_x}}} W^d_{ih}, \  d_x \in D_n, \ h=i,\cdots,m_d.
\label{eq:Wd_x} 
\end{equation}

\textbf{Fine-Grained Similarity Selection.}
A lower local interaction consistency indicates that similar drugs usually interact with different targets, which further implies that noisy information is possessed by the corresponding similarity view. 
%Involving noisy similarities in the integration process would impact the quality of the fused similarity matrix. 
To mitigate the influence of noise, we conduct a fine-grained selective procedure to filter corrupted similarities for each drug by sparsifying the weight matrix.
Specifically, for each row of $\mathbf{W}^{d}$, its $\rho m_d$ smallest weights, whose corresponding similarities are considered as noisy, are set to 0:
\begin{equation}
    \mathbf{W}^{d}_{ih} = 0, \  h \in \mathcal{H}_i, \  i=1,\cdots,n_d,  
\label{eq:Wd_filter}
\end{equation}
where $\mathcal{H}_i$ is a set containing indices of the $\rho m_d$ smallest values in $\mathbf{W}^{d}_{i\cdot}$, and $\rho$ is the filter ratio. Similarities with zero weights are discarded, while those with non-zero weights are eventually used to generate the integrated similarity matrix.
Different from existing similarity selective strategies that delete whole similarity matrices, the selection process in FGS evaluates similarities at a finer granularity. For each similarity view, FGS gets rid of noisy similarities for some drugs, while retaining informative similarities for others. As shown in the step 4 of Figure \ref{fig:FGS}, the first similarity view is filtered for drugs $d_1$ and $d_5$, but retained for the rest three drugs.

\textbf{Normalization.} The row-wise normalization is applied to the fine-grained similarity weight matrix, so as to keep the range of similarity values unchanged:
\begin{equation}
    \mathbf{W}^{d}_{ih} = \frac{\mathbf{W}^{d}_{ih}}{\sum_{h'=1}^{m_d} \mathbf{W}^{d}_{ih'}}, \  i=1,\cdots, n_d.
\label{eq:Wd_norm}
\end{equation}

\begin{spacing}{0.5}
\begin{algorithm}[h]
\small
\caption{Integrating multiple drug similarities by FGS}
\label{alg:FGS}
\SetKwData{Left}{left}\SetKwData{This}{this}\SetKwData{Up}{up}
\SetKwInOut{Input}{input} \SetKwInOut{Output}{output}
\Input{$\bm{Y}$, $D$, $\{\mathbf{S}^{d,h}\}_{h=1}^{m_d}$, $\rho$, $k$}
\Output{$\mathbf{S}^d$} 
$\mathbf{S}^d \leftarrow \mathbf{0}$ \;
Compute $\mathbf{C}^d$ via Eq.\eqref{eq:Cdh_ij}\;
Find new drug set $D_n=\{d_i|\mathbf{Y}_{i\cdot}=\mathbf{0}, d_i \in D \}$ \;
\For(\tcc*[h]{Initilization}){$i\leftarrow 1$ \KwTo $n_d$}{
    \For {$h\leftarrow 1$ \KwTo $m_d$}{
       $W^{d}_{ih} = \sum_{j=1}^{n_t}C^{d,h}_{ij}Y_{ij}$ \tcc*[r]{Eq.\eqref{eq:Wd_ih}}
    }
}
Compute global weight vector $\mathbf{v}^{d} = \sum_{i=1}^{n_d}\mathbf{W}^d_{i\cdot}$ \;
\ForEach(\tcc*[h]{Compelete zero-weights for konwn drugs}){$d_i \in D/D_n$}{
    \If{$\mathbf{W}^{d}_{i\cdot} =\mathbf{0}$} 
    {$\mathbf{W}^{d}_{i\cdot} \leftarrow \mathbf{v}^{d}$ \tcc*[r]{Eq. \eqref{eq:Wd_zeroWight})} } 
}
\ForEach(\tcc*[h]{Infer weights for new drugs}){$d_x \in D_n$}{
     \For {$h\leftarrow 1$ \KwTo $m_d$}{
     $W^{d}_{xh} \leftarrow \sum_{d_i \in {\mathcal{N}^{k,h}_{d_x}}} W^d_{ih}$ \tcc*[r]{Eq.\eqref{eq:Wd_x}}
    }    
}
\For(\tcc*[h]{Select similarity}){$i\leftarrow 1$ \KwTo $n_d$}{
    $\mathcal{H}_i \leftarrow$ indices of the $\rho m_d$ smallest values in $\mathbf{W}^{d}_{i\cdot}$ \;
    \ForEach{$h \in \mathcal{H}_i$}{
        $\mathbf{W}^{d}_{ih} \leftarrow 0$ \tcc*[r]{Eq.\eqref{eq:Wd_filter})}
    } 
}
\For(\tcc*[h]{Normalize weights}){$i\leftarrow 1$ \KwTo $n_d$}{
$s \leftarrow \sum_{h'=1}^{m_d} \mathbf{W}^{d}_{ih'}$ \;
    \For {$h\leftarrow 1$ \KwTo $m_d$}{
         $\mathbf{W}^{d}_{ih} = \mathbf{W}^{d}_{ih}/s$ \tcc*[r]{Eq.\eqref{eq:Wd_norm})}
    }
}
\For(\tcc*[h]{Integrate similarity}){$i\leftarrow 1$ \KwTo $n_d$}{
    $\mathbf{S}^{d}_{i\cdot} \leftarrow \sum_{h=1}^{m_d}W^{d}_{ih}\mathbf{S}^{d,h}_{i\cdot}$ \tcc*[r]{Eq.\eqref{eq:FGS_integration})}
}
\end{algorithm}
\end{spacing}

The process of fusing multiple drug similarities by FGS is summarized in Algorithm \ref{alg:FGS}.
The procedure of integrating target similarities with fine-grained weights is achieved in the same way, except for replacing $Y_{ij}$ with $Y_{ij}^\top$ in Eq.\eqref{eq:Wd_ih} and substituting superscript $d$ with $t$ in Eq.\eqref{eq:FGS_integration}-\eqref{eq:Wd_norm}.
% weights, denoted as $\mathbf{W}^t \in \mathbb{R}^{n_t \times m_t}$,

\subsection{Computational Complexity Analysis}
In FGS, the computational complexity of fine-grained weight initialization, including computing $\{\mathbf{C}^{d,h}\}_{h=1}^{m_d}$, is equal to $O(m_d(n_d^2+n_dk\log(k)))$. The computational costs of zero-weights imputation for known and new drugs are $O(m_dn_d)$ and $O(m_d|D_n|k\log(k))$, respectively, where $|D_n|$ is the number of new drugs. The computational complexity of similarity selection and weight normalization is $O(m_dn_d)$. Considering $|D_n|<n_d$, the computational complexity of the fine-grained weight calculation is $O(m_d(n_d^2+n_dk\log(k)))$. 
For weighted similarity integration, its computational complexity is $O(m_dn_d^2)$, which is the same as other conventional linear combination approaches using a singular weight value for each similarity matrix. 
The overall complexity of FGS is therefore $O(m_d(n_d^2+n_dk\log(k)))$.

\subsection{Overview of Similarity Integration Methods}

Table \ref{tab:SI_methods} summarizes the characteristics and computational complexity of the proposed FGS and other similarity integration methods introduced in Section \ref{sec:related_work}. 

First, compared with other linear similarity integration approaches, FGS incorporates similarity selection to filter noisy information and is aware of the distinctive properties of drugs (targets) in each similarity view. 
In addition, the computational complexity of similarity integration in FGS is the same as  other linear methods, which is linear to the number of similarity views and quadratic to the number of drugs (targets).

The advantages of FGS over nonlinear methods are twofold. 
% First, FGS enables to efficiently leverage supervised information provided by interactions, while SNF approaches either ignore the interaction information or employ it in an implicit way. 
First, FGS selects similarity at a fine granular perspective, while the two non-linear methods discard whole similarities without considering the entity-wise utility of similarity views.
Second, FGS is more computationally efficient than the two SNF-based approaches, whose computational cost is cubic to the number of drugs (targets).
% Third, FSG is suitable for the inductive learning setting, i.e. to predict the interaction of any unseen drug-target pairs, while nonlinear methods can not.
 % versatile

% {\color{green}\xmark} {\color{red}\cmark}
\begin{table*}[t]
\centering
\small
\setlength{\tabcolsep}{2pt}
\caption{Summarization of Similarity Integration Approaches}
\label{tab:SI_methods}
\begin{tabular}{@{}ccccccc@{}}
\toprule
\multirow{2}{*}{Method} & \multirow{2}{*}{Type} & \multirow{2}{*}{Supervised} & \multicolumn{2}{c}{Capturing entity-wise utility} & \multicolumn{2}{c}{Computational complexity} \\ \cmidrule(l){4-7} 
 &  &  & Similarity selection & Similarity integration & Weight calculation & Similarity integration \\ \midrule
AVE~\cite{Nascimento2016APrediction} & Linear & {\color{red}\xmark} & - & {\color{red}\xmark} & $O(m_d)$. & $O(m_dn_d^2)$ \\
KA~\cite{Qiu09_KA_KernelAlignment1} & Linear & {\color{green}\cmark} & - & {\color{red}\xmark} & $O(m_dn_d^2)$ & $O(m_dn_d^2)$ \\
HSIC~\cite{Ding2020IdentificationFusion} & Linear & {\color{green}\cmark} & - & {\color{red}\xmark} & $O(m_dn_d^3+rm_d^3)$ & $O(m_dn_d^2)$ \\
LIC~\cite{liu2021optimizing} & Linear & {\color{green}\cmark} & - & {\color{red}\xmark} & $O(m_d(n_d^2+n_d k\log(k)+|P_1|))$ & $O(m_dn_d^2)$ \\
SNF-H~\cite{Olayan2018DDR:Approaches} & Non-linear & {\color{red}\xmark} & {\color{red}\xmark} & {\color{green}\cmark} & - & $O(rm'_dn_d^3)$ \\
SNF-F~\cite{Thafar2020DTiGEMS+:Techniques} & Non-linear & {\color{green}\cmark} & {\color{red}\xmark} & {\color{green}\cmark} & - & $O(m_d^2\Theta(n_d,n_t)+rm'_dn_d^3)$ \\ 
\textbf{FGS} (ours) & Linear & {\color{green}\cmark} & {\color{green}\cmark} & {\color{green}\cmark} & $O(m_d(n_d^2+n_dk\log(k)))$ & $O(m_dn_d^2)$ \\
\bottomrule
\end{tabular}
\end{table*}

\section{Experiments}
\label{sec:experiment}

\subsection{Dataset}
In the experiments, we utilize five benchmark DTI datasets, including four collected by Yamanishi~\cite{Yamanishi2008PredictionSpaces}, namely Nuclear Receptors (NR), G-protein coupled receptors (GPCR), Ion Channel (IC), and Enzyme (E), and one obtained from~\cite{Luo2017AInformation} (denoted as Luo). Table \ref{tab:dataset} summarizes the information of these datasets. More details on similarity calculations can be found in Supplementary Section A2.

\textbf{Yamanishi datasets} only contain interactions discovered before they were constructed (in 2007). Therefore, we update them by adding newly discovered DTIs recorded in the up-to-date version of KEGG, DrugBank, and ChEMBL databases~\cite{Liu2022_MDMF}. There are 175, 1350, 3201, and 4640 interactions in the updated NR, GPCR, IC, and E datasets. 
Nine drug and nine target similarities from~\cite{Nascimento2016APrediction} are used to describe the drug and target relations in various aspects. Six drug similarities are constructed based on chemical structure (CS) with various metrics, while the other three are derived from drug side effects (SE). Seven target similarities are calculated based on protein amino-acid sequence (AAS) using different similarity assessments, while the other two are obtained from Gene Ontology (GO) terms and protein-protein interactions (PPI), respectively.

\textbf{Luo dataset} includes 1923 DTIs obtained from DrugBank 3.0. The similarities of the Luo dataset are obtained from~\cite{Luo2017AInformation}. Specifically, four types of drug similarity are computed based on CS, drug-drug interactions (DDI), SE, and drug-disease associations (DIA). Three kinds of target similarities are derived from AAS, PPI, and target-disease associations (TIA).

\begin{table*}[th]
\centering
\small
\caption{Characteristic of DTI datasets, where different colors denotes similarities derived from data diverse sources}
\label{tab:dataset}
\begin{tabular}{@{}c>{\centering\arraybackslash}p{2cm}>{\centering\arraybackslash}p{2cm}>{\centering\arraybackslash}p{2cm}>{\centering\arraybackslash}p{2cm}|c@{}}
\toprule
Dataset & NR & GPCR & IC & E & Luo \\ \hline
Drug & 54 & 223 & 210 & 445 & 708 \\
Target & 26 & 95 & 204 & 664 & 1512 \\
Interaction & 166 & 1096 & 2331 & 4256 & 1923 \\
Sparsity & 0.118 & 0.052 & 0.054 & 0.014 & 0.002 \\ \hline
\begin{tabular}[c]{@{}c@{}}Drug Similarity \\ (Source) $^\ddag$ \end{tabular} & \multicolumn{4}{c|}{\begin{tabular}[c]{@{}c@{}}{\color{red}SIMCOMP, LAMBDA, MARG, MINMAX, SPEC, TAN (CS)}; \\ {\color{blue}AERS-b, AERS-f, SIDER (SE)}\end{tabular}} & \begin{tabular}[c]{@{}c@{}}{\color{red}TAN (CS)}, {\color{green}DDI}, \\ {\color{olive} DIA}, {\color{blue}SE}\end{tabular} \\ \hline
\begin{tabular}[c]{@{}c@{}}Target Similarity \\ (Source)$^\S$\end{tabular} & \multicolumn{4}{c|}{\begin{tabular}[c]{@{}c@{}}{\color{red} SW, MIS-k3m1, MIS-k4m1, MIS-k3m2, MIS-k4m2, SPEC-k3, SPEC-k4 (AAS)};\\  {\color{green} PPI}; {\color{violet} GO}\end{tabular}} & \begin{tabular}[c]{@{}c@{}}{\color{red}SW (AAS)}, {\color{green} PPI}, \\ {\color{olive} TIA} \end{tabular} \\ \bottomrule
\multicolumn{6}{l}{
\setlength\extrarowheight{-2pt}
\begin{tabular}[l]{@{}l@{}}
{\tiny $\ddag$ {\color{red}CS: chemical structure}, {\color{blue}SE: drug side effect}, {\color{green}DDI: drug-drug interaction}, {\color{olive} DIA: drug-disease association}} 
\\
{\tiny $\S$ {\color{red}AAS: amino-acid sequence}, {\color{green}PPI: protein-protein interaction}, {\color{violet} GO: Gene Ontology annotation}, {\color{olive} TIA: target-disease association}} 
\end{tabular} } 
\end{tabular}
\end{table*}

\subsection{Experiment Setup} 

In the empirical study, we first compare the proposed FGS with six similarity integration methods, namely AVE~\cite{Nascimento2016APrediction}, KA~\cite{Qiu09_KA_KernelAlignment1}, %~\cite{Nascimento2016APrediction},
HSIC~\cite{Ding2020IdentificationFusion}, LIC~\cite{liu2021optimizing}, SNF-H~\cite{Olayan2018DDR:Approaches}, and SNF-F~\cite{Thafar2020DTiGEMS+:Techniques}. The similarity integration method needs to collaborate with a DTI prediction model to infer unseen interactions. Therefore, we employ five DTI prediction approaches, including WKNNIR~\cite{Liu2022Drug-targetRecovery}, NRLMF~\cite{Liu2016NeighborhoodPrediction}, GRMF~\cite{Ezzat2017Drug-targetFactorization}, BRDTI~\cite{Peska2017}, and DLapRLS~\cite{Ding2020IdentificationFusion}, as base models of similarity integration methods.  
Each base model is trained on the fused drug and target similarities generated by each similarity integration method. Within each base model, better prediction performance indicates the superiority of the corresponding similarity integration approach.

In addition, we further compare FGS, which collaborates with well-performing base models, with state-of-the-art DTI prediction models that can directly handle heterogeneous information. Specifically, the competitors include four traditional machine learning-based approaches (MSCMF~\cite{Zheng2013CollaborativeInteractions},
KRLSMKL~\cite{Nascimento2016APrediction}, MKTCMF~\cite{Ding2022_MKTCMF}, and NEDTP~\cite{An2021AInteractions}) and four deep learning models (NeoDTI~\cite{Wan2019NeoDTI:Interactions}, DTIP~\cite{Xuan2021IntegratingPrediction}, DCFME~\cite{chen2022DCFME}, and SupDTI~\cite{chen2022SupDTI}).

To verify the effectiveness of our method, three more challenging prediction settings that estimate interactions for new drugs and/or new targets are considered. Three types of cross validation (CV) are utilized accordingly~\cite{Pliakos2020Drug-targetReconstruction}:
\begin{itemize}
    \item CVS$_d$: 10-fold CV on drugs to predict DTIs between new drugs and known targets, where one drug fold is split as a test set.  
    \item CVS$_t$: 10-fold CV on targets to predict DTIs between known drugs and new targets, where one target fold is split as a test set.
    \item CVS$_{dt}$: $3 \times 3$-fold CV on both drugs and targets to infer DTIs between new drugs and new targets, where one drug fold and one target fold are split as a test set. 
\end{itemize}

Area under the precision-recall curve (AUPR) and the receiver operating characteristic curve (AUC) are employed as evaluation metrics.  
% To examine the statistical significance of the differences among the competing methods, the Friedman test, followed by the Wilcoxon signed rank test with Bergman-Hommel’s correction at the 5\% level~\cite{Benavoli2016} is applied.
% In regard to the hyperparameter settings of the proposed  
In FGS, we set $k=5$ for all datasets. The filter ratio $\rho$ is set as 0.5 for Yamanishi datasets. For the Luo dataset, $\rho=0.2$ in CVS$_d$ and CVS$_{dt}$ and $\rho=0.7$ in CVS$_t$. 
The hyperparameters of comparing similarity integration methods, DTI prediction models, and base prediction models are set based on the suggestions in the respective articles or chosen by grid search (see Supplementary Table A1 and Section A3).

\subsection{FGS vs. Similarity Integration Methods}

Table \ref{tab:averank_results_SI} summarizes the average ranks of FGS and the six similarity integration competitors with the five base prediction models. Detailed results can be found in Supplementary Tables A2–A7.
FGS achieves the best average rank under all prediction settings in terms of both AUPR and AUC. Furthermore, as shown in Table \ref{tab:statistical_test_SI}, FGS achieves statistically significant improvement over all competitors under all three prediction settings, according to the Wilcoxon signed-rank test with Bergman-Hommel’s correction~\cite{Benavoli2016} at 5\% level. This demonstrates the effectiveness of the selective fine-grained weighting scheme in FGS. 
LIC and SNF-F are the second and third methods. Although LIC also relies on local interaction consistency-based weights, it does not include fine-grained similarity weighting and selection. SNF-F is outperformed by FGS, mainly because its global similarity selection can not exploit the utility of similarity views for individual drugs and targets. 
The following two methods, KA and HSIC, do not remove noisy information and do not exploit the entity-wise similarity utility. AVE and SNF-F are the two worst approaches, since they totally ignore any interaction information, which is vital in the DTI prediction task.

\begin{table}[t]
\centering
\setlength{\tabcolsep}{2pt}
\small
\caption{Average ranks of similarity integration methods under three prediction settings}
\label{tab:averank_results_SI}
\begin{tabular}{@{}ccccccccc@{}}
\toprule
Metric & Setting & AVE & KA & HSIC & LIC & SNF-H & SNF-F & FGS \\ \midrule
\multirow{4}{*}{AUPR} & CVS$_d$ & 5.3 & 4.7 & 4.4 & 2.52 & 6.68 & 3.2 & \textbf{1.2} \\
 & CVS$_t$ & 5.26 & 4.14 & 4.8 & 2.9 & 5.04 & 3.92 & \textbf{1.94} \\
 & CVS$_{dt}$ & 5.3 & 4.9 & 4.8 & 2.44 & 6.92 & 2.24 & \textbf{1.4} \\ \cline{2-9} 
 & \textit{Summary} & 5.29 & 4.58 & 4.67 & 2.62 & 6.21 & 3.12 & \textbf{1.51} \\ \midrule
\multirow{4}{*}{AUC} & CVS$_d$ & 4.98 & 4.65 & 4.48 & 2.73 & 6.85 & 2.78 & \textbf{1.55} \\
 & CVS$_t$ & 4.43 & 4.45 & 4.08 & 3.3 & 5.73 & 3.48 & \textbf{2.55} \\
 & CVS$_{dt}$ & 5.38 & 4.83 & 4.8 & 2.8 & 6.25 & 2.5 & \textbf{1.45} \\ \cline{2-9} 
& \textit{Summary} & 4.93 & 4.64 & 4.45 & 2.94 & 6.28 & 2.92 & \textbf{1.85} \\ \bottomrule
\end{tabular}
\end{table}

\begin{table}[t]
\centering
\setlength{\tabcolsep}{2pt}
\small
\caption{P-values of Wilcoxon signed-rank test with Bergman-Hommel’s correction at 5\% level for similarity integration methods under each prediction setting, where p-values less than 0.05 are highlighted by bold to indicate that FGS achieves statistically significant improvement over the corresponding competitor}
\label{tab:statistical_test_SI}
\begin{tabular}{@{}cccccccc@{}}
\toprule
Metric & Setting & \multicolumn{1}{c}{AVE} & \multicolumn{1}{c}{KA} & \multicolumn{1}{c}{HSIC} & \multicolumn{1}{c}{LIC} & \multicolumn{1}{c}{SNF-H} & \multicolumn{1}{c}{SNF-F} \\ \midrule
\multirow{4}{*}{AUPR} & CVS$_d$ & \textbf{1.3e-4} & \textbf{1.3e-4} & \textbf{1.3e-4} & \textbf{1.3e-4} & \textbf{1.3e-4} & \textbf{1.3e-4} \\
 & CVS$_t$ & \textbf{2.4e-4} & \textbf{5.0e-4} & \textbf{2.4e-4} & \textbf{6.9e-4} & \textbf{4.8e-3} &  \textbf{4.8e-2} \\
 & CVS$_{dt}$ & \textbf{1.3e-4} & \textbf{1.3e-4} & \textbf{1.3e-4} & \textbf{2.6e-4} & \textbf{1.3e-4} & \textbf{7.7e-3} \\ \midrule
\multirow{4}{*}{AUC} & CVS$_d$ & \textbf{1.3e-4} & \textbf{1.3e-4} & \textbf{1.3e-4} & \textbf{5.2e-4} & \textbf{1.3e-4} & \textbf{5.3e-4} \\
 & CVS$_t$ & \textbf{5.9e-3} & \textbf{4.7e-3} & \textbf{5.9e-3} & \textbf{9.7e-3} & \textbf{3.8e-4} & \textbf{3.9e-2} \\
 & CVS$_{dt}$ & \textbf{1.3e-4} & \textbf{1.3e-4} & \textbf{1.3e-4} & \textbf{1.4e-4} & \textbf{1.3e-4} & \textbf{2.6e-2} \\ \bottomrule
\end{tabular}
\end{table}

\begin{figure}[th]
\centering
\includegraphics[width=0.47\textwidth]{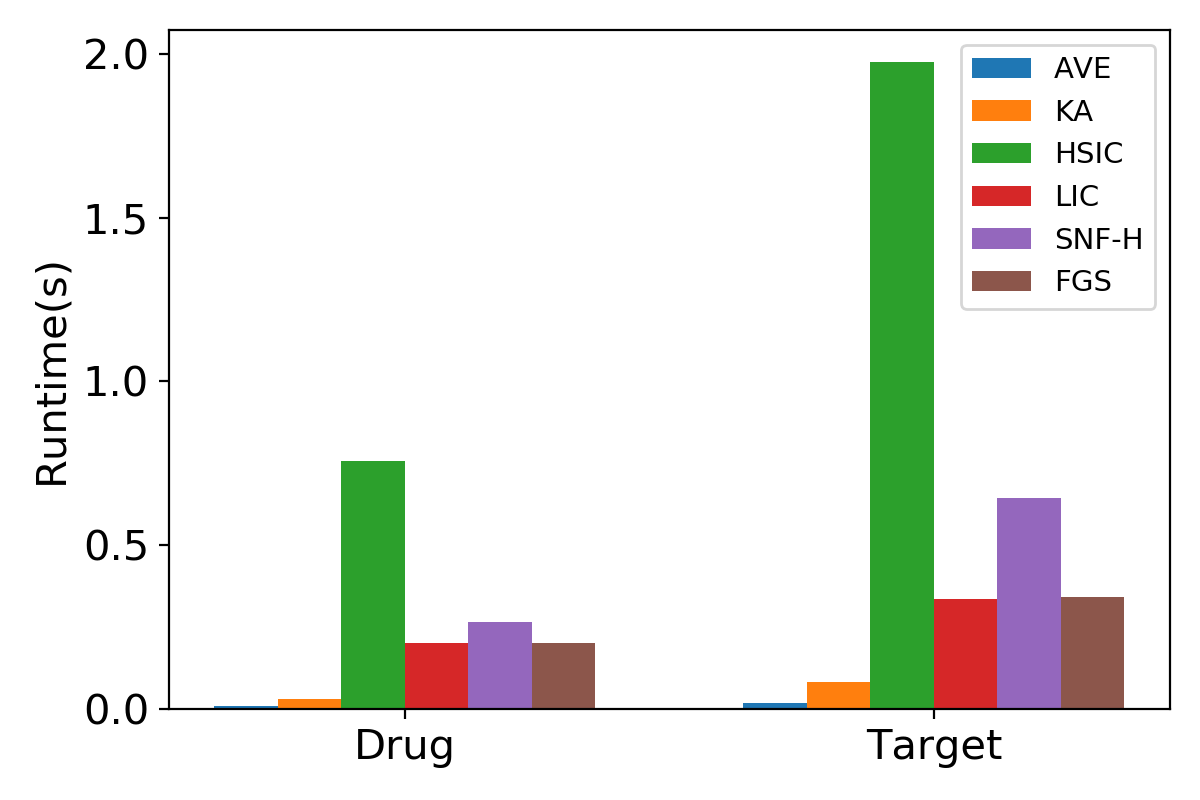}
\caption{Runtime of similarity integration methods on E dataset} 
\label{fig:runtime}
\end{figure}

Figure \ref{fig:runtime} shows the runtime of FGS and other baselines for integrating drug and target similarities of the E dataset. AVE and KA are the two fastest methods due to their efficient linear fusion strategy. LIC and the proposed FGS are slower than the top two, because they search kNNs to calculate the local interaction consistency. The nonlinear approach, SNF-H, which has a cubic computational complexity w.r.t. the drug (target) size, is the fifth. HSIC requires solving a quadratic optimization problem, which leads to the most time-consuming integration procedure.

We further examine the robustness of FGS w.r.t. the variation of base model hyperparameters as well as the test set without homologous drugs, and analyze the sensitivity of hyperparameters in FGS. See Supplementary Section A4.2, A4.3 and A4.4 for more details.
% We do not consider SNF-F in the comparison, since its running time varies based on the used base model.

\subsection{FGS with Base Models vs. DTI Prediction Models}
In this part, we first compare the proposed FGS combined with WkNNRI, denoted as FGS\textsubscript{WkNNRI}, with four traditional machine learning-based DTI prediction methods, namely MSCMF,
KRLSMKL, MKTCMF, and NEDTP, that can directly handle multiple types of similarities. The AUPR and AUC results in three prediction settings are shown in Table \ref{tab:result_AUPR_ML} and \ref{tab:result_AUC_ML}, respectively.
FGS\textsubscript{WkNNRI} is the best method and significantly outperforms all competitors, according to the Wilcoxon signed-rank statistical test with Bergman-Hommel on the results of all prediction settings. This demonstrates the effectiveness of the proposed fine-grained selective similarity fusion manner to extract vital information from multiple similarities. There are some cases, e.g., the Luo dataset in CVS$_d$ and CVS$_t$ and the NR dataset in CVS$_{dt}$, where AUC results of FGS are slightly lower than the corresponding best comparison method (NEDTP, KronRLSMKL, and MKTCMF). Nevertheless, in these cases, the three competitors perform much worse than FGS in AUPR, which is more informative in assessing a model under the extremely sparse interaction distribution.

\begin{table}[h]
\centering
\setlength{\tabcolsep}{1pt}
\small
\caption{AUPR Results of FGS with WkNNRI and traditional machine learning based competitors}
\label{tab:result_AUPR_ML}
\begin{tabular}{@{}ccccccc@{}}
\toprule
Setting & Dataset & MSCMF & KRLSMKL & MKTCMF & NEDTP & FGS\textsubscript{WkNNRI} \\ \midrule
\multirow{5}{*}{CVS$_d$} & NR & 0.44(4) & 0.474(3) & 0.415(5) & 0.486(2) & \textbf{0.551(1)} \\
 & GPCR & 0.46(2) & 0.302(5) & 0.417(4) & 0.451(3) & \textbf{0.547(1)} \\
 & IC & 0.408(2) & 0.236(5) & 0.337(4) & 0.407(3) & \textbf{0.486(1)} \\
 & E & 0.231(4) & 0.208(5) & 0.256(3) & 0.33(2) & \textbf{0.418(1)} \\
 & Luo & 0.371(4) & 0.452(3) & 0.485(2) & 0.077(5) & \textbf{0.49(1)} \\
\multicolumn{2}{c}{\textit{AveRank}} & 3.2 & 4.2 & 3.6 & 3 & \textbf{1} \\ \midrule
\multirow{5}{*}{CVS$_t$} & NR & 0.466(4) & 0.497(2) & 0.49(3) & 0.375(5) & \textbf{0.565(1)} \\
 & GPCR & 0.583(4) & 0.579(5) & 0.726(2) & 0.684(3) & \textbf{0.776(1)} \\
 & IC & 0.781(4) & 0.744(5) & 0.849(2) & 0.8(3) & \textbf{0.855(1)} \\
 & E & 0.574(4) & 0.464(5) & 0.708(2) & 0.615(3) & \textbf{0.725(1)} \\
 & Luo & 0.08(4) & 0.21(3) & 0.248(2) & 0.046(5) & \textbf{0.459(1)} \\
\multicolumn{2}{c}{\textit{AveRank}} & 4 & 4 & 2.2 & 3.8 & \textbf{1} \\ \midrule
\multirow{5}{*}{CVS$_{dt}$} & NR & 0.228(3) & 0.191(4) & 0.242(2) & 0.16(5) & \textbf{0.251(1)} \\
 & GPCR & 0.284(3) & 0.239(5) & 0.248(4) & 0.29(2) & \textbf{0.389(1)} \\
 & IC & 0.232(3) & 0.131(5) & 0.199(4) & 0.251(2) & \textbf{0.339(1)} \\
 & E & 0.1(3) & 0.091(4) & 0.084(5) & 0.112(2) & \textbf{0.241(1)} \\
 & Luo & 0.018(5) & 0.125(3) & 0.129(2) & 0.03(4) & \textbf{0.203(1)} \\
\multicolumn{2}{c}{\textit{AveRank}} & 3.4 & 4.2 & 3.4 & 3 & \textbf{1} \\ \midrule
\multicolumn{2}{c}{\textit{Summary}} & 3.53 & 4.13 & 3.07 & 3.27 & \textbf{1} \\
\multicolumn{2}{c} {p-value} & \textbf{3.2e-3} & \textbf{3.2e-3} & \textbf{3.2e-3} & \textbf{3.2e-3} & - \\ \bottomrule
\end{tabular}
\end{table}

\begin{table}[h]
\centering
\setlength{\tabcolsep}{1pt}
\small
\caption{AUC Results of FGS with WkNNRI and traditional machine learning based competitors}
\label{tab:result_AUC_ML}
\begin{tabular}{@{}ccccccc@{}}
\toprule
Setting & Dataset & MSCMF & KRLSMKL & MKTCMF & NEDTP & FGS\textsubscript{WkNNRI} \\ \midrule
\multirow{5}{*}{CVS$_d$} & NR & 0.779(5) & 0.787(2) & 0.785(4) & 0.786(3) & \textbf{0.825(1)} \\
 & GPCR & 0.884(3) & 0.839(5) & 0.855(4) & 0.885(2) & \textbf{0.911(1)} \\
 & IC & 0.795(2) & 0.767(4) & 0.764(5) & 0.794(3) & \textbf{0.863(1)} \\
 & E & 0.75(5) & 0.819(3) & 0.763(4) & 0.837(2) & \textbf{0.875(1)} \\
 & Luo & 0.897(3) & 0.855(5) & 0.89(4) & \textbf{0.907(1)} & 0.905(2) \\
\multicolumn{2}{c}{\textit{AveRank}} & 3.6 & 3.8 & 4.2 & 2.2 & \textbf{1.2} \\ \midrule
\multirow{5}{*}{CVS$_t$} & NR & 0.755(4) & 0.759(3) & 0.776(2) & 0.726(5) & \textbf{0.817(1)} \\
 & GPCR & 0.891(4) & 0.889(5) & 0.916(3) & 0.918(2) & \textbf{0.95(1)} \\
 & IC & 0.941(3) & 0.935(5) & 0.948(2) & 0.939(4) & \textbf{0.958(1)} \\
 & E & 0.902(3.5) & 0.897(5) & 0.902(3.5) & 0.918(2) & \textbf{0.933(1)} \\
 & Luo & 0.826(5) & \textbf{0.9(1)} & 0.841(4) & 0.86(3) & 0.862(2) \\
\multicolumn{2}{c}{\textit{AveRank}} & 3.9 & 3.8 & 2.9 & 3.2 & \textbf{1.2} \\ \midrule
\multirow{5}{*}{CVS$_{dt}$} & NR & 0.581(3) & 0.552(4) & \textbf{0.616(1)} & 0.548(5) & 0.603(2) \\
 & GPCR & 0.814(3) & 0.792(4) & 0.729(5) & 0.816(2) & \textbf{0.869(1)} \\
 & IC & 0.701(3) & 0.637(5) & 0.65(4) & 0.702(2) & \textbf{0.768(1)} \\
 & E & 0.742(4) & 0.754(2) & 0.619(5) & 0.744(3) & \textbf{0.821(1)} \\
 & Luo & 0.752(4) & 0.795(3) & 0.671(5) & 0.822(2) & \textbf{0.862(1)} \\
\multicolumn{2}{c}{\textit{AveRank}} & 3.4 & 3.6 & 4 & 2.8 & \textbf{1.2} \\ \midrule
\multicolumn{2}{c}{\textit{Summary}} & 3.63 & 3.73 & 3.7 & 2.73 &  \textbf{1.2} \\
\multicolumn{2}{c} {p-value} & \textbf{3.2e-3} & \textbf{3.2e-3} & \textbf{3.2e-3} & \textbf{3.2e-3} & - \\ \bottomrule
\end{tabular}
\end{table}

Furthermore, FGS is compared with four deep learning based DTI prediction models, including NeoDTI, DTIP, DCFME, and SupDTI.  
These four deep learning competitors, inferring new DTIs by mining heterogeneous networks, can only handle the Luo dataset, which could be formulated as a heterogeneous DTI network containing four types of nodes (drugs, targets, diseases, and side effects). Besides, considering that the four deep learning competitors mainly focus on predicting new DTIs between known drugs and known targets, we utilize the 10-fold CVs on drug-target pairs, named CVS$_p$, to simulate this prediction setting. GRMF and NRLMF that can be applied to CVS$_p$ are used as the base model of FGS. As the results shown in Table \ref{tab:Result_Luo_DL}, FGS\textsubscript{GRMF} and FGS\textsubscript{NRLMF} are the top two methods in terms of AUPR, which achieve 6.7\% and 2.5\% improvement over the best baseline (DCFME). DTIP attains the best AUC result, followed by FGS\textsubscript{GRMF} and FGS\textsubscript{NRLMF} that are 4.4\% and 2.8\% lower. However, DTIP suffers a huge decline in AUPR, indicating that it heavily emphasizes the AUC results at the expense of a decrease in AUPR.

Overall, the comparison with various state-of-the-art DTI prediction models under different prediction settings confirms the superiority of FGS. 

\begin{table}[h]
\centering
\setlength{\tabcolsep}{2pt}
\small
\caption{Results of FGS with GRMF and NRLMF and deep learning competitors on Luo dataset under CVS$_p$}
\label{tab:Result_Luo_DL}
\begin{tabular}{@{}ccccccc@{}}
\toprule
Metric & NeoDTI & DTIP & DCFME & SupDTI & FGS\textsubscript{GRMF} & FGS\textsubscript{NRLMF} \\ \midrule
AUPR & 0.573(5) & 0.399(6) & 0.596(3) & 0.585(4) & \textbf{0.639(1)} & 0.611(2) \\
AUC & 0.921(6) & \textbf{0.981(1)} & 0.936(4) & 0.933(5) & 0.939(3) & 0.954(2) \\ \bottomrule
\end{tabular}
\end{table}

%This domesticates that filtering similarities futile to the improvement of overall predicting performance may lead to severe information loss, if the deleted similarities are reliable to a few drugs or targets. 

\subsection{Investigation of Similarity Weights }

\begin{figure*}[h]
\centering
\includegraphics[width=0.8\textwidth]{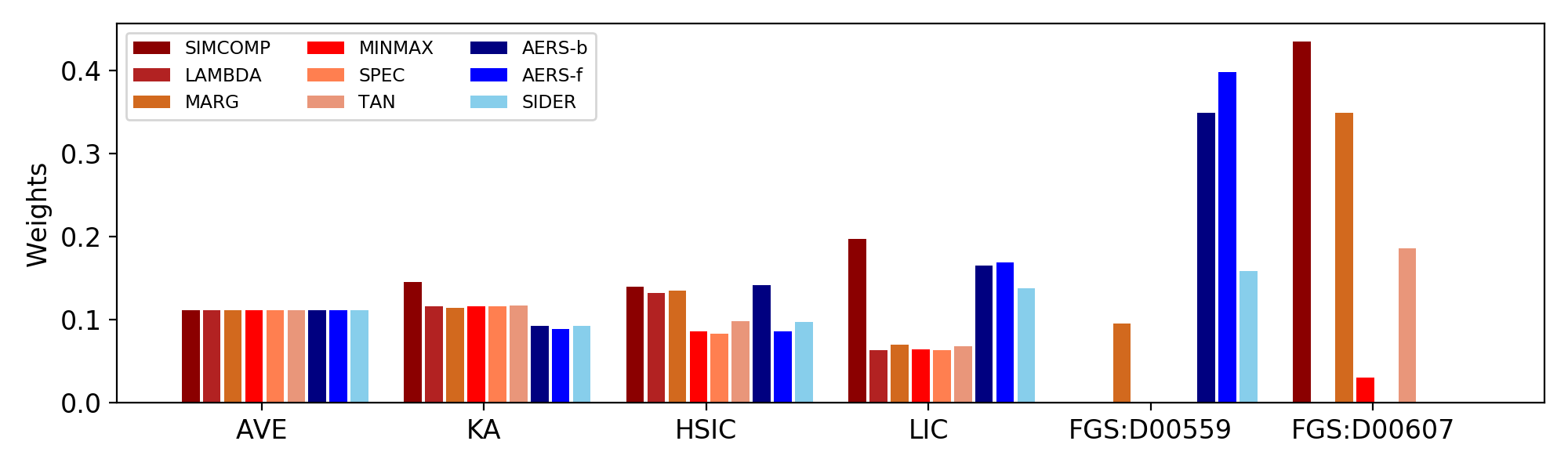}
\caption{Drug similarity weights for the GPCR dataset. Various red and blue bars denote the weights of different chemical structure (CS) and side effect (SE)-based similarities, respectively.} 
\label{fig:sim_weight}
\end{figure*}

Similarity weights play a vital role in the linear similarity integration method, determining the contribution of each view to the final fused similarity space. Hence, we further investigate drug similarity weights on the GPCR dataset generated by FGS and four linear methods.

As shown in Figure  \ref{fig:sim_weight}, AVE assigns equal weight to each type of similarity. KA increases the importance of CS-based similarities. HSIC considers SIMCOMP, LAMBDA, MARG, and AERS-b similarities more informative than others. LIC emphasizes SIMCOMP and three SE-based similarities. All four above methods do not discard any noisy similarity, i.e., all weights are non-zero. 
FGS produces different weight vectors for each drug, so two representative drugs in the GPCR dataset, namely D00559 (pramipexole) and D00607 (guanadrel), are considered in this case study. 

\begin{figure}[!h]
\centering
\subfloat[pramipexole]{\includegraphics[width=0.20\textwidth]{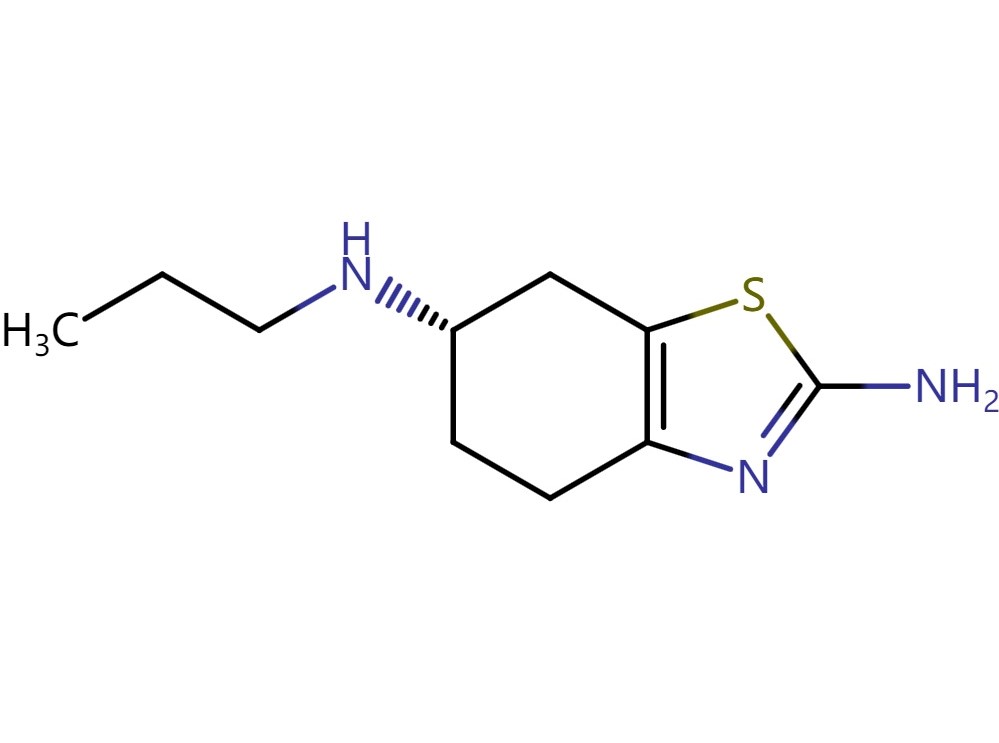}} 
%\hspace{1em}% Space between image A and B
\subfloat[levodopa]{\includegraphics[width=0.20\textwidth]{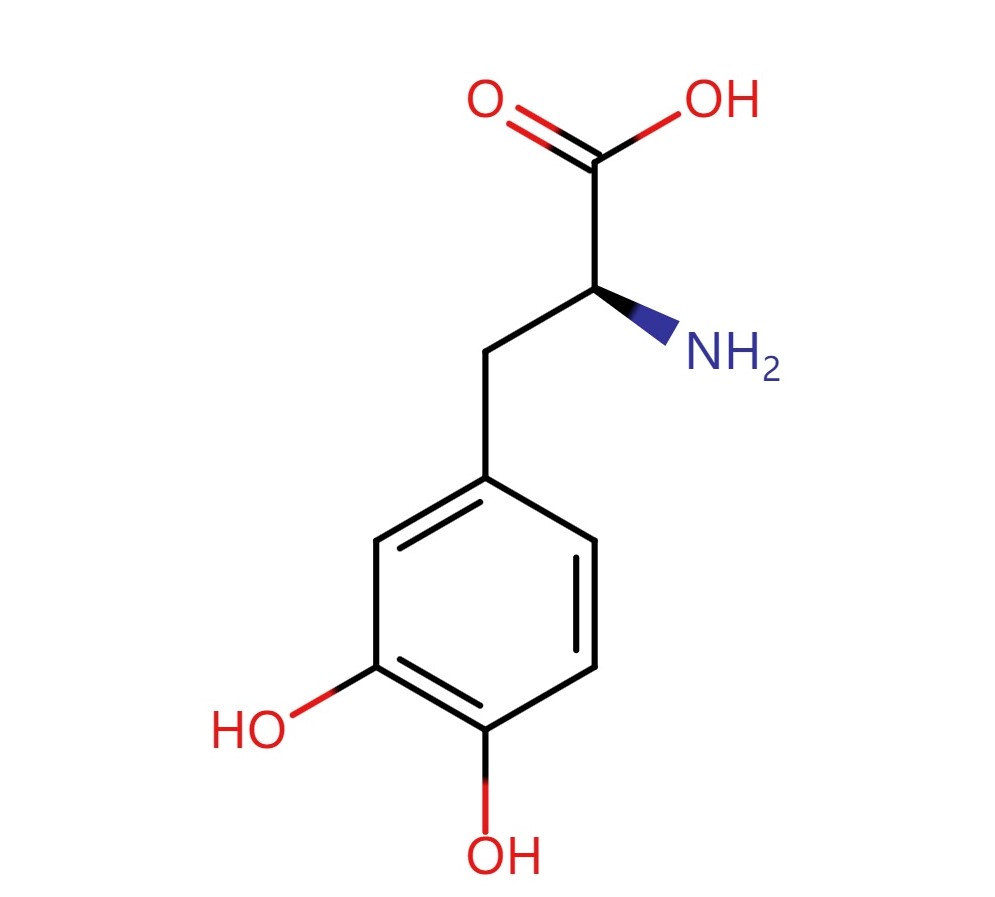}} \\
\subfloat[guanadrel]{\includegraphics[width=0.20\textwidth]{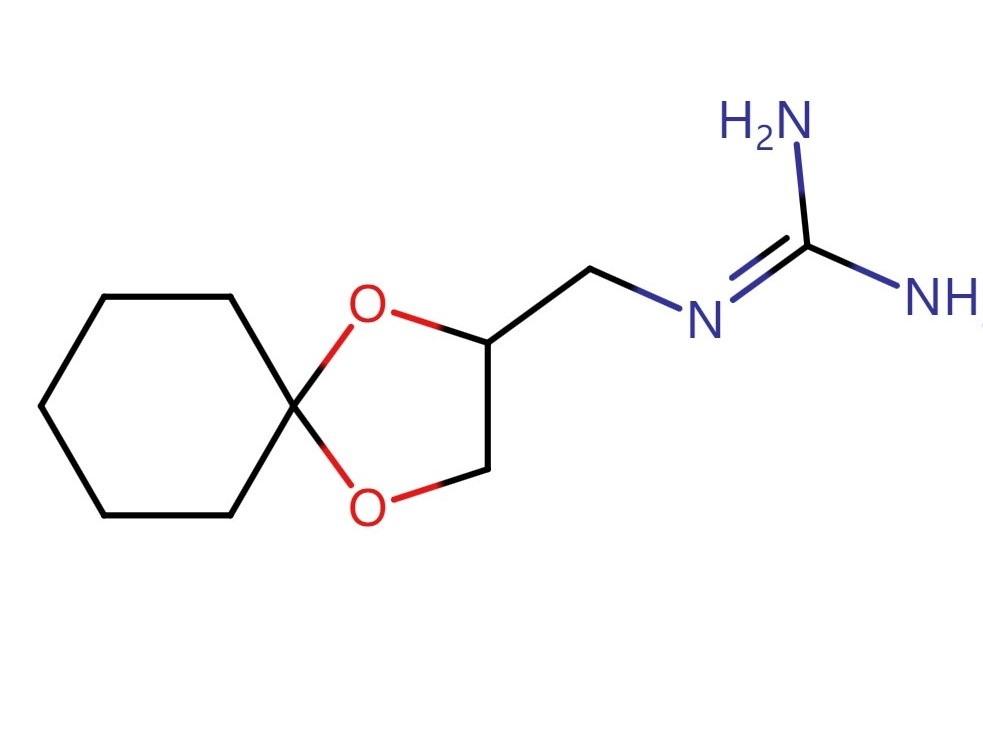}}
\subfloat[guanethidine]{\includegraphics[width=0.20\textwidth]{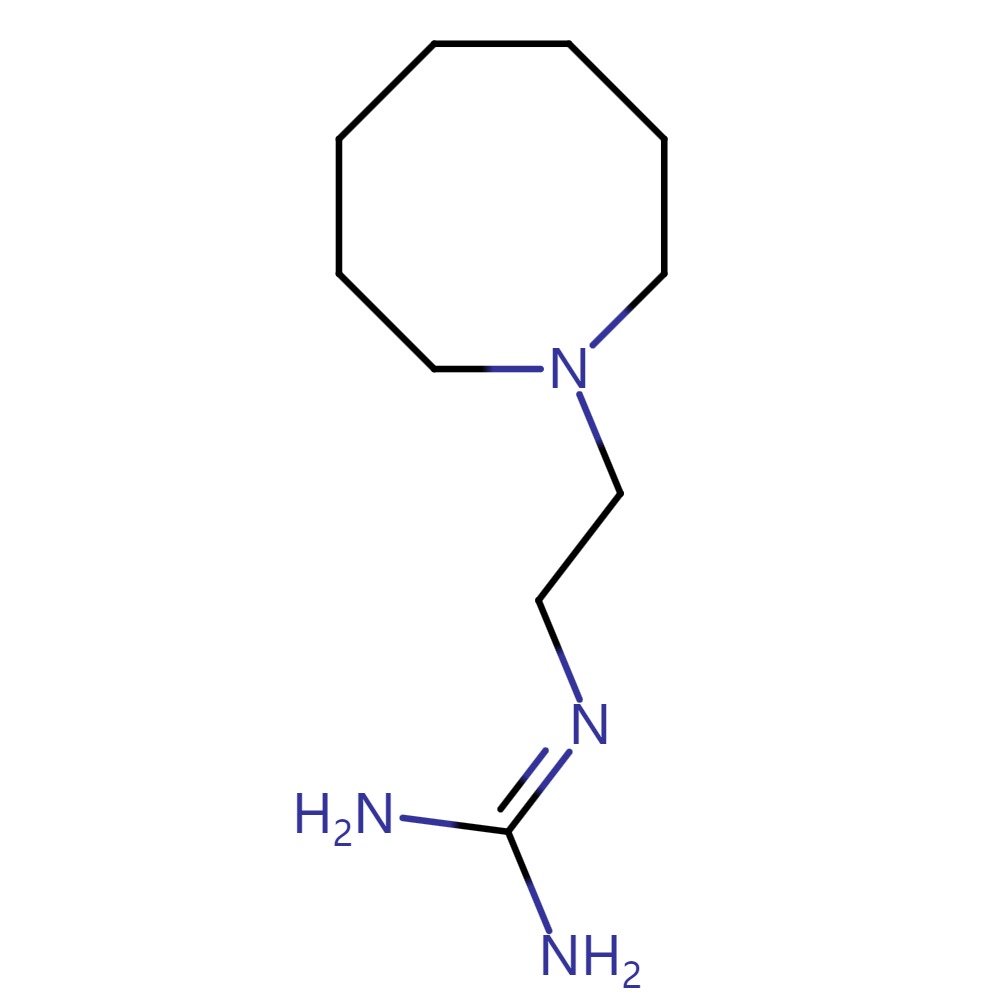}}
\caption{Chemical structures of pramipexole, levodopa, guanadrel, and guanethidine.}
\label{fig:Chemical_Structure}
\end{figure}

Pramipexole, a commonly used Parkinson's disease (PD) medication, interacts with three types of dopaminergic receptors, e.g., DRD2, DRD3, and DRD4, whose genetic polymorphisms are potential determinants of PD development, symptoms, and treatment response \cite{magistrelli2021polymorphisms}. D00059 (levodopa), which also interacts with dopaminergic receptors, could be combined with pramipexole to relief the symptoms in PD \cite{tayarani2010pramipexole_levodopa1,huang2020pramipexole_levodopa2}. The two drugs frequently coexist in reported adverse events, resulting in larger AERS-f and AERS-b similarities. 
Nevertheless, their chemical structures are discrepant, since they do not share many common substructures. Pramipexole (Figure \ref{fig:Chemical_Structure} (a)) is a member of the class of benzothiazoles (\ce{C7H5NS}) in which the hydrogens at the 2 and 6-pro-S-positions are substituted by amino (\ce{-NH2}) and propylamino groups (\ce{C3H9N}), respectively. While levodopa (Figure \ref{fig:Chemical_Structure} (b)) is a hydroxyphenylalanine (\ce{C9H11NO3}) carrying hydroxy substituents (\ce{-OH}) at positions 3 and 4 of the benzene ring. 
%i.e., levodopa ranks 148th out of 223 drugs w.r.t. pramipexole according to the SIMCOMP similarity. 
Therefore, for pramipexole, FGS assigns higher weights to SE-based similarities that are more reliable in identifying drugs interacting with the same targets as pramipexole (e.g., levodopa), and treats most CS-based ones as noise.

Guanadrel is an antihypertensive that inhibits the release of norepinephrine from the sympathetic nervous system to suppress the responses mediated by alpha-adrenergic receptors, leading to the relaxation of blood vessels and the improvement of blood circulation \cite{palmer1983guanadrel,reid1986Alpha_adrenergic_receptors}. D02237 (guanethidine), another medication to treat hypertension, is pharmacologically and structurally similar to guanadrel, which also interacts with alpha-adrenergic receptors and is identified as the most similar drug to guanadrel based on the SIMCOMP similarity. As shown in Figure \ref{fig:Chemical_Structure} (c) and (d), both of them include guanidine (\ce{CH5N3}). 
On the other hand, the side effects of guanadrel and guanethidine are not included in the SIDER database and are merely recorded in AERS, causing the invalidity of SE-based similarities. Therefore, when it comes to guanadrel, FGS emphasizes CS-based similarities and discards all SE-based ones.

Compared with other linear integration methods, FGS not only distinguishes the importance of similarities for each drug based on the corresponding pharmacological properties, but also diminishes the influence of noisy similarities.

\subsection{Discovery of New DTIs}
We check the validity of novel DTIs discovered by FGS from the Luo dataset. Specifically, we leverage 10-fold CVs on non-interacting pairs, i.e., using all interacting pairs and nine folds of non-interacting pairs as the training set, and leaving the remaining non-interacting fold as the test set. The top 10 non-interacting pairs with the highest prediction scores across all ten folds are considered new interacting candidates. Table \ref{tab:new_DTIs} lists the new DTIs found by FGS\textsubscript{GRMF} and FGS\textsubscript{NRLMF}, along with their supportive evidence retrieved from DrugBank (DB)~\cite{Wishart2018DrugBank2018} and DrugCentral (DC)~\cite{Avram2021DrugCentralRepositioning} databases. There are 18 out of 20 new interactions confirmed by at least one database, which is comparable to other similarity integration methods (Supplementary Table A10), demonstrating the effectiveness of FGS in trustworthy new DTI discovery.
In addition, compared with new DTIs predicted by other similarity integration methods (Supplementary Tables A8–A9), there are three novel DTIs (bold ones in Table \ref{tab:new_DTIs}) only discovered by FGS, including two unverified ones.

\begin{comment}
\begin{table}[h]
\centering
\setlength{\tabcolsep}{1pt}
\small
\caption{Top 10 new DTIs discovered by FGS\textsubscript{GRMF} and FGS\textsubscript{NRLMF} from the Luo dataset}
\label{tab:new_DTIs}
\begin{tabular}{@{}c|ccc|ccc@{}}
\toprule
\multirow{2}{*}{Rank} & \multicolumn{3}{c|}{FGS\textsubscript{GRMF}} & \multicolumn{3}{c}{FGS\textsubscript{NRLMF}} \\ 
  & Drug ID & Target ID & Database & Drug ID & Target ID & Database \\ \midrule
1 & DB00829 & P48169 & DB & DB00829 & P48169 & DB \\
2 & DB01215 & P48169 & DB & DB01215 & P48169 & DB \\
3 & DB00363 & P21918 & DC & DB00363 & P21918 & DC \\
4 & DB00734 & P28222 & DC & DB06216 & P21918 & DC \\
5 & DB06800 & P41143 & DC & DB06216 & P28221 & DC \\
6 & DB00652 & P41143 & DC & DB00696 & P08908 & DB, DC \\
7 & \textbf{DB06216} & \textbf{P21918} & \textbf{DC} & DB00543 & P28223 & DB, DC \\
8 & DB00333 & P41145 & DC & DB00652 & P41143 & DC \\
9 & DB06216 & P28221 & DC & \textbf{DB01273} & \textbf{Q15822} & \textbf{-} \\
10 & DB00734 & P21918 & DC & \textbf{DB01186} & \textbf{P34969} & \textbf{-} \\ \bottomrule
\end{tabular}
\end{table}
\end{comment}

\begin{table*}[h]
\centering
\setlength{\tabcolsep}{1pt}
\small
\caption{Top 10 new DTIs discovered by FGS\textsubscript{GRMF} and FGS\textsubscript{NRLMF} from the Luo dataset}
\label{tab:new_DTIs}
\begin{tabular}{@{}c|ccccc|ccccc@{}}
\toprule
\multirow{2}{*}{Rank} & \multicolumn{5}{c|}{FGS\textsubscript{GRMF}} & \multicolumn{5}{c}{FGS\textsubscript{NRLMF}} \\ 
  & Drug ID & Drug Name & Target ID & Target Name & Database & Drug ID & Drug Name & Target ID & Target Name & Database \\ \midrule
1 & DB00829 & Diazepam & P48169 & GABRA4 & DB & DB00829 & Diazepam & P48169 & GABRA4 & DB \\
2 & DB01215 & Estazolam & P48169 & GABRA4 & DB & DB01215 & Estazolam & P48169 & GABRA4 & DB \\
3 & DB00363 & Clozapine & P21918 & DRD5 & DC & DB00363 & Clozapine & P21918 & DRD5 & DC \\
4 & DB00734 & Risperidone & P28222 & HTR1B & DC & DB06216 & Asenapine & P21918 & DRD5 & DC \\
5 & DB06800 & Methylnaltrexone & P41143 & OPRD1 & DC & DB06216 & Asenapine & P28221 & HTR1D & DC \\
6 & DB00652 & Pentazocine & P41143 & OPRD1 & DC & DB00696 & Ergotamine & P08908 & HTR1A & DB, DC \\
7 & \textbf{DB06216} & \textbf{Asenapine} & \textbf{P21918} & \textbf{DRD5} & \textbf{DC} & DB00543 & Amoxapine & P28223 & HTR2A & DB, DC \\
8 & DB00333 & Methadone & P41145 & OPRK1 & DC & DB00652 & Pentazocine & P41143 & OPRD1 & DC \\
9 & DB06216 & Asenapine & P28221 & HTR1D & DC & \textbf{DB01273} & \textbf{Varenicline} & \textbf{Q15822} & \textbf{CHRNA2} & \textbf{-} \\
10 & DB00734 & Risperidone & P21918 & DRD5 & DC & \textbf{DB01186} & \textbf{Pergolide} & \textbf{P34969} & \textbf{HTR7} & \textbf{-} \\ \bottomrule
\end{tabular}
\end{table*}

\begin{figure*}[!h]
\centering
\subfloat[DB01273 (varenicline)-Q15822 (CHRNA2)]{\includegraphics[width=0.4\textwidth]{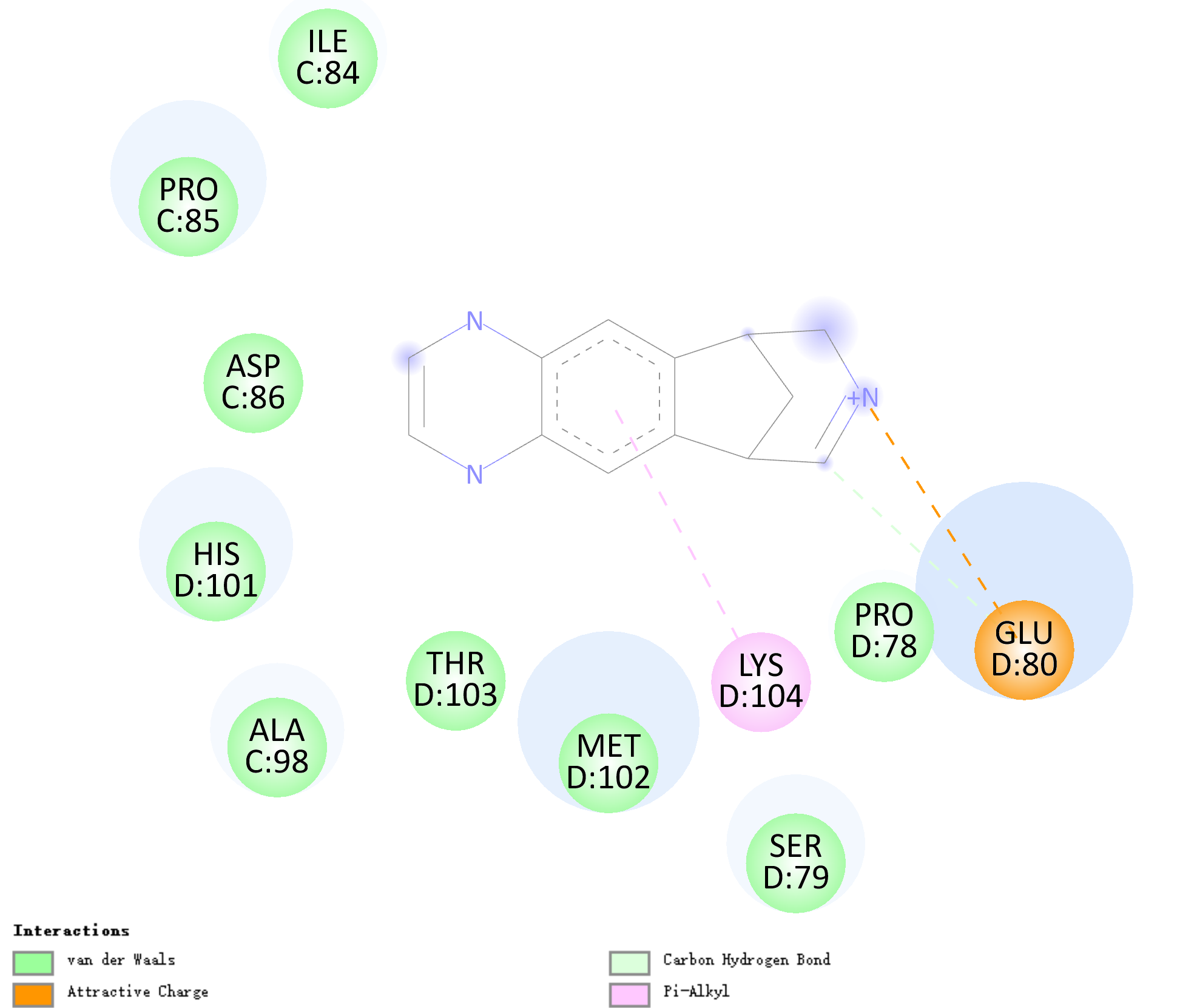}} 
%\hspace{1em}% Space between image A and B
\subfloat[DB01186 (pergolide)-P34969 (HTR7)]{\includegraphics[width=0.4\textwidth]{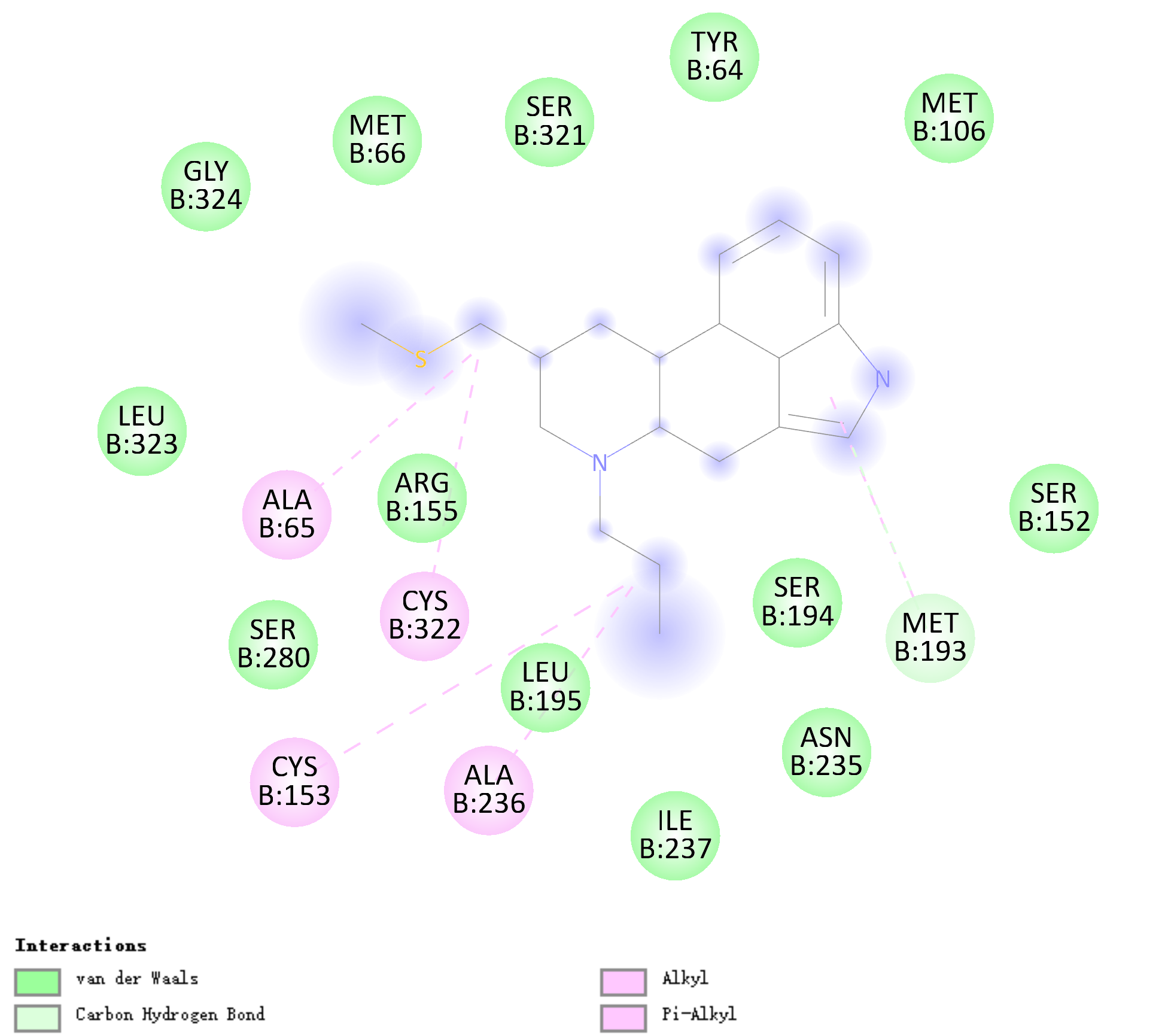}} 
\caption{2D visualization of docking results}
\label{fig:Docking_2d}
\end{figure*}

For the two drug-target pairs (DB01273-Q15822 and DB01186-P34969) that have not been recorded in the databases, we check their binding affinities by conducting docking simulations with PyRx\footnote{\href{https://pyrx.sourceforge.io/}{https://pyrx.sourceforge.io/}}. 
%\BL{REF:Small-molecule library screening by docking with PyRx}. 
The 2D and 3D visualizations of docking interactions of DB01273 with Q15822 and DB01186 with P34969, produced by Discovery Studio\footnote{\href{https://discover.3ds.com/discovery-studio-visualizer-download}{https://discover.3ds.com/discovery-studio-visualizer-download}}, are shown in Figure \ref{fig:Docking_2d} and Supplementary Figure A6. 
The binding energy between DB01273 and Q15822 is -8.9 kcal$\backslash$mol. DB01273 binds with Q15822 by forming a Pi-alkyl bond with Lys104, an attractive charge and a carbon hydrogen bond with Glu80, and Van der Waals bonds with nine amino acid residues. 
DB01273 (varenicline) is a medicine for smoking cessation via partial agonists at nicotinic acetylcholine receptors, which is more likely to interact with Q15822 (CHRNA2, alpha-2 subtype of neuronal acetylcholine receptor). 
Regarding DB01186 and P34969, 13 residues of P34969 are involved in Van der Waals interactions, five residues are bonded to DB01186 through Pi-alkyl or alkyl interactions, and Met193 is also linked with the ligand through carbon hydrogen interaction, which results in the binding score of -7.2 kcal$\backslash$mol. 
DB01186 (pergolide), a treatment for Parkinson’s Disease, interacts with 5-hydroxytryptamine receptors (HTR1A, HTR2B, HTR1D, etc.) that have similar function (serotonin receptor activity) with P34969 (HTR7).
The docking results of two drug-target pairs indicate that their potential interactions are reliable.

% amino acid residues with different interactions 
% There are Q15822 (Neuronal acetylcholine receptor subunit alpha-2)
%The  docking results of DB01273 (Varenicline)-Q15822 (Neuronal acetylcholine receptor subunit alpha-2) and DB01186 (Pergolide)-P34969 (5-hydroxytryptamine receptor 7), generated by Discovery Studio Visualizer \BL{FootNote: https://discover.3ds.com/discovery-studio-visualizer-download} are shown in   

\section{Conclusion}
\label{sec:conclusion}
This paper proposed FGS, a fine-grained selective similarity integration approach, to produce more compact and beneficial inputs for DTI prediction models via entity-wise utility-aware similarity selection and fusion.
%entity-wise similarity utility-aware crucial information extraction and noisy similarity filter.
We conducted extensive experiments under various DTI prediction settings, and the results demonstrate the superiority of FGS to the state-of-the-art similarity integration methods and DTI prediction models with the cooperation of simple base models. Furthermore, the case studies verify the importance of the fine-grained similarity weighting scheme and the practical ability of FGS to assist prediction models in discovering reliable new DTIs.

In the future, we plan to extend our work to handle heterogeneous information from multi-modal views, i.e., drug and target information in different formats, such as drug chemical structure graphs, target genomic sequences, textual descriptions from scientific literature, and vectorized feature representations.

\hspace*{\fill} % an new line without any content

\noindent\fbox{\parbox{0.46\textwidth}{\textbf{Key Points}:
\begin{itemize}
    \item FGS is an entity-wise utility-aware similarity integration approach that generates a reliable fused similarity matrix for any DTI prediction model by capturing crucial information and removing noisy similarities at a finer granularity.
    \item Extensive experimental studies on five public DTI datasets demonstrate the effectiveness of FGS.
    \item Case studies further exemplify that the fine-grained similarity weighting scheme used in FGS is more suitable than conventional ones for DTI prediction with multiple similarities.
\end{itemize}}}

\section*{Data and Code availability}
The source code and data of this work can be found at 
\href{https://github.com/Nanfeizhilu/FGS_DTI_Predition}{https://github.com/Nanfeizhilu/FGS\_DTI\_Predition}

%\section{Author contributions statement}
%Must include all authors, identified by initials, for example: S.R. and D.A. conceived the experiment(s),  S.R. conducted the experiment(s), S.R. and D.A. analysed the results.  S.R. and D.A. wrote and reviewed the manuscript.

\section*{Funding}
This work was partially supported by Science Innovation Program of Chendu-Chongqing Economic Circle in Southwest China (grant no. KJCXZD2020027)
% This work was supported by the China Scholarship Council (CSC) [201708500095]; and the French National Research Agency (ANR) under the JCJC project GraphIA [ANR-20-CE23-0009-01].

\section{Biography}
\textbf{Bin Liu} is a lecturer at Key Laboratory of Data Engineering and Visual Computing, Chongqing University of Posts and Telecommunications. 
%and received his PhD Degree in computer science from Aristotle University of Thessaloniki. 
His research interests include multi-label learning and bioinformatics.

%\textbf{Dimitrios Papadopoulos} is a PhD student at the School of Informatics, Aristotle University of Thessaloniki. His research interests include supervised machine learning, graph mining, and drug discovery.

%\textbf{Fragkiskos D. Malliaros} is an Assistant Professor at Paris-Saclay University, CentraleSupélec and associate researcher at Inria Saclay. His research interests include graph mining, machine learning, and graph-based information extraction.

\textbf{Jin Wang} is a Professor at Key Laboratory of Data Engineering and Visual Computing, Chongqing University of Posts and Telecommunications. His research interests include large scale data mining and machine learning.
 
\textbf{Kaiwei Sun} is an Associate Professor at Key Laboratory of Data Engineering and Visual Computing, Chongqing University of Posts and Telecommunications. His research interests in data mining and machine learning.

\textbf{Grigorios Tsoumakas} is an Associate Professor at the Aristotle University of Thessaloniki. His research interests include machine learning (ensembles, multi-target prediction) and natural language processing (semantic indexing, keyphrase extraction, summarization)
 
%\textbf{Apostolos N. Papadopoulos} is Associate Professor at the School of Informatics, Aristotle University of Thessaloniki. His research interests include data management, data mining and big data analytics.

%\bibliographystyle{plain}
\bibliographystyle{unsrt}
%\bibliography{LB}
\bibliography{BL_Nov_25}

\begin{comment}

%USE THE BELOW OPTIONS IN CASE YOU NEED AUTHOR YEAR FORMAT.
%\bibliographystyle{abbrvnat}
%\bibliography{reference}

%% sample for biography with author's image
\begin{biography}{{\color{black!20}\rule{77pt}{77pt}}}{\author{Author Name.} This is sample author biography text. The values provided in the optional argument are meant for sample purposes. There is no need to include the width and height of an image in the optional argument for live articles. This is sample author biography text this is sample author biography text this is sample author biography text this is sample author biography text this is sample author biography text this is sample author biography text this is sample author biography text this is sample author biography text.}
\end{biography}

%% sample for biography without author's image
\begin{biography}{}{\author{Author Name.} This is sample author biography text this is sample author biography text this is sample author biography text this is sample author biography text this is sample author biography text this is sample author biography text this is sample author biography text this is sample author biography text.}
\end{biography}

\end{comment}

\end{document}

\maketitle

\renewcommand{\thesection}{A\arabic{section}}
\captionsetup[table]{name={Supplementary Table},labelsep=colon}
\renewcommand{\thetable}{A\arabic{table}}
\captionsetup[figure]{name={Supplementary Figure},labelsep=colon}
\renewcommand{\thefigure}{A\arabic{figure}}

\section{Base and Comparing DTI Prediction Models}
In this part, we review representative DTI prediction approaches, all of which are employed in our empirical study as either a base model or a comparing method. Those methods are categorized into three groups according to their required input format and used machine learning technique.
% their ability to handle multiple types of similarities

\textbf{1)} Traditional machine learning technique-based DTI prediction methods that can handle a single type of drug and target similarities only:
\begin{itemize}
     \item \textbf{WkNNIR}~\cite{Liu2022Drug-targetRecovery} is a neighborhood method that recovers potential missing interactions and predicts new DTIs based on the interaction information of proximate known drugs and targets.
     \item \textbf{NRLMF}~\cite{Liu2016NeighborhoodPrediction} is a matrix factorization model, which employs a weighted logistic loss function to emphasize the importance of the minority interacting data, and applies neighborhood regularization on drug and target similarities to preserve the local invariance of the learned latent feature space.
    \item \textbf{GRMF}~\cite{Ezzat2017Drug-targetFactorization}, another matrix factorization approach, performs graph regularization on latent features to learn a manifold for label propagation in drug and target spaces, and incorporates a neighborhood based interaction recovery procedure to mitigate the inaccuracy of the supervised information caused by possible undetected interactions.
    \item \textbf{BRDTI}~\cite{Peska2017} is a ranking method, which learns an interaction value based ideal target sequence for each drug and utilizes the content alignment based regularization to capture structural similarities of drugs and targets. 
    \item \textbf{DLapRLS}~\cite{Ding2020IdentificationFusion} is a kernel classifier that incorporates both drug and target similarity spaces into the graph regularized least squares (RLS) framework, which can be solved by an efficient alternating least squares algorithm. 
\end{itemize}
These five approaches in Group 1 are the base models of similarity integration methods, leveraging the fused drug and target similarity matrices produced by a similarity integration method as input.

\textbf{2)} Traditional machine learning technique based DTI prediction methods that can process multiple types of drug and target similarities:
\begin{itemize}
    \item \textbf{MSCMF}~\cite{Zheng2013CollaborativeInteractions} is a matrix factorization model that learns two low-rank features that align to linear combinations of multiple drug and target similarities derived from diverse data sources, respectively.
    \item \textbf{KRLSMKL}~\cite{Nascimento2016APrediction} is a kernel model that unifies linear multiple kernel fusion and Kronecker kernel based RLS classifier training in one optimization objective.  
    \item \textbf{MKTCMF}~\cite{Ding2022_MKTCMF} is a DTI prediction method that incorporates both multiple kernel learning and triple collaborative matrix factorization int a uniform optimization framework. 
    \item \textbf{NEDTP}~\cite{An2021AInteractions} is a network embedding based model, which generates drug and target embeddings using a Node2Vec model with random walks sampled from diverse drug and target related sub-networks, and trains a Gradient Boosting Decision Tree classifier using the binary dataset constructed upon the learned embeddings to predict new interactions.
\end{itemize}

\textbf{3)} Deep Learning DTI prediction models that typically works on the heterogeneous DTI network:
\begin{itemize}
    \item \textbf{NeoDTI}~\cite{Wan2019NeoDTI:Interactions} implements a graph convolution architecture to leverage information from multiple heterogeneous sources and conducts the network reconstruction to discover new DTIs. 
    \item \textbf{DTIP}~\cite{Xuan2021IntegratingPrediction} uses a feature-level attention mechanism to extract fine-grained information for drug, target, and disease nodes. It also conducts random walk with restart in the heterogeneous network to acquire node sequences, which are processed by a Bi-GRU based model with attention mechanism to generate the drug and target embeddings. 
    \item \textbf{DCFME}~\cite{chen2022DCFME} leverages CNN and MLP to learn drug and target features that capture both global/local and deep representations of drug-target couplings, and employs the focal loss to emphasize the hard examples during the training procedure.  
    \item \textbf{SupDTI}~\cite{chen2022SupDTI}, a deep learning DTI prediction framework enhanced by the self-supervised learning strategy, incorporates a contrastive learning and a generative learning modules to improve the global and local node-level agreement of learned embeddings in the heterogeneous DTI network. 
\end{itemize}
The eight DTI prediction models in Groups 2 and 3, which can handle input information from multiple data sources, are considered as the competitors of our FGS method collaborating with base models.

\section{Similarities}

\subsection{Updated Golden Standard Datasets}

Following~\cite{Nascimento2016APrediction}, we leverage different characteristics or profiles to compute nine drug similarities. \textbf{SIMCOMP} similarity is obtained by using SIMCOMP algorithm to assess the ratio of common substructures between two drugs based on their chemical graphs~\cite{Hattori2003DevelopmentPathways}. In addition, we also compute other five chemical structure based similarities using various graph kernels, namely \textbf{LAMBDA} (the Lambda-k kernel) \cite{Lambdak_MINMAX_Kernel}, \textbf{MARG} (the Marginalized kernel)~\cite{KashimaTI03_Marginalized_Kernel}, \textbf{MINMAX} (the MINMAX kernel)~\cite{Lambdak_MINMAX_Kernel}, \textbf{SPEC} (the Spectrum kernel)~\cite{RalaivolaSSB05_Spectrum_Tanimoto_Kernel} and \textbf{TAN} (the Tanimoto kernel)~\cite{RalaivolaSSB05_Spectrum_Tanimoto_Kernel}. Furthermore, we retrieve three kinds of side effect profiles to describe the pharmacological properties of drugs, and compute the weighted cosine correlation coefficient between associated profiles for each drug pair to obtain three side effects based drug similarities. Specifically, for \textbf{AERS-f} and \textbf{AERS-b} similarities, each drug is described by a profile (vector) that records the frequency and presence of side effect keywords in the adverse event reports system (AERS)~\cite{Takarabe2012DrugApproach}, respectively. For \textbf{SIDER} similarity, each drug is represented by a binary profile that reports its verified side effects from the SIDER database~\cite{Kuhn2016TheEffects}. 

We compute nine target similarities derived from various sources to describe the relation between targets in multi-aspects~\cite{Nascimento2016APrediction}. Seven measurements are used to obtain different target similarities based on the amino-acid protein sequence, including \textbf{SW} (normalized Smith-Waterman scores), \textbf{MIS-k3m1}, \textbf{MIS-k4m1}, \textbf{MIS-k3m2}, \textbf{MIS-k4m2} (four Mismatch kernels with different parameterization, e.g., $k=3$ and $m=1$, $k=4$ and $m=1$, etc., where $k$ is the mers (subsequence) length and $m$ is the number of maximal mismatches per $k$-mer)~\cite{LeslieEWN02_Mismatch_Kernel}, \textbf{SPEC-k3} and \textbf{SPEC-k4} (two Spectrum kernels with length of mers $k$ being 3 and 4 respectively)~\cite{LeslieEN02_Spectrum_Protein_Kernel}.
\textbf{GO} similarity reflects the functional semantic likeness between targets, where the gene ontology term annotation profiles of targets are retrieved from the BioMART database~\cite{smedley2015biomart}, and the semantic similarity scores between annotations of two targets are computed using the Resnik algorithm~\cite{resnik1999semantic}.
\textbf{PPI} similarity indicates the shortest distance between two target proteins in the human protein-protein network (PPI) retrieved from the BioGRID database~\cite{StarkBRBBT06_BioGRID}.

\subsection{Luo Dataset}
The Luo dataset is originally formulated as a heterogeneous network that contains four types of nodes (i.e., drugs, targets (proteins), diseases and side effects) and six types of edges (i.e., DTIs, drug–drug interactions, drug–disease associations, drug–side effect associations, target–disease associations and target-target interactions). Drug-target interactions and drug-drug interactions are extracted from  DrugBank3.0~\cite{Wishart2018DrugBank2018}. SIDER (version 2)~\cite{Kuhn2016TheEffects} provides drug side effects associations. Target–target interactions are obtained from HPRD database (Release 9)~\cite{KeshavaPrasad2009HumanUpdate}. Drug–disease and target–disease associations are downloaded from Comparative Toxicogenomics Database~\cite{Davis2013The2013}.

We construct four types of drug similarities based on the chemical structure and associations /interactions embedded in the heterogeneous network constructed by Luo. Similar with Yaminish's datasets, \textbf{TAN} assesses the common chemical sub-structure of two drugs using the Tanimoto kernel~\cite{RalaivolaSSB05_Spectrum_Tanimoto_Kernel}. For \textbf{DDI} (drug-drug interaction), \textbf{DIA} (drug-disease association) and \textbf{SE} (drug-side effect association) similarities, the similarity value between two drugs is computed as the Jaccard coefficient of their interacting/associating entity sets. For example, given two drugs $d_i$ and $d_j$, as well as their associated disease set $DI_i$ and $DI_j$, the drug–disease association based similarity between $d_i$ and $d_j$ is computed as:
\begin{equation}
   S^{DI}_{ij} = \frac{|DI_i \cap DI_j|}{|DI_i \cup DI_j|} 
\end{equation}

We construct three kinds of target similarities based on the target sequence and target-related network topology. \textbf{SW} similarity uses the normalized Smith-Waterman score to assess the similarity between two target sequences. For \textbf{PPI} (protein-protein interaction, which is equivalent to target-target interaction) and \textbf{TIA} (target-disease association) similarities, the similarity value between two targets is defined as the Jaccard coefficient of their interacting/associating entity sets, which is the same as drug interaction/association based similarities.

\section{Hyperparameter Settings}

The hyperparameters of comparing methods, including similarity integration and DTI prediction approaches, and base models used in our experiments are listed in Supplementary Table \ref{tab:param_baseline}. Their hyperparameters are either set as a fixed value according to the suggestions in the respective papers or tuned by using the grid search. As the example of GRMF, $\eta=0.7$ indicates that a fixed value is used for this parameter, and 0.7 is suggested by the corresponding paper~\cite{Ezzat2017Drug-targetFactorization}. On the other hand, $r$, another hyperparameter in GRMF, is set as the optimal value (to achieve the best prediction performance, e.g., the sum of AUC and AUPR result) chosen from $\{50,100\}$. For a method with more than one hyperparameter to be tuned, all hyperparameter value combinations are used as searching candidates. For example, GRMF needs to tune four hyperparameters, including $r$ with two values, $\lambda_l$, $\lambda_d$, and $\lambda_t$ with four values, so the number of hyperparameter setting candidates is $2 \times 4 \times 4 \times4 =128$.

\begin{table*}[h]
\centering
\footnotesize
\caption{Hyperparameter Settings for Comparing Methods and Base Models}
\label{tab:param_baseline}
\begin{tabular}{@{}ccc@{}}
\toprule
% & \multicolumn{2}{c}{\textbf{Similarity Integration Baselines}} \\ \hline
Category & Method & Hyperparameter  Settings \\ \hline
\multirow{6}{*}{\begin{tabular}[c]{@{}c@{}} \textbf{Similarity} \\ \textbf{Integration} \\ \textbf{Baselines} \end{tabular}} & AVE & - \\
& KA & - \\
& HSIC & $v_1, v_2 \in \{2^{-1},2^{-2},2^{-3},2^{-4}\}$ \\
& LIC & $k=5$ \\
& SNF-H & $k=5$, $\#iters=2$, $\alpha=1$, $c_1=0.7$, $c_2=0.6$ \\
& SNF-F & $k=5$, $\#iters=2$, $\alpha=1$ \\ \midrule
% & \multicolumn{2}{c}{\textbf{Base Models}} \\ \hline
% Method & Parameter  Settings \\ \hline
\multirow{6}{*}{\begin{tabular}[c]{@{}c@{}} \textbf{Base} \\ \textbf{Models} \end{tabular}} & WkNNIR & $k \in \{1,2,3,5,7,9\}$, $\eta \in \{0.1,0.2,0.3…1.0\}$ \\
& DLapRLS & $\lambda_d, \lambda_t \in \{0.9, 0.95, 1.0\}$, $u \in \{0.9,1.0,1.1\}$ \\
& GRMF & $\eta=0.7$, $k=5$, $r \in   \{50,100\}$, $\lambda_d, \lambda_t \lambda_r \in   \{2^{-4},2^{-2},2^0,2^2\}$ \\
& NRLMF & $c$=5, $k1=k2=5$, $\theta=0.1$, $r \in \{50,100\}$ , $\lambda_d,   \lambda_t, \alpha, \beta \in \{2^{-4},2^{-2},2^0,2^2\}$ \\
& BRDTI & $k=5$, $r \in \{50,100\}$ , $\lambda_r=\{0.01, 0.05, 0.1, 0.3\}$,   $\lambda_c=\{0.05, 0.1, 0.5, 0.9\}$ \\ \midrule
%& \multicolumn{2}{c}{\textbf{DTI Prediction Competitors}} \\ \hline
%& Method & Parameter  Settings \\ \hline
\multirow{6}{*}{\begin{tabular}[c]{@{}c@{}} \textbf{DTI} \\ \textbf{Prediction} \\ \textbf{Competitors} \end{tabular}} & MSCMF & $r \in \{50,100\}$, $\lambda_l, \lambda_d, \lambda_t \in   \{2^{-4},2^{-2},2^0,2^2\}$, $\lambda_w \in   \{2^1, 2^3, 2^5, 2^7, 2^9\}$ \\
& KRLSMKL & $\lambda \in \{2^{-15}, 2^{-10}, 2^{-5}, 2^{0}, 2^5, 2^{10},    2^{15}, 2^{20}, 2^{25}, 2^{30}\}$,   $\sigma \in \{0, 0.25, 0.5, 0.75, 1\}$ \\
& MKTCMF & $\lambda_{\Theta}, \lambda_l, \lambda_d, \lambda_t, \lambda_{\beta} \in \{0.5,1\}$, $r_d \in \{53, 200, 300 \}$, $r_t \in \{21, 80, 200, 300 \}$ \\ 
%, $iter \in \{2,5\}$ \\
% Optimal settings for each dataset listed in Table 3 of \BL{MKTCMF paper} \\
& NEDTP & $p=1$, $q=3$, $k=4$, $\#walk=8$, $walk\_length=80$, $r \in \{50,100\}$, $\#iter=3000$\\
& NeoDTI & learning rate = $10^{-4}$, $d \in \{256, 512, 1024\}$, $k \in \{256,   512, 1024\}$ \\
& DTIP & kernel size = $2 \times 2$, stride = 1, \#filters = 16 \& 32 \\
& DCFME & $\gamma=2$, $in_d=256$, $out_d=64$, $k=256$ \\
& SupDTI & {\begin{tabular}[c]{@{}c@{}} $\lambda_{l2} \in \{1,2,3,4\}{e-2}$, $\lambda_j \in \{0.05,0.1,0.5,1,1.5\}$, $\lambda_c \in \{0.1,0.5,1,1.5\}$, \\ $\lambda_{gen} \in \{0.1,0.2,0.5,1,1.5\}$, $\lambda_c \in \{0.05,0.1,0.2,0.5,1\}$  \end{tabular}}  \\ \bottomrule
\end{tabular}
\end{table*}
\clearpage

\section{Supplementary Results}

\subsection{Detailed Results for Comparison of FGS with Similarity Integration Methods}

\begin{table}[h]
\centering
\footnotesize
\caption{Detailed AUPR Results in CVS$_d$}
\label{Atab_AUPR_S2}
\begin{tabular}{@{}ccccccccc@{}}
\toprule
Base model & Dataset & AVE & KA & HSIC & LIC & SNF-H & SNF-F & FGS \\ \midrule
\multirow{5}{*}{WkNNRI} & NR & 0.417 & 0.426 & 0.426 & 0.48 & 0.38 & 0.467 & \textbf{0.549} \\
 & GPCR & 0.41 & 0.419 & 0.412 & 0.509 & 0.36 & 0.507 & \textbf{0.547} \\
 & IC & 0.336 & 0.338 & 0.346 & 0.45 & 0.28 & 0.411 & \textbf{0.486} \\
 & E & 0.259 & 0.265 & 0.262 & 0.376 & 0.18 & 0.358 & \textbf{0.418} \\
 & Luo & 0.485 & 0.462 & 0.486 & 0.488 & 0.27 & 0.446 & \textbf{0.49} \\ \midrule
\multirow{5}{*}{DLapRLS} & NR & 0.416 & 0.434 & 0.429 & 0.489 & 0.4 & 0.453 & \textbf{0.549} \\
 & GPCR & 0.441 & 0.448 & 0.441 & 0.521 & 0.33 & 0.495 & \textbf{0.527} \\
 & IC & 0.35 & 0.354 & 0.356 & 0.45 & 0.27 & 0.387 & \textbf{0.46} \\
 & E & 0.279 & 0.278 & 0.279 & 0.367 & 0.16 & 0.325 & \textbf{0.394} \\
 & Luo & \textbf{0.503} & 0.475 & \textbf{0.503} & 0.498 & 0.25 & 0.425 & 0.494 \\ \midrule
\multirow{5}{*}{GRMF} & NR & 0.422 & 0.438 & 0.431 & 0.493 & 0.38 & 0.465 & \textbf{0.551} \\
 & GPCR & 0.386 & 0.402 & 0.386 & 0.497 & 0.34 & 0.506 & \textbf{0.53} \\
 & IC & 0.334 & 0.336 & 0.34 & 0.453 & 0.28 & 0.41 & \textbf{0.481} \\
 & E & 0.242 & 0.253 & 0.243 & 0.368 & 0.17 & 0.343 & \textbf{0.409} \\
 & Luo & 0.48 & 0.453 & 0.48 & 0.481 & 0.27 & 0.426 & \textbf{0.482} \\ \midrule
\multirow{5}{*}{NRLMF} & NR & 0.382 & 0.393 & 0.388 & 0.449 & 0.37 & 0.472 & \textbf{0.517} \\
 & GPCR & 0.365 & 0.378 & 0.366 & 0.464 & 0.27 & \textbf{0.534} & 0.511 \\
 & IC & 0.303 & 0.296 & 0.31 & 0.428 & 0.22 & 0.437 & \textbf{0.47} \\
 & E & 0.235 & 0.24 & 0.237 & 0.335 & 0.13 & 0.378 & \textbf{0.403} \\
 & Luo & \textbf{0.459} & 0.442 & \textbf{0.459} & 0.456 & 0.3 & 0.437 & \textbf{0.459} \\ \midrule
\multirow{5}{*}{BRDTI} & NR & 0.398 & 0.405 & 0.4 & 0.457 & 0.42 & 0.487 & \textbf{0.523} \\
 & GPCR & 0.364 & 0.376 & 0.367 & 0.487 & 0.39 & 0.536 & \textbf{0.538} \\
 & IC & 0.298 & 0.294 & 0.304 & 0.413 & 0.3 & 0.439 & \textbf{0.461} \\
 & E & 0.227 & 0.231 & 0.228 & 0.298 & 0.21 & 0.348 & \textbf{0.352} \\
 & Luo & 0.446 & 0.432 & 0.446 & 0.446 & 0.29 & 0.42 & \textbf{0.45} \\  \midrule
\multicolumn{2}{c}{\textit{Average Rank}} & 5.3 & 4.7 & 4.4 & 2.52 & 6.68 & 3.2 & \textbf{1.2} \\ \bottomrule
\end{tabular}
\end{table}

\begin{table}[t]
\centering
\footnotesize
\caption{Detailed AUPR Results in CVS$_t$}
\label{Atab_AUPR_S3}
\begin{tabular}{@{}ccccccccc@{}}
\toprule
Base model & Dataset & AVE & KA & HSIC & LIC & SNF-H & SNF-F & FGS \\ \midrule
\multirow{5}{*}{WkNNRI} & NR & 0.538 & 0.534 & 0.537 & 0.549 & 0.56 & 0.496 & \textbf{0.565} \\
 & GPCR & 0.735 & 0.752 & 0.738 & 0.76 & 0.67 & 0.75 & \textbf{0.776} \\
 & IC & 0.852 & 0.854 & 0.852 & \textbf{0.856} & 0.83 & 0.818 & 0.855 \\
 & E & 0.71 & 0.724 & 0.711 & 0.724 & 0.62 & 0.703 & \textbf{0.725} \\
 & Luo & 0.216 & 0.14 & 0.216 & 0.293 & 0.08 & 0.384 & \textbf{0.459} \\ \midrule
\multirow{5}{*}{DLapRLS} & NR & 0.509 & 0.514 & 0.51 & 0.517 & \textbf{0.55} & 0.501 & 0.546 \\
 & GPCR & 0.725 & 0.742 & 0.729 & 0.747 & 0.63 & 0.733 & \textbf{0.754} \\
 & IC & 0.86 & \textbf{0.861} & 0.86 & 0.856 & 0.82 & 0.826 & 0.847 \\
 & E & 0.713 & \textbf{0.719} & 0.713 & 0.71 & 0.59 & 0.692 & 0.711 \\
 & Luo & 0.249 & 0.163 & 0.249 & 0.316 & 0.09 & 0.422 & \textbf{0.426} \\ \midrule
\multirow{5}{*}{GRMF} & NR & 0.522 & 0.518 & 0.515 & 0.523 & \textbf{0.57} & 0.505 & 0.563 \\
 & GPCR & 0.731 & 0.747 & 0.734 & 0.758 & 0.7 & \textbf{0.773} & 0.769 \\
 & IC & 0.858 & 0.859 & 0.858 & 0.859 & 0.86 & 0.846 & \textbf{0.863} \\
 & E & 0.704 & 0.714 & 0.704 & \textbf{0.718} & 0.64 & 0.702 & \textbf{0.718} \\
 & Luo & 0.248 & 0.168 & 0.248 & 0.319 & 0.1 & 0.42 & \textbf{0.448} \\ \midrule
\multirow{5}{*}{NRLMF} & NR & 0.421 & 0.434 & 0.426 & 0.447 & \textbf{0.55} & 0.502 & 0.486 \\
 & GPCR & 0.685 & 0.695 & 0.688 & 0.708 & 0.69 & \textbf{0.769} & 0.725 \\
 & IC & 0.83 & 0.832 & 0.832 & 0.835 & \textbf{0.86} & 0.834 & 0.845 \\
 & E & 0.67 & 0.684 & 0.671 & 0.692 & 0.66 & \textbf{0.717} & 0.693 \\
 & Luo & 0.206 & 0.153 & 0.206 & 0.262 & 0.04 & \textbf{0.4} & 0.354 \\ \midrule
\multirow{5}{*}{BRDTI} & NR & 0.42 & 0.435 & 0.425 & 0.442 & \textbf{0.52} & 0.494 & 0.449 \\
 & GPCR & 0.59 & 0.605 & 0.593 & 0.615 & 0.56 & \textbf{0.655} & 0.631 \\
 & IC & 0.748 & 0.75 & 0.749 & 0.753 & \textbf{0.78} & 0.746 & 0.764 \\
 & E & 0.655 & 0.669 & 0.656 & \textbf{0.674} & 0.61 & 0.661 & 0.672 \\
 & Luo & 0.185 & 0.141 & 0.186 & 0.251 & 0.08 & \textbf{0.38} & 0.36 \\ \midrule
\multicolumn{2}{c}{\textit{Average Rank}} & 5.26 & 4.14 & 4.8 & 2.9 & 5.04 & 3.92 & \textbf{1.94} \\ \bottomrule
\end{tabular}
\end{table}

\begin{table}[t]
\centering
\footnotesize
\caption{Detailed AUPR Results in CVS$_{dt}$}
\label{Atab_AUPR_S4}
\begin{tabular}{@{}ccccccccc@{}}
\toprule
Base model & Dataset & AVE & KA & HSIC & LIC & SNF-H & SNF-F & FGS \\ \midrule
 & NR & 0.241 & 0.241 & 0.243 & 0.247 & 0.2 & 0.245 & \textbf{0.251} \\
 & GPCR & 0.289 & 0.303 & 0.294 & 0.374 & 0.24 & 0.35 & \textbf{0.389} \\
 & IC & 0.226 & 0.233 & 0.235 & 0.324 & 0.18 & 0.243 & \textbf{0.339} \\
 & E & 0.125 & 0.135 & 0.125 & 0.207 & 0.07 & 0.201 & \textbf{0.241} \\
\multirow{-5}{*}{WkNNRI} & Luo & 0.132 & 0.079 & 0.132 & 0.192 & 0.02 & 0.186 & \textbf{0.203} \\ \midrule
 & NR & 0.19 & 0.188 & 0.189 & 0.212 & 0.18 & 0.22 & \textbf{0.251} \\
 & GPCR & 0.077 & 0.075 & 0.077 & 0.086 & 0.06 & 0.165 & \textbf{0.251} \\
 & IC & 0.074 & 0.072 & 0.075 & 0.08 & 0.06 & 0.091 & \textbf{0.217} \\
 & E & 0.019 & 0.023 & 0.019 & 0.044 & 0.02 & 0.049 & \textbf{0.124} \\
\multirow{-5}{*}{DLapRLS} & Luo & 0.068 & 0.039 & 0.068 & 0.103 & 0 & 0.121 & \textbf{0.147} \\ \midrule
 & NR & 0.236 & 0.246 & 0.237 & 0.25 & 0.19 & \textbf{0.264} & 0.239 \\
 & GPCR & 0.273 & 0.295 & 0.276 & 0.363 & 0.24 & 0.357 & \textbf{0.37} \\
 & IC & 0.222 & 0.238 & 0.227 & 0.329 & 0.17 & 0.271 & \textbf{0.341} \\
 & E & 0.11 & 0.116 & 0.11 & 0.201 & 0.07 & 0.196 & \textbf{0.229} \\
\multirow{-5}{*}{GRMF} & Luo & 0.101 & 0.055 & 0.101 & 0.164 & 0.02 & 0.157 & \textbf{0.174} \\ \midrule
 & NR & 0.2 & 0.194 & 0.2 & 0.206 & 0.15 & \textbf{0.234} & 0.217 \\
 & GPCR & 0.242 & 0.255 & 0.246 & 0.307 & 0.17 & \textbf{0.388} & 0.334 \\
 & IC & 0.195 & 0.199 & 0.198 & 0.29 & 0.15 & 0.307 & \textbf{0.327} \\
 & E & 0.115 & 0.113 & 0.118 & 0.179 & 0.04 & \textbf{0.223} & 0.209 \\
\multirow{-5}{*}{NRLMF} & Luo & 0.098 & 0.07 & 0.098 & 0.147 & 0 & \textbf{0.189} & 0.173 \\ \midrule
 & NR & 0.201 & 0.212 & 0.203 & 0.234 & 0.21 & \textbf{0.272} & 0.21 \\
 & GPCR & 0.206 & 0.221 & 0.213 & 0.279 & 0.19 & 0.314 & \textbf{0.316} \\
 & IC & 0.194 & 0.198 & 0.193 & 0.279 & 0.18 & 0.236 & \textbf{0.316} \\
 & E & 0.115 & 0.114 & 0.117 & 0.179 & 0.05 & 0.152 & \textbf{0.203} \\
\multirow{-5}{*}{BRDTI} & Luo & 0.09 & 0.061 & 0.09 & 0.144 & 0.02 & 0.124 & \textbf{0.167} \\ \midrule
\multicolumn{2}{c}{\textit{Average Rank}} & 5.3 & 4.9 & 4.8 & 2.44 & 6.92 & 2.24 & \textbf{1.4} \\ \bottomrule
\end{tabular}
\end{table}

\begin{table}[t]
\centering
\footnotesize
\caption{Detailed AUC Results in CVS$_d$}
\label{Atab_AUC_S2}
\begin{tabular}{@{}ccccccccc@{}}
\toprule
Base model & Dataset & AVE & KA & HSIC & LIC & SNF-H & SNF-F & FGS \\ \midrule
\multirow{5}{*}{WkNNRI} & NR & 0.78 & 0.784 & 0.781 & 0.795 & 0.689 & 0.799 & \textbf{0.812} \\
 & GPCR & 0.863 & 0.872 & 0.866 & 0.903 & 0.792 & 0.909 & \textbf{0.911} \\
 & IC & 0.766 & 0.766 & 0.773 & 0.798 & 0.68 & 0.807 & \textbf{0.818} \\
 & E & 0.825 & 0.832 & 0.826 & 0.858 & 0.682 & 0.855 & \textbf{0.875} \\
 & Luo & 0.904 & 0.892 & 0.904 & 0.901 & 0.855 & 0.901 & \textbf{0.905} \\ \midrule
\multirow{5}{*}{DLapRLS} & NR & 0.779 & 0.787 & 0.782 & 0.802 & 0.734 & 0.797 & \textbf{0.811} \\
 & GPCR & 0.862 & 0.863 & 0.862 & \textbf{0.886} & 0.773 & 0.872 & 0.881 \\
 & IC & 0.771 & 0.769 & 0.772 & \textbf{0.804} & 0.669 & 0.779 & 0.789 \\
 & E & 0.773 & 0.766 & 0.773 & 0.794 & 0.666 & \textbf{0.81} & 0.803 \\
 & Luo & 0.865 & 0.863 & 0.865 & 0.865 & 0.791 & \textbf{0.882} & 0.873 \\ \midrule
\multirow{5}{*}{GRMF} & NR & 0.792 & 0.799 & 0.79 & 0.806 & 0.695 & 0.801 & \textbf{0.825} \\
 & GPCR & 0.828 & 0.834 & 0.829 & 0.887 & 0.788 & 0.89 & \textbf{0.904} \\
 & IC & 0.739 & 0.74 & 0.742 & 0.792 & 0.691 & 0.792 & \textbf{0.802} \\
 & E & 0.775 & 0.79 & 0.775 & 0.815 & 0.665 & 0.805 & \textbf{0.825} \\
 & Luo & 0.871 & 0.869 & 0.871 & 0.871 & 0.791 & 0.852 & \textbf{0.876} \\ \midrule
\multirow{5}{*}{NRLMF} & NR & 0.773 & 0.779 & 0.774 & 0.794 & 0.722 & 0.811 & \textbf{0.821} \\
 & GPCR & 0.848 & 0.854 & 0.85 & 0.893 & 0.78 & 0.903 & \textbf{0.909} \\
 & IC & 0.763 & 0.757 & 0.764 & 0.803 & 0.683 & 0.808 & \textbf{0.815} \\
 & E & 0.81 & 0.821 & 0.81 & 0.85 & 0.667 & 0.836 & \textbf{0.868} \\
 & Luo & \textbf{0.915} & 0.911 & \textbf{0.915} & 0.913 & 0.89 & 0.901 & 0.914 \\ \midrule
\multirow{5}{*}{BRDTI} & NR & 0.764 & 0.769 & 0.765 & 0.783 & 0.745 & 0.819 & \textbf{0.823} \\
 & GPCR & 0.86 & 0.864 & 0.862 & 0.894 & 0.859 & 0.907 & \textbf{0.908} \\
 & IC & 0.761 & 0.758 & 0.763 & 0.794 & 0.764 & \textbf{0.816} & 0.809 \\
 & E & 0.831 & 0.844 & 0.831 & 0.867 & 0.823 & 0.878 & \textbf{0.881} \\
 & Luo & \textbf{0.881} & 0.88 & \textbf{0.881} & 0.879 & 0.832 & 0.866 & 0.876 \\ \midrule
\multicolumn{2}{c}{\textit{Average Rank}} & 4.98 & 4.65 & 4.48 & 2.73 & 6.85 & 2.78 & \textbf{1.55} \\ \bottomrule
\end{tabular}
\end{table}

\begin{table}[t]
\centering
\footnotesize
\caption{Detailed AUC Results in CVS$_t$}
\label{Atab_AUC_S3}
\begin{tabular}{@{}ccccccccc@{}}
\toprule
Base model & Dataset & AVE & KA & HSIC & LIC & SNF-H & SNF-F & FGS \\ \midrule
\multirow{5}{*}{WkNNRI} & NR & 0.787 & 0.787 & 0.787 & 0.792 & 0.798 & 0.809 & \textbf{0.817} \\
 & GPCR & 0.937 & 0.939 & 0.938 & 0.945 & 0.924 & 0.948 & \textbf{0.95} \\
 & IC & 0.957 & 0.956 & 0.957 & \textbf{0.959} & 0.952 & 0.956 & 0.955 \\
 & E & 0.935 & 0.934 & \textbf{0.936} & 0.935 & 0.912 & 0.932 & 0.933 \\
 & Luo & 0.816 & 0.767 & 0.816 & 0.857 & 0.692 & \textbf{0.886} & 0.862 \\ \midrule
\multirow{5}{*}{DLapRLS} & NR & 0.781 & 0.783 & 0.781 & 0.784 & 0.791 & 0.781 & \textbf{0.815} \\
 & GPCR & 0.919 & 0.922 & 0.92 & 0.919 & 0.887 & 0.916 & \textbf{0.932} \\
 & IC & \textbf{0.95} & \textbf{0.95} & \textbf{0.95} & 0.946 & 0.938 & 0.942 & 0.943 \\
 & E & 0.894 & 0.892 & 0.895 & 0.885 & 0.864 & \textbf{0.899} & 0.897 \\
 & Luo & 0.8 & 0.761 & 0.8 & 0.827 & 0.625 & 0.845 & \textbf{0.848} \\ \midrule
\multirow{5}{*}{GRMF} & NR & 0.794 & 0.795 & 0.794 & 0.796 & 0.793 & 0.804 & \textbf{0.814} \\
 & GPCR & 0.941 & 0.942 & 0.942 & 0.949 & 0.932 & \textbf{0.954} & 0.953 \\
 & IC & 0.959 & 0.958 & 0.959 & \textbf{0.96} & 0.953 & 0.956 & \textbf{0.958} \\
 & E & \textbf{0.923} & 0.921 & \textbf{0.923} & 0.921 & 0.888 & 0.912 & 0.922 \\
 & Luo & 0.776 & 0.749 & 0.777 & 0.844 & 0.646 & 0.851 & \textbf{0.859} \\ \midrule
\multirow{5}{*}{NRLMF} & NR & 0.77 & 0.775 & 0.771 & 0.786 & \textbf{0.818} & 0.801 & 0.809 \\
 & GPCR & 0.934 & 0.936 & 0.935 & 0.942 & 0.918 & \textbf{0.946} & 0.943 \\
 & IC & \textbf{0.959} & 0.957 & \textbf{0.959} & \textbf{0.959} & \textbf{0.959} & 0.958 & 0.958 \\
 & E & \textbf{0.942} & 0.941 & \textbf{0.942} & 0.939 & 0.911 & 0.925 & 0.938 \\
 & Luo & 0.818 & 0.798 & 0.818 & 0.874 & 0.632 & \textbf{0.902} & 0.892 \\ \midrule
\multirow{5}{*}{BRDTI} & NR & 0.747 & 0.75 & 0.746 & 0.757 & 0.773 & 0.774 & \textbf{0.787} \\
 & GPCR & 0.924 & 0.928 & 0.926 & 0.932 & 0.913 & 0.936 & \textbf{0.937} \\
 & IC & 0.949 & 0.951 & 0.949 & 0.952 & \textbf{0.953} & 0.951 & 0.951 \\
 & E & \textbf{0.923} & \textbf{0.923} & \textbf{0.923} & \textbf{0.923} & 0.907 & 0.913 & 0.922 \\
 & Luo & 0.75 & 0.714 & 0.75 & 0.833 & 0.638 & \textbf{0.868} & 0.852 \\ \midrule
\multicolumn{2}{c}{\textit{Average Rank}} & 4.43 & 4.45 & 4.08 & 3.3 & 5.73 & 3.48 & \textbf{2.55} \\ \bottomrule
\end{tabular}
\end{table}

\begin{table}[t]
\centering
\footnotesize
\caption{Detailed AUC Results in CVS$_{dt}$}
\label{Atab_AUC_S4}
\begin{tabular}{@{}ccccccccc@{}}
\toprule
Base model & Dataset & AVE & KA & HSIC & LIC & SNF-H & SNF-F & FGS \\ \midrule
\multirow{5}{*}{WkNNRI} & NR & 0.599 & 0.596 & 0.598 & 0.598 & 0.577 & \textbf{0.608} & 0.603 \\
 & GPCR & 0.816 & 0.827 & 0.818 & 0.857 & 0.74 & \textbf{0.872} & 0.869 \\
 & IC & 0.722 & 0.724 & 0.725 & 0.764 & 0.637 & 0.757 & \textbf{0.768} \\
 & E & 0.768 & 0.773 & 0.768 & 0.8 & 0.615 & 0.789 & \textbf{0.821} \\
 & Luo & 0.819 & 0.778 & 0.819 & 0.859 & 0.662 & \textbf{0.864} & 0.862 \\ \midrule
\multirow{5}{*}{DLapRLS} & NR & 0.588 & 0.584 & 0.588 & 0.587 & 0.558 & 0.586 & \textbf{0.633} \\
 & GPCR & 0.443 & 0.448 & 0.445 & 0.453 & 0.536 & 0.65 & \textbf{0.698} \\
 & IC & 0.467 & 0.463 & 0.471 & 0.483 & 0.511 & 0.563 & \textbf{0.661} \\
 & E & 0.342 & 0.353 & 0.342 & 0.396 & 0.526 & \textbf{0.629} & 0.59 \\
 & Luo & 0.454 & 0.386 & 0.454 & 0.536 & 0.468 & \textbf{0.666} & 0.657 \\ \midrule
\multirow{5}{*}{GRMF} & NR & 0.621 & 0.621 & 0.622 & 0.621 & 0.578 & \textbf{0.638} & 0.63 \\
 & GPCR & 0.789 & 0.802 & 0.79 & 0.843 & 0.724 & 0.821 & \textbf{0.851} \\
 & IC & 0.693 & 0.702 & 0.694 & \textbf{0.753} & 0.666 & 0.748 & 0.749 \\
 & E & 0.721 & 0.73 & 0.721 & 0.761 & 0.6 & 0.737 & \textbf{0.774} \\
 & Luo & 0.8 & 0.763 & 0.801 & 0.837 & 0.65 & 0.824 & \textbf{0.838} \\ \midrule
\multirow{5}{*}{NRLMF} & NR & 0.589 & 0.589 & 0.59 & 0.594 & 0.558 & 0.605 & \textbf{0.606} \\
 & GPCR & 0.802 & 0.813 & 0.808 & 0.844 & 0.714 & \textbf{0.87} & 0.865 \\
 & IC & 0.71 & 0.713 & 0.71 & 0.756 & 0.639 & 0.761 & \textbf{0.764} \\
 & E & 0.775 & 0.776 & 0.775 & 0.803 & 0.602 & 0.771 & \textbf{0.814} \\
 & Luo & 0.796 & 0.763 & 0.796 & 0.835 & 0.58 & \textbf{0.849} & 0.842 \\ \midrule
\multirow{5}{*}{BRDTI} & NR & 0.602 & 0.609 & 0.6 & 0.622 & 0.568 & \textbf{0.663} & 0.643 \\
 & GPCR & 0.773 & 0.788 & 0.783 & 0.83 & 0.692 & 0.808 & \textbf{0.856} \\
 & IC & 0.655 & 0.664 & 0.657 & 0.728 & 0.642 & \textbf{0.73} & 0.726 \\
 & E & 0.747 & 0.755 & 0.748 & 0.778 & 0.64 & 0.739 & \textbf{0.783} \\
 & Luo & 0.711 & 0.642 & 0.711 & 0.783 & 0.559 & 0.732 & \textbf{0.786} \\ \midrule
\multicolumn{2}{c}{\textit{Average Rank}} & 5.38 & 4.83 & 4.8 & 2.8 & 6.25 & 2.5 & \textbf{1.45} \\ \bottomrule
\end{tabular}
\end{table}
\clearpage

\subsection{Influence of Base Model Hyperparameter Variation}

We use the optimal hyperparameter values of base models via grid research to obtain the results listed in Tables \ref{Atab_AUPR_S2}-\ref{Atab_AUC_S4}. However, the optimal hyperparameter value is not always attainable due to computational or time limitation, especially for extremely complex base model and enormous amount of hyperparameter candidates. Therefore, we investigate that if FGS still outperforms other similarity integration methods when the base model uses non-optimal (arbitrary) hyperparameter settings.

\begin{figure}[!h]
\centering
\subfloat[$k=2$]{\includegraphics[width=0.45\textwidth]{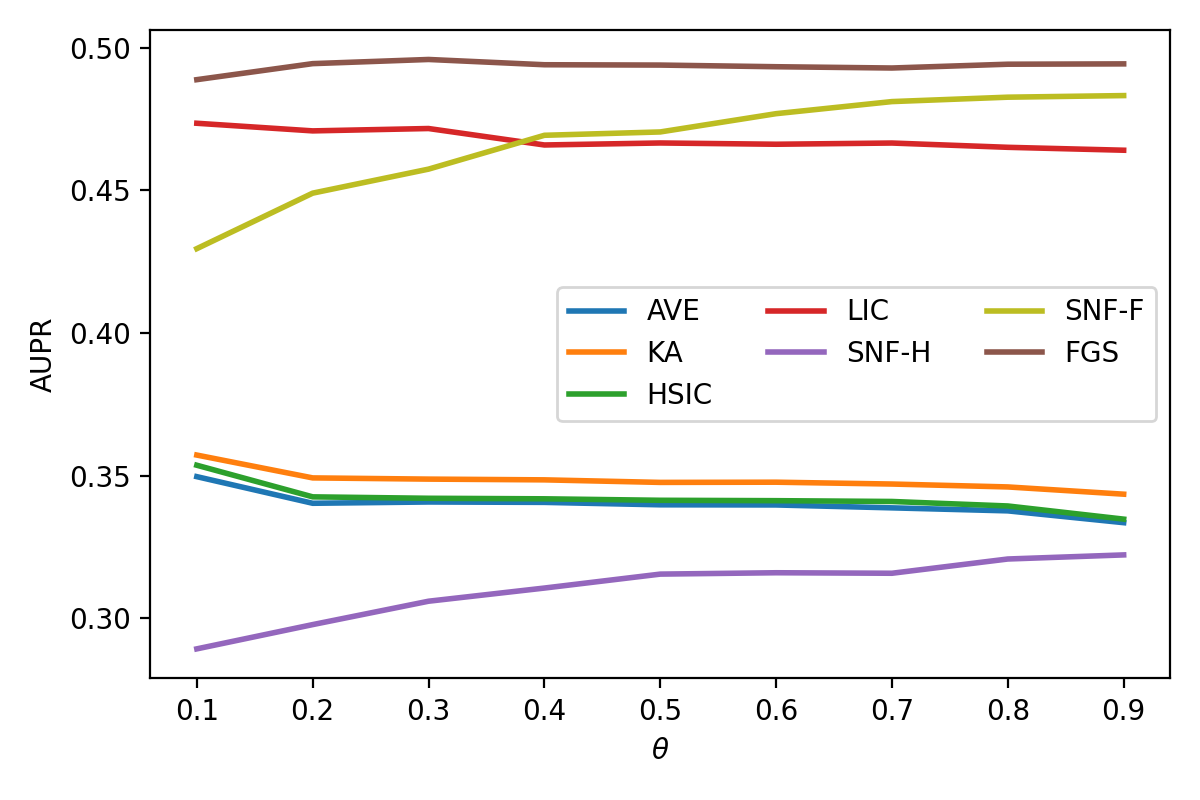}}
\subfloat[$k=3$]{\includegraphics[width=0.45\textwidth]{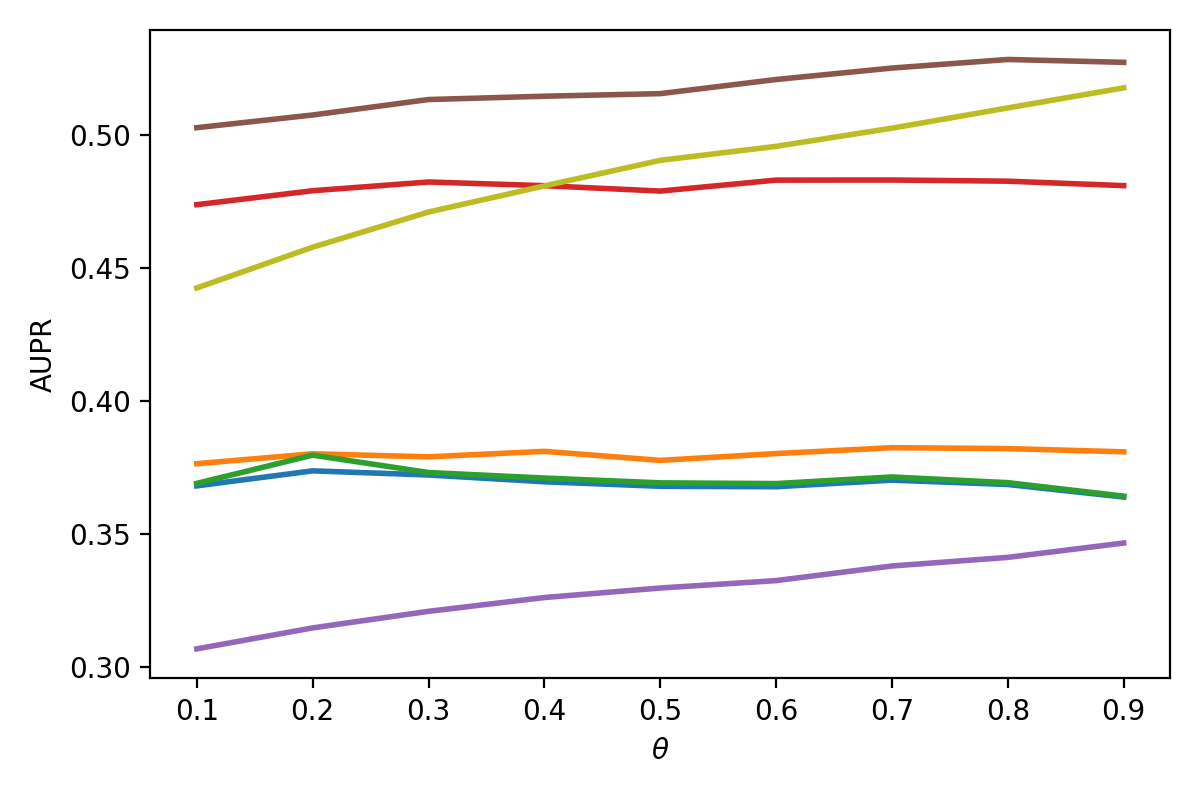}}  \\
\subfloat[$k=5$]{\includegraphics[width=0.45\textwidth]{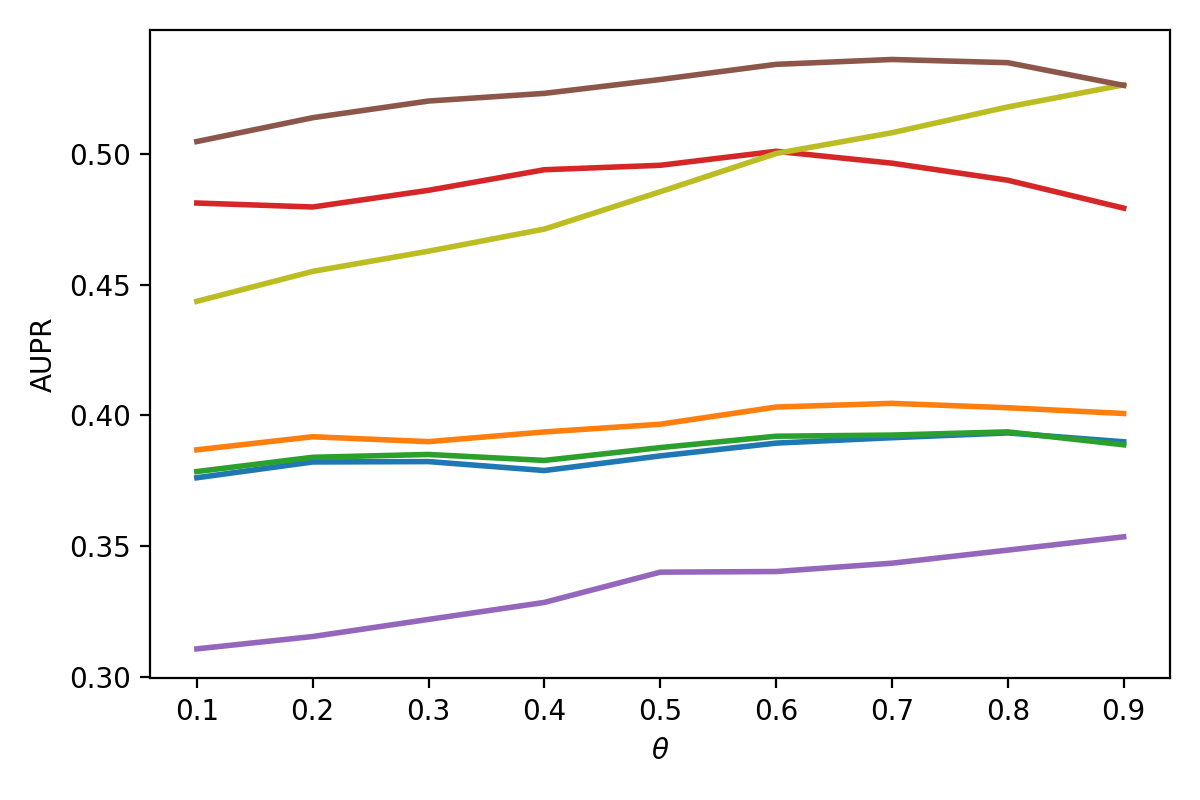}}
\subfloat[$k=7$]{\includegraphics[width=0.45\textwidth]{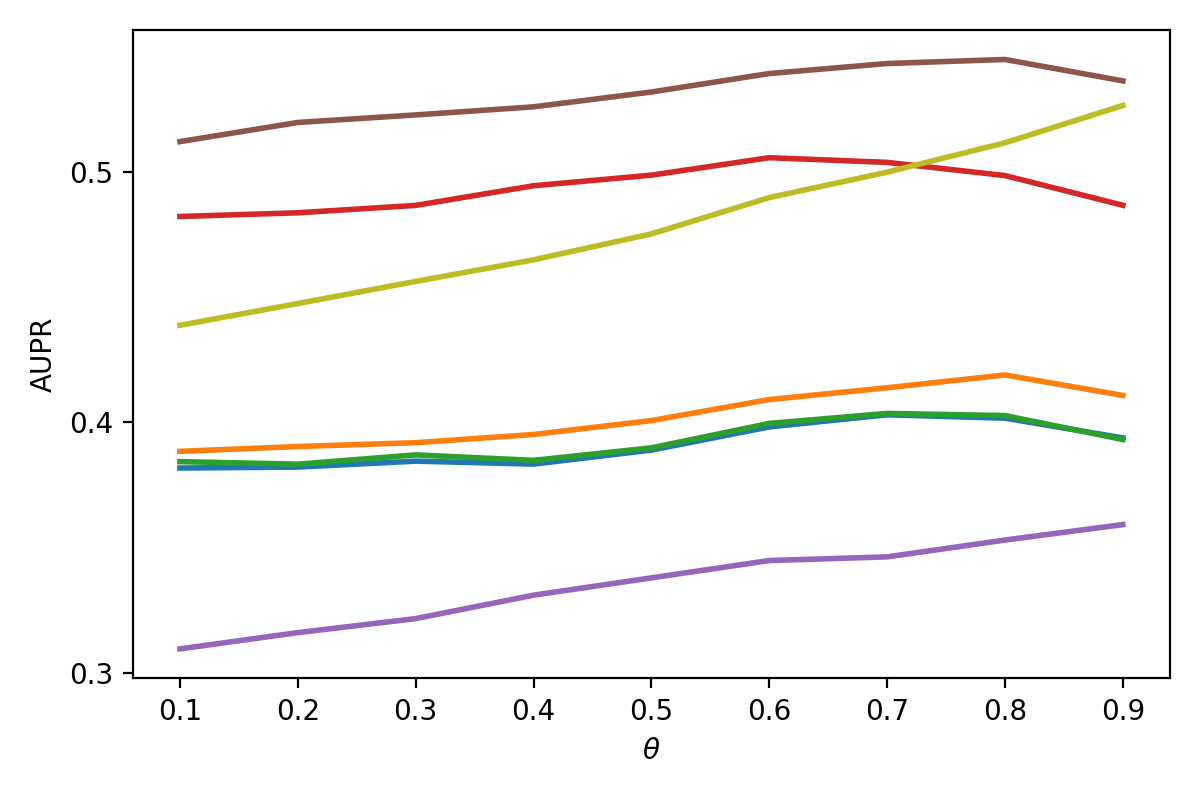}}  \\
\subfloat[$k=9$]{\includegraphics[width=0.45\textwidth]{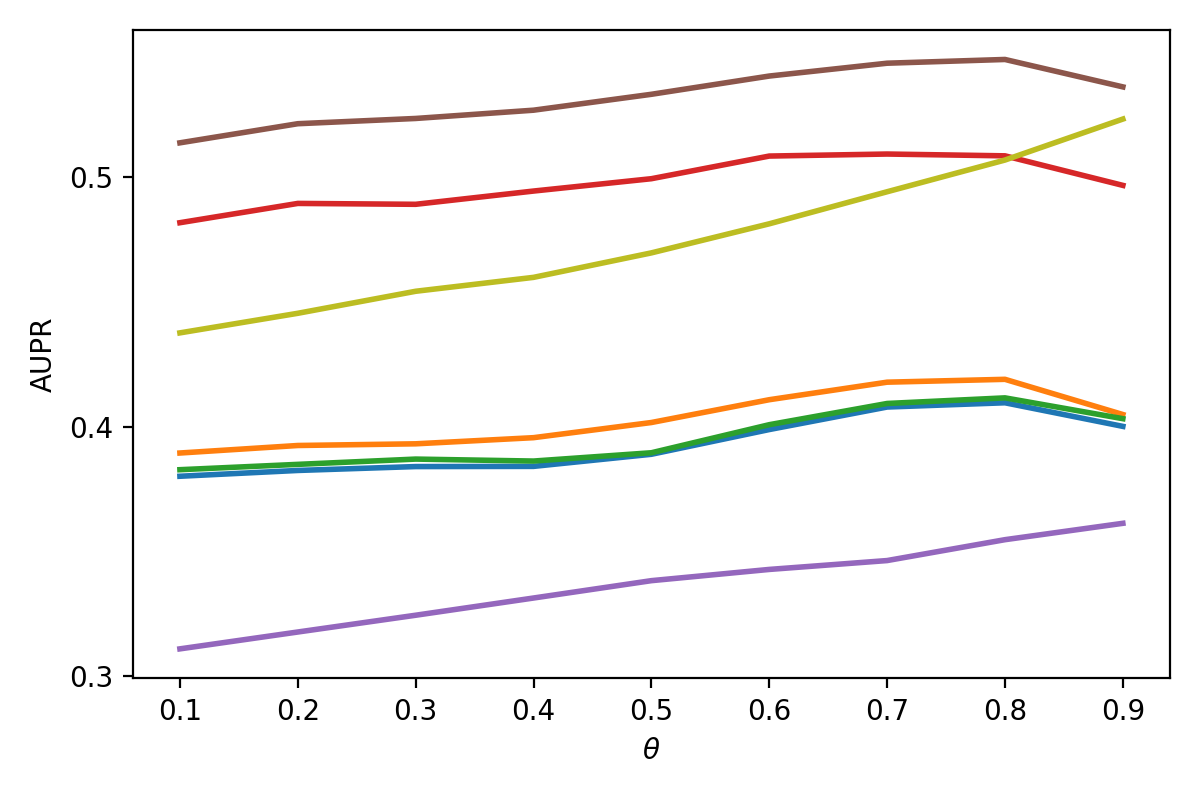}}
\caption{AUPR results of similarity integration methods with different hyperparameter settings of WkNNIR under CVS$_d$}
\label{fig:Var_Param_Base_Model}
\end{figure}

Figure \ref{fig:Var_Param_Base_Model} shows the AUPR results of seven similarity integration methods with WkNNIR using various hyperparameter settings on the GPCR dataset under CVS$_d$. As we can see, FGS is always the best method under all hyperparameter settings, which demonstrates the robustness of FGS with respect to the hyperparameter variance of the base model.

\subsection{Results of Cluster Cross-Validation}

% \BL{\textbf{Comments 2.7}: What are the similarities between drugs in the training sets and testing sets? How to avoid over fitting? An independent test is encouraged.}

\begin{figure}[!h]
\centering
\subfloat[NR]{\includegraphics[width=0.32\textwidth]{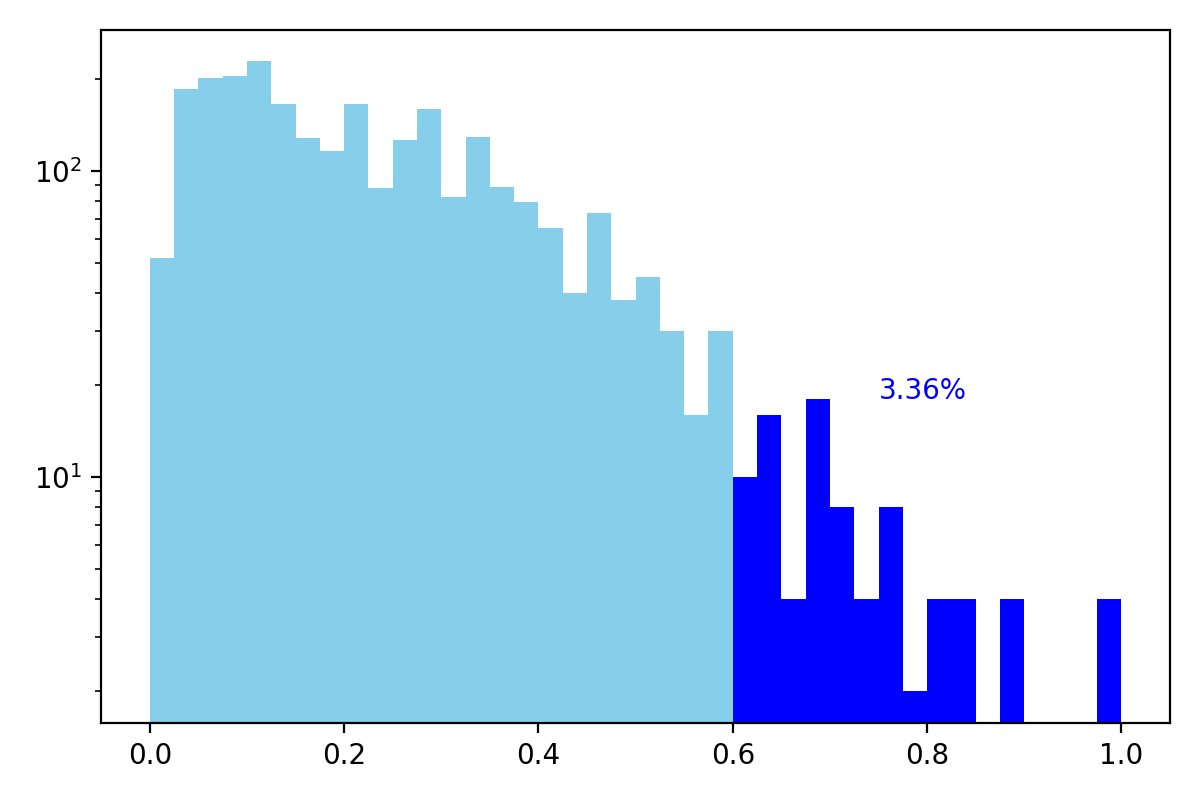}}
\subfloat[GPCR]{\includegraphics[width=0.32\textwidth]{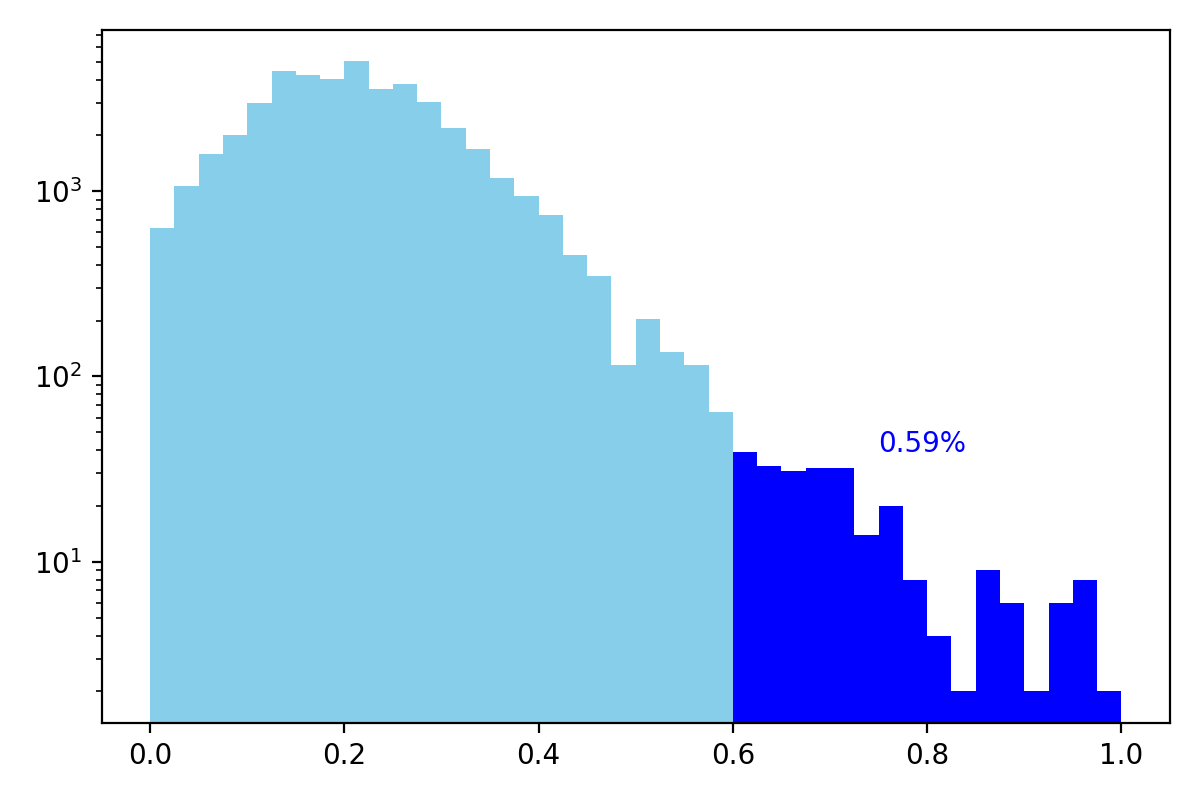}} 
\subfloat[IC]{\includegraphics[width=0.32\textwidth]{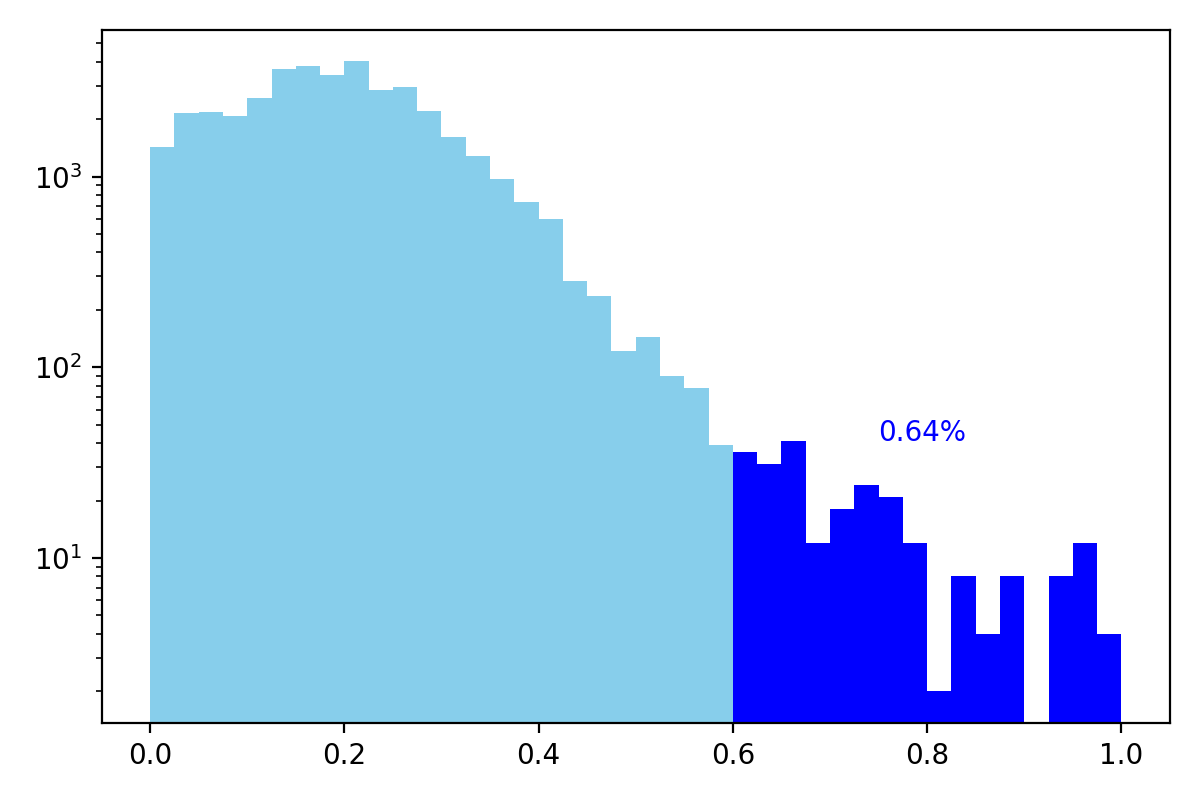}} \\
\subfloat[E]{\includegraphics[width=0.32\textwidth]{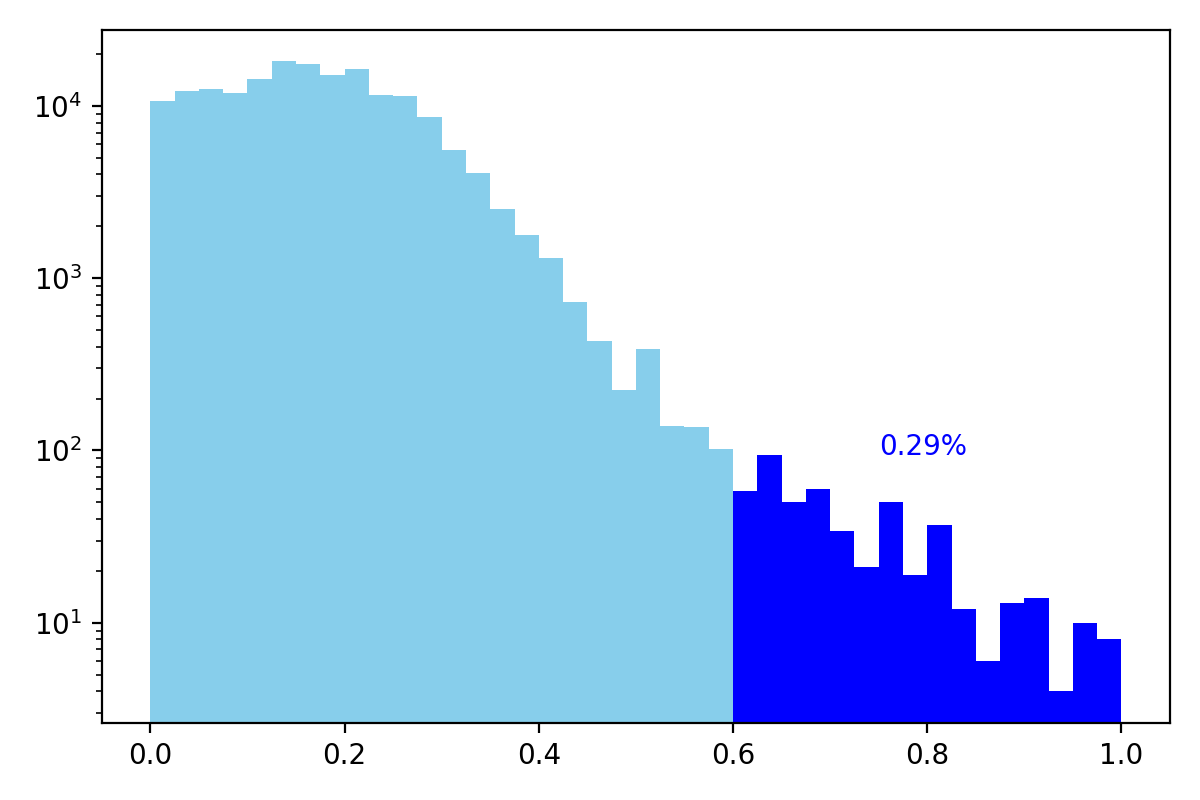}}
\subfloat[Luo]{\includegraphics[width=0.32\textwidth]{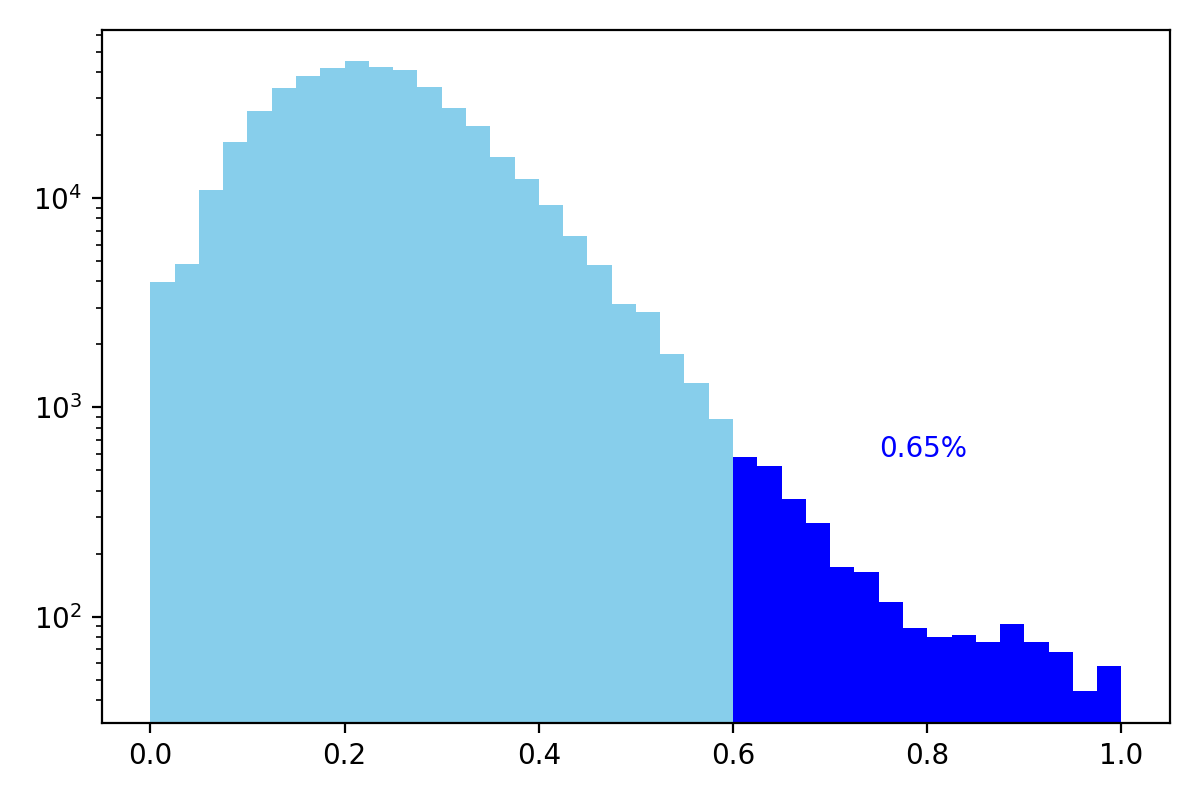}} \\
%\hspace{1em}% Space between image A and B
\caption{The distribution of SIMCOMP similarities between training and test drugs under  CVS$_d$, where the dark blue percentage is the proportion of similarity values of homologous drugs (larger than 0.6)}
\label{fig:SimDistri_CV}
\end{figure}

The datasets used in our empirical study contain some \textit{homologous drugs} that are structurally similar and interact with same targets. Inferring DTIs based on homologous drugs is considered as an easy prediction. Figure \ref{fig:SimDistri_CV} shows SIMCOMP similarities between training and test drugs under CVS$_d$, where the training and test drugs whose similarity value is larger than 0.6 are homologous ones~\cite{Luo2017AInformation,Wan2019NeoDTI:Interactions}. Since CVS$_d$ splits training and test drugs randomly, there are some homologous drugs in both training and test sets. 
Therefore, it is necessary to examine if FGS still perform well if there is no easy prediction, i.e., test drugs are not similar to training ones.

\begin{figure}[!h]
\centering
\subfloat[NR]{\includegraphics[width=0.32\textwidth]{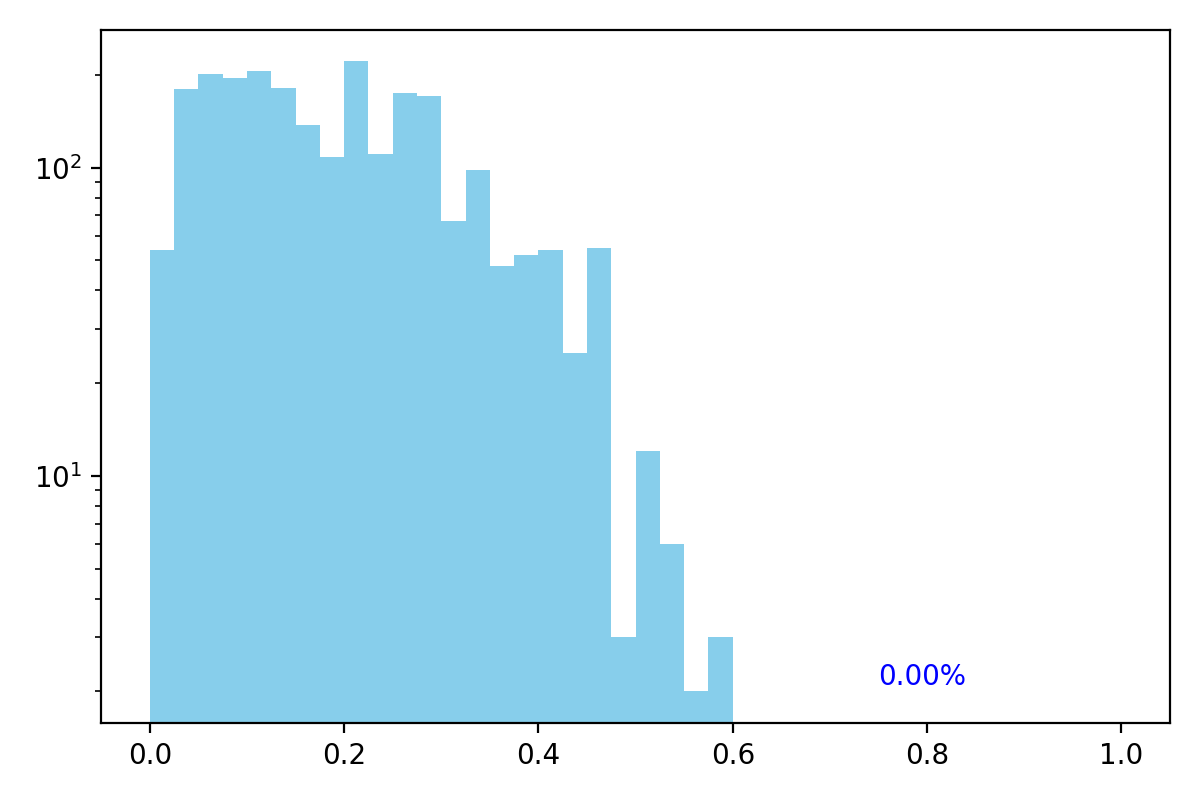}}
\subfloat[GPCR]{\includegraphics[width=0.32\textwidth]{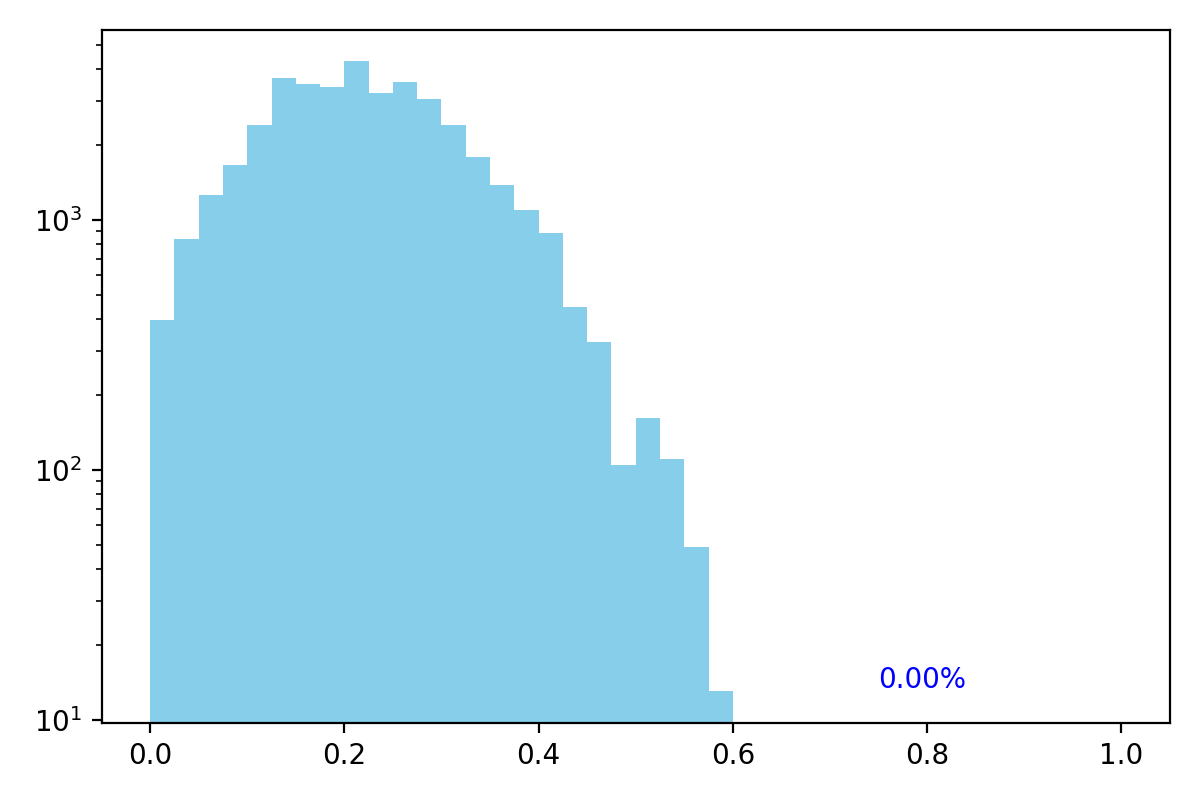}} 
\subfloat[IC]{\includegraphics[width=0.32\textwidth]{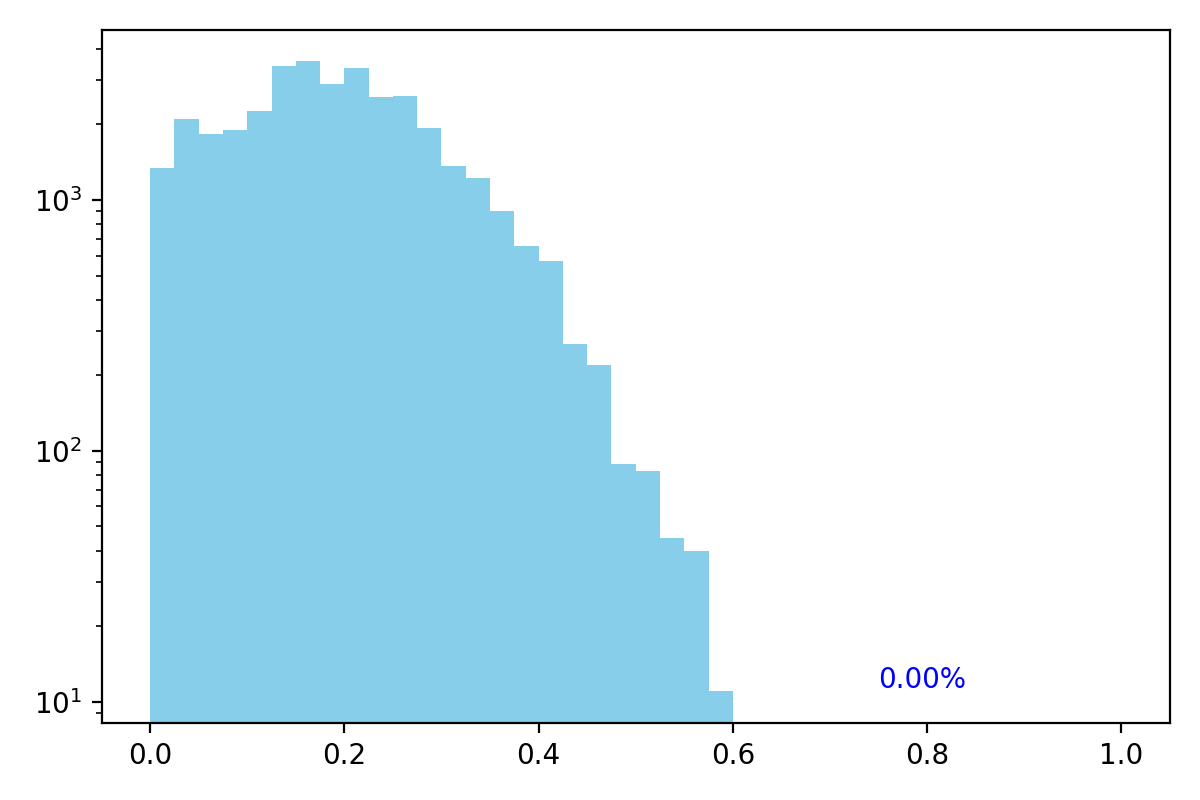}} \\
\subfloat[E]{\includegraphics[width=0.32\textwidth]{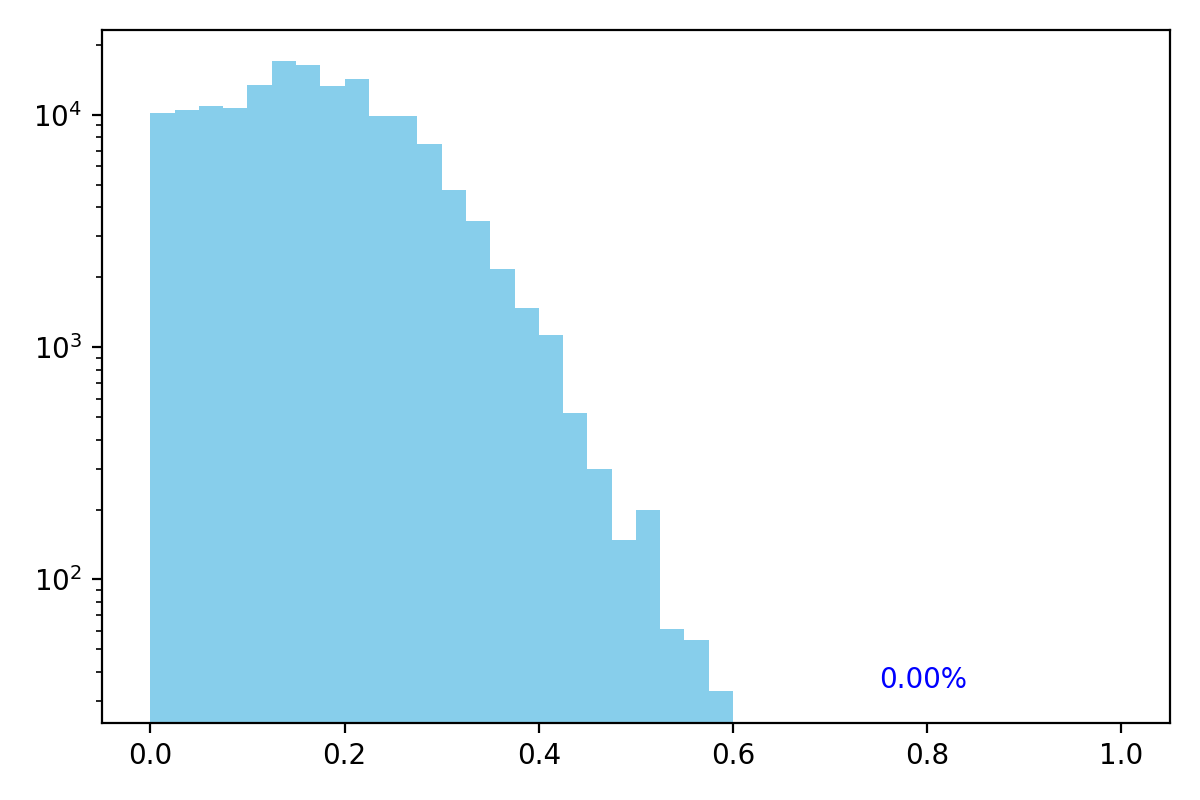}}
\subfloat[Luo]{\includegraphics[width=0.32\textwidth]{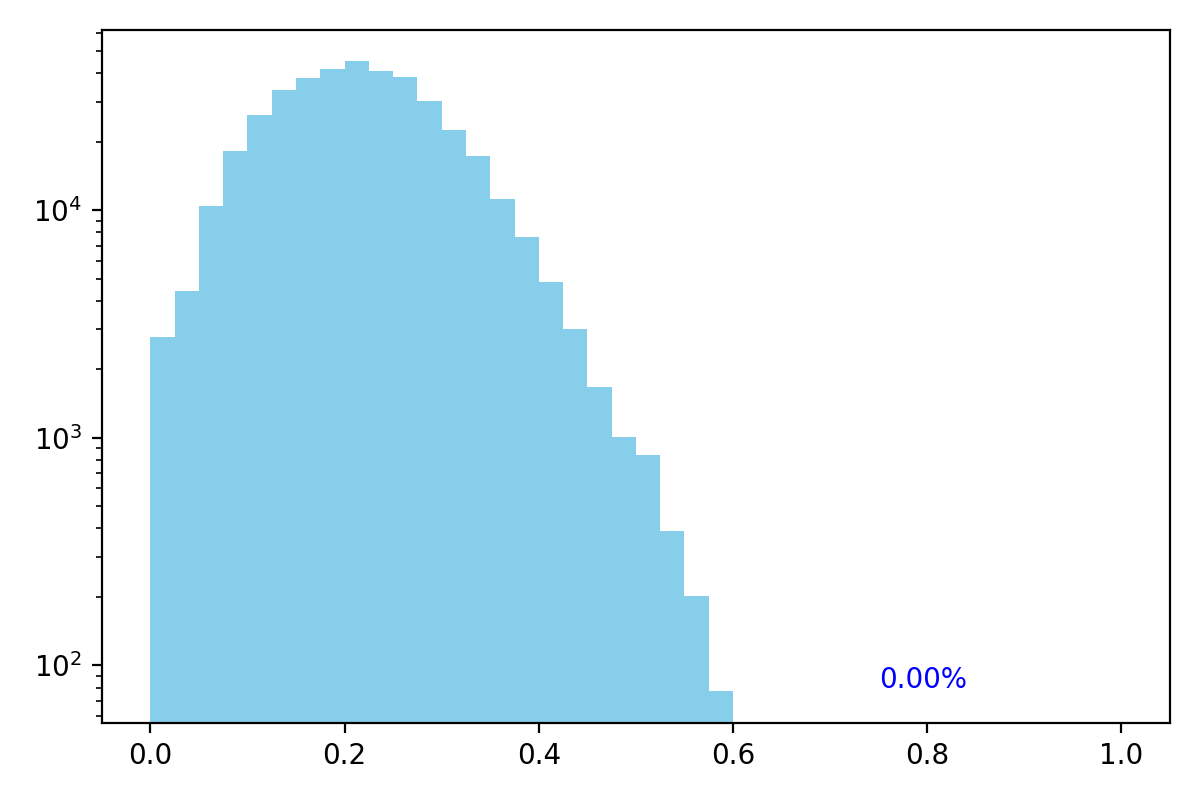}} \\
%\hspace{1em}% Space between image A and B
\caption{The distribution of SIMCOMP similarities between training and test drugs under  cluCVS$_d$, where the dark blue percentage is the proportion of similarity values of homologous drugs (larger than 0.6)}
\label{fig:SimDistri_clusterCV}
\end{figure}

To avoid homologous drugs being allocated to both training and test sets, we employ cluster cross-validation for drugs (cluCVS$_d$)~\cite{Mayr2018_Large_ChEMBL}. We first apply single-linkage clustering to all drugs, with SIMCOMP similarity as the input and a threshold of 0.6 for merging clusters. For obtained drug clusters, the \textit{single-linkage similarity} between any two drug clusters is less than 0.6, where the single-linkage similarity is defined as the most similar drugs in the two clusters. Then, we randomly allocate drug clusters in training and test sets, and conduct the allocation ten times. 
As shown in Figure~\ref{fig:SimDistri_clusterCV}, SIMCOMP similarities between all training and test drugs are less than 0.6 under cluCVS$_d$, which indicates the non-existence of easy predictions.

\begin{figure}[!h]
\centering
\subfloat[NR]{\includegraphics[width=0.32\textwidth]{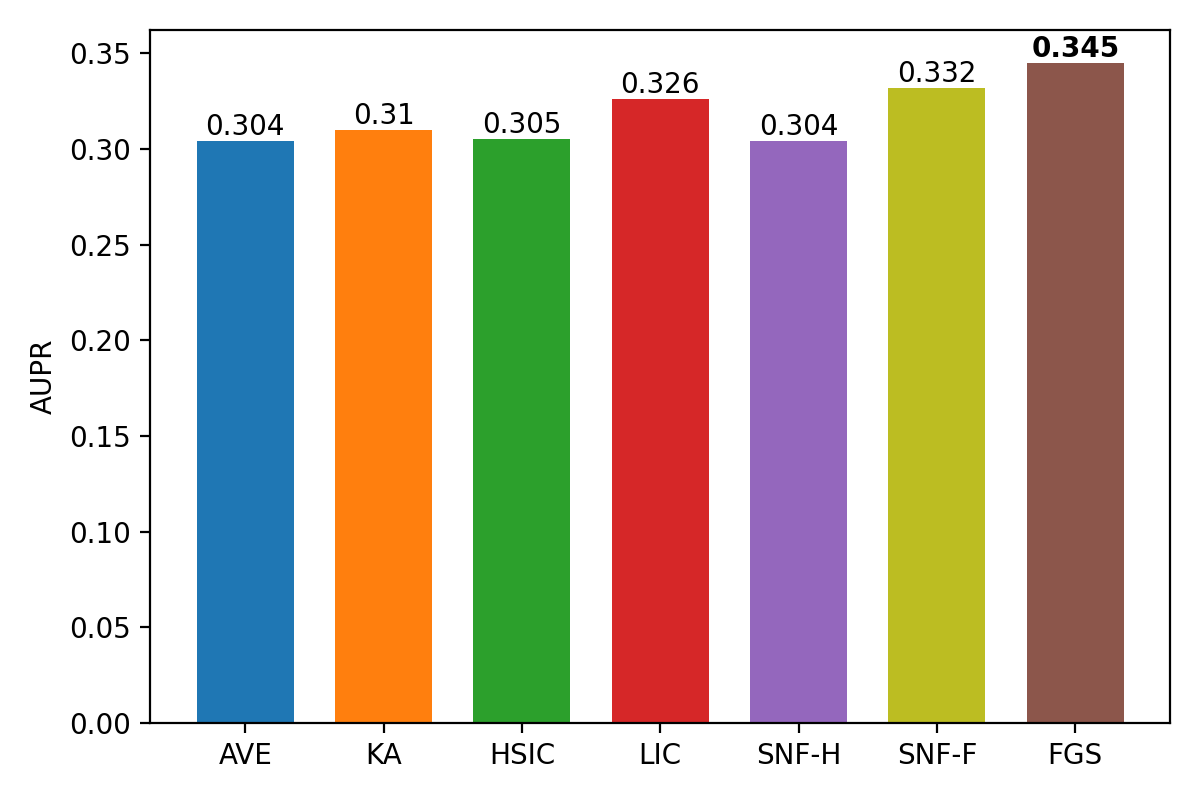}}
\subfloat[GPCR]{\includegraphics[width=0.32\textwidth]{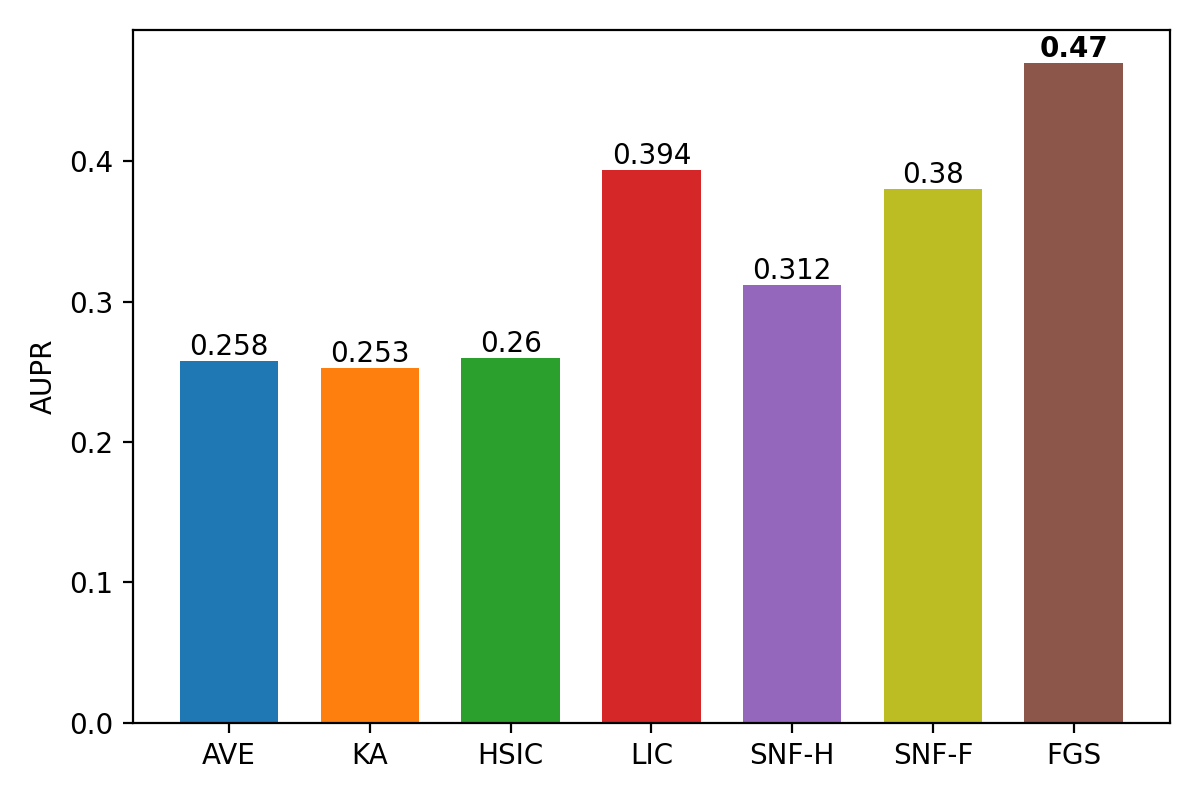}} 
\subfloat[IC]{\includegraphics[width=0.32\textwidth]{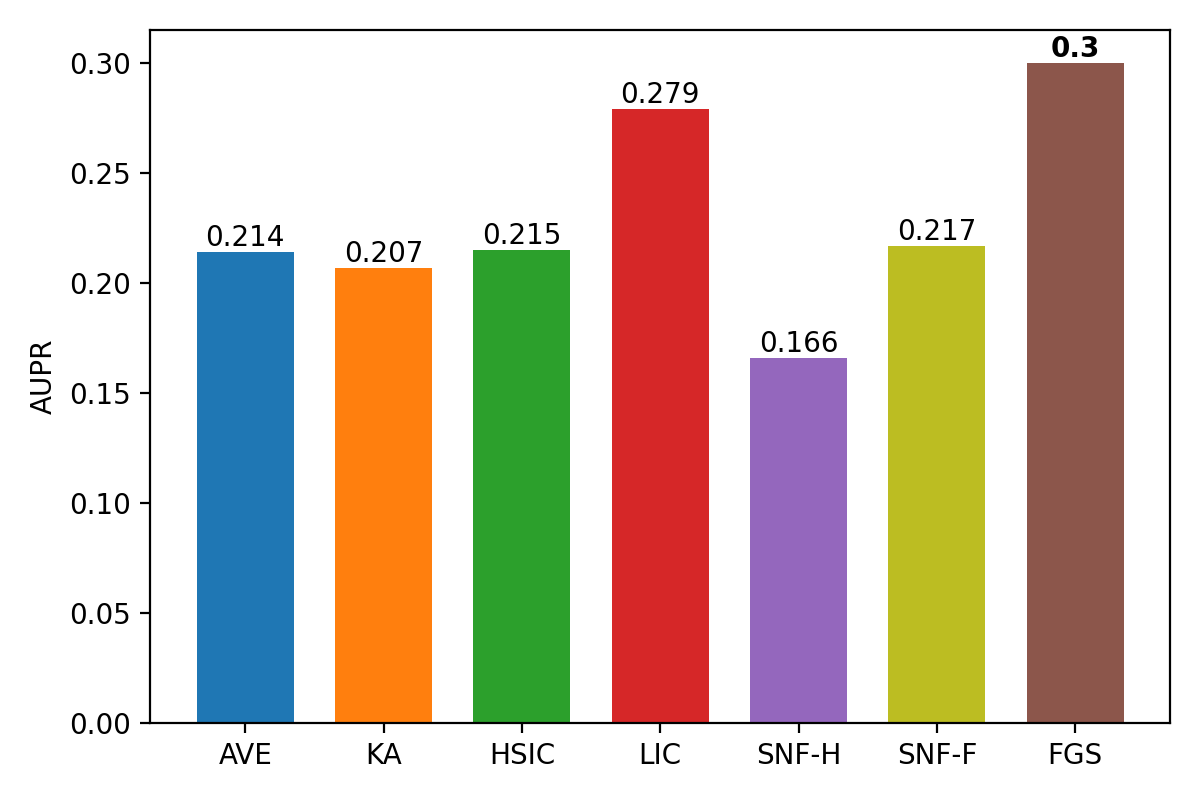}} \\
\subfloat[E]{\includegraphics[width=0.32\textwidth]{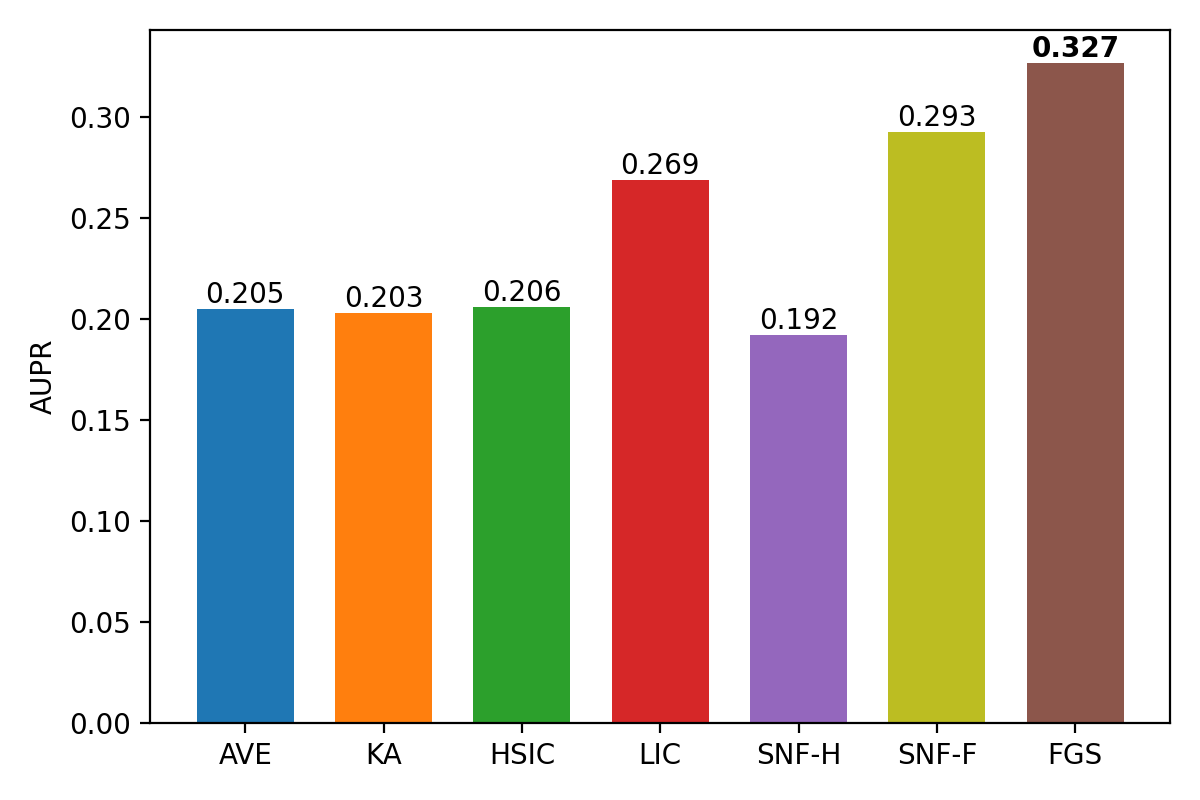}}
\subfloat[Luo]{\includegraphics[width=0.32\textwidth]{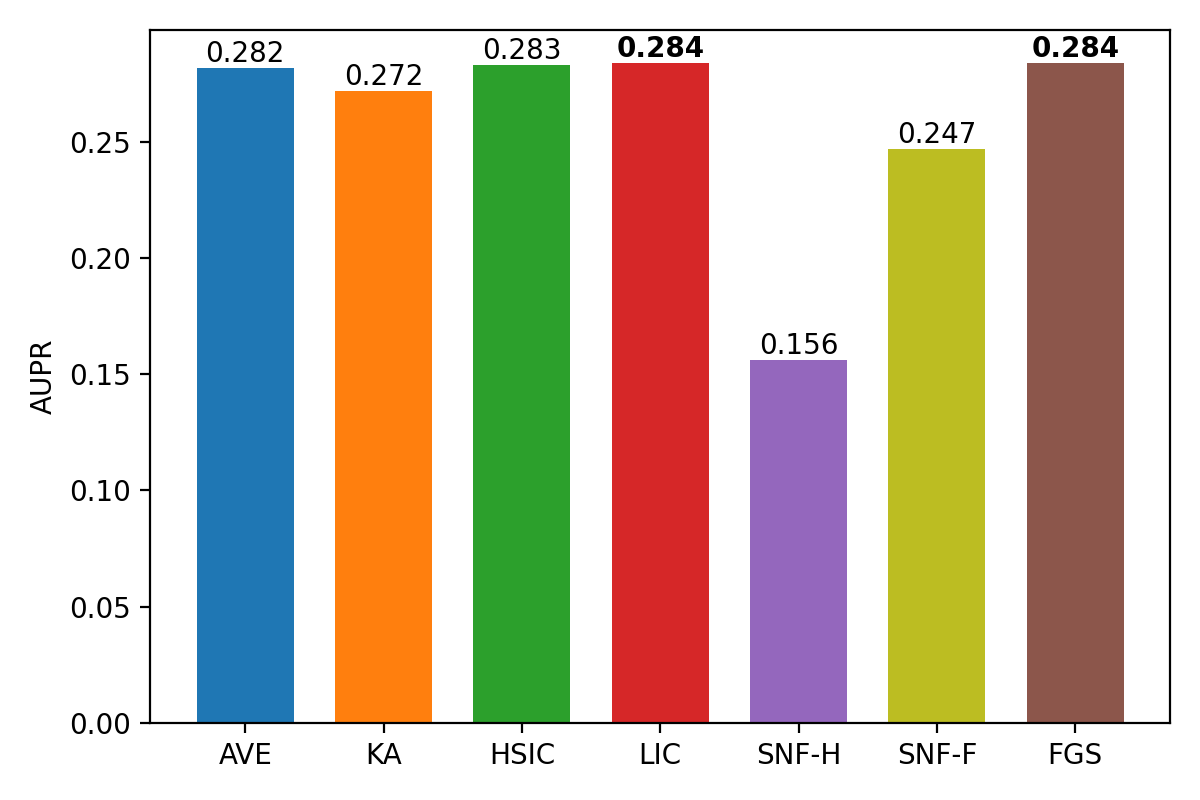}} \\
%\hspace{1em}% Space between image A and B
\caption{AUPR results with WkNNIR under cluCVS$_d$}
\label{fig:clusterCV_result}
\end{figure}

We further compare AUPR results of FGS and seven comparing similarity integration methods using WkNNIR as the base model under cluCVS$_d$. As shown in Figure \ref{fig:clusterCV_result}, FGS is still the best method for all datasets, demonstrating the robustness of FGS after excluding easy predictions. In addition, there is a drop in AUPR results for all methods, because the removal of easy predictions leads to a more difficult task.

\clearpage

\subsection{Hyperparameter Sensitivity Analysis}
\begin{figure}[!h]
\centering
\subfloat[$\rho$]{\includegraphics[width=0.48\textwidth]{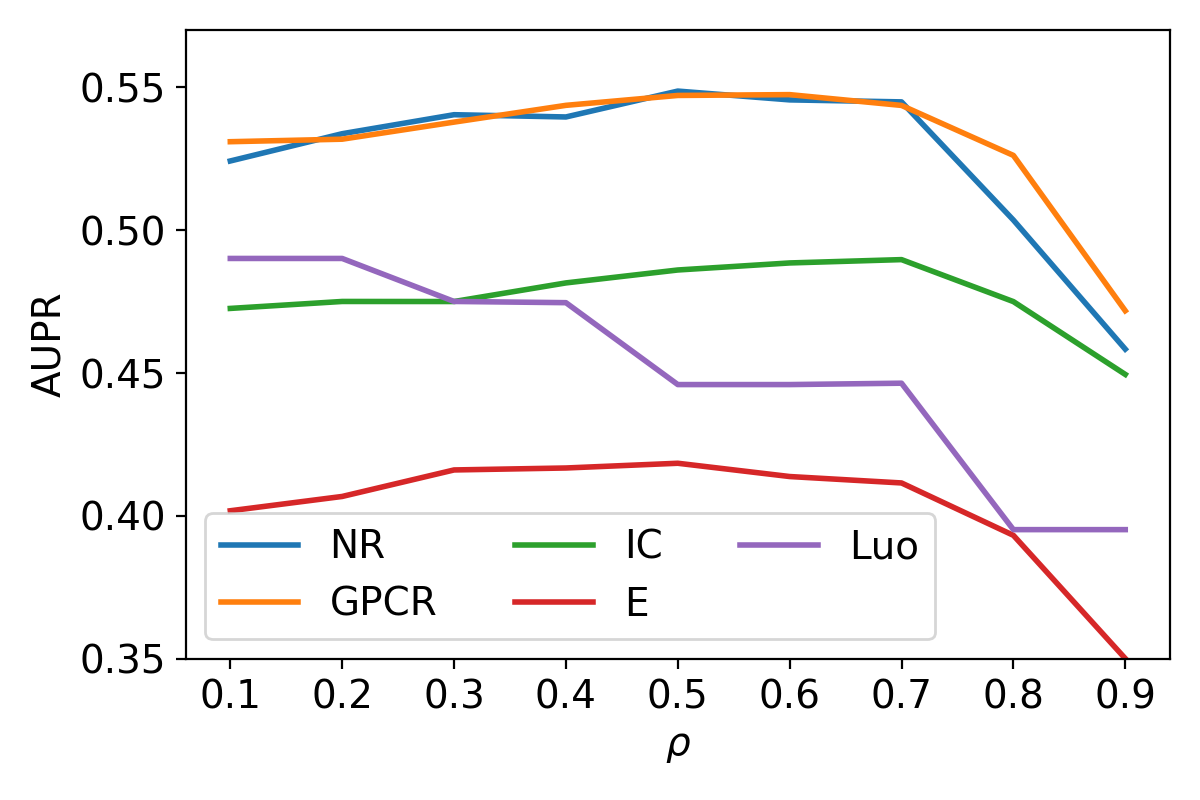}}
%\hspace{1em}% Space between image A and B
\subfloat[$k$]{\includegraphics[width=0.48\textwidth]{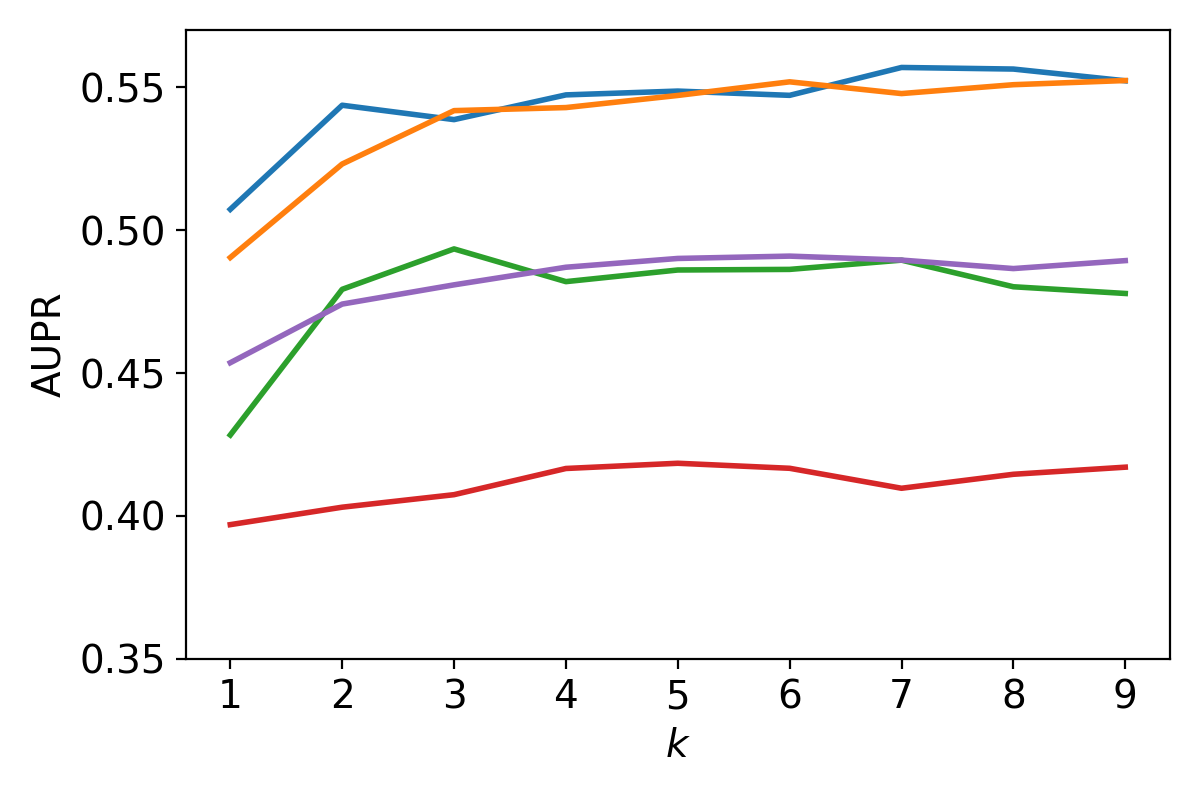}}
\caption{AUPR results of FGS with WkNNIR as base model (FGS$_{\text{WkNNIR}}$) under different hyperparameter configurations under in CVS$_d$.}
\label{fig:Var_Params}
\end{figure}

In this part, we analyze the sensitivity of two hyperparameters, i.e., the similarity filter ratio $\rho$ and the number of neighbors $k$. Figure \ref{fig:Var_Params} shows the AUPR performance of FGS with WkNNIR as base model (FGS$_{\text{WkNNIR}}$) w.r.t. different settings of $k$ and $\rho$ under CVS$_d$. 

We first focus on similarity filter ratio $\rho$.
For the four updated Golden Standard datasets containing nine types of drug and target similarities, the performance of FGS$_{\text{WkNNIR}}$ improves with the increase of $\rho$, as more noisy information is filtered. The FGS$_{\text{WkNNIR}}$ usually achieves the best performance when $\rho$ is around 0.6, but it becomes worse if $\rho$ is excessively large, which indicates the removal of more crucial and informative similarities.
For the Luo dataset, lower $\rho$ values lead to better performance. This is because the Luo dataset includes fewer types of drug and target similarities, all of which are critical to the DTI prediction task. 

The performance of FGS$_{\text{WkNNIR}}$ is more insensitive w.r.t. the changing of $k$. For all five datasets, the performance slightly improves with more neighbors considered, and reaches a plateau when $k$ is large enough.

\clearpage

\subsection{Supplementary Tables and Figures for New DTI Discovery}

\begin{table}[h]
\centering
\footnotesize
\caption{New DTIs predicted by other similarity integration methods combined with GRMF, where DB and DC denotes DrugBank and DrugCentral, respectively}
\label{Stab:newDTI_GRMF}
\begin{tabular}{@{}c|ccc|ccc|ccc@{}}
\toprule
\multirow{2}{*}{Rank} & \multicolumn{3}{c|}{AVE} & \multicolumn{3}{c|}{KA} & \multicolumn{3}{c}{HISC} \\ 
 & Drug ID & Target ID & Database & Drug ID & Target ID & Database & Drug ID & Target ID & Database \\ \midrule
1 & DB01215 & P48169 & DB & DB01215 & P48169 & DB & DB01215 & P48169 & DB \\
2 & DB00829 & P48169 & DB & DB00829 & P48169 & DB & DB00829 & P48169 & DB \\
3 & DB00734 & P28222 & DC & DB00734 & P28222 & DC & DB00734 & P28222 & DC \\
4 & DB00652 & P41143 & DC & DB00696 & P08908 & DB,DC & DB00652 & P41143 & DC \\
5 & DB06800 & P41143 & DC & DB00953 & P08908 & DC & DB06800 & P41143 & DC \\
6 & DB00696 & P08908 & DB,DC & DB06216 & P28221 & DC & DB00696 & P08908 & DB,DC \\
7 & DB00953 & P08908 & DC & DB00652 & P41143 & DC & DB00953 & P08908 & DC \\
8 & DB00734 & P21918 & DC & DB00918 & P08908 & - & DB00734 & P21918 & DC \\
9 & DB00185 & P08172 & DC & DB00734 & P21918 & DC & DB00185 & P08172 & DC \\
10 & DB00335 & P07550 & DB,DC & DB06800 & P41143 & DC & DB00335 & P07550 & DB,DC \\
\midrule
\multirow{2}{*}{Rank} & \multicolumn{3}{c|}{LIC} & \multicolumn{3}{c|}{SNF-H} & \multicolumn{3}{c}{SNF-F} \\
 & Drug ID & Target ID & Database & Drug ID & Target ID & Database & Drug ID & Target ID & Database \\\midrule
1 & DB00829 & P48169 & DB & DB01215 & P48169 & DB & DB01215 & P48169 & DB \\
2 & DB01215 & P48169 & DB & DB00734 & P28222 & DC & DB00829 & P48169 & DB \\
3 & DB00652 & P41143 & DC & DB00953 & P08908 & DC & DB00652 & P41143 & DC \\
4 & DB00734 & P28222 & DC & DB00829 & P48169 & DB & DB06800 & P41143 & DC \\
5 & DB06800 & P41143 & DC & DB01273 & Q15822 & - & DB00734 & P28222 & DC \\
6 & DB00363 & P21918 & DC & DB00696 & P08908 & DB,DC & DB00810 & P08172 & DC \\
7 & DB00696 & P08908 & DB,DC & DB00998 & P08908 & DC & DB00809 & P08912 & DC \\
8 & DB00734 & P21918 & DC & DB06216 & P28221 & DC & DB00734 & P21918 & DC \\
9 & DB00335 & P07550 & DB,DC & DB00333 & P41145 & DC & DB00333 & P41145 & DC \\
10 & DB06216 & P28221 & DC & DB00918 & P08908 & - & DB00802 & P41145 & - \\ \bottomrule
\end{tabular}
\end{table}

\begin{table}[h]
\centering
\footnotesize
\caption{New DTIs predicted by other similarity integration methods combined with NRLMF, where DB and DC denotes DrugBank and DrugCentral, respectively}
\label{Stab:newDTI_NRLMF}
\begin{tabular}{@{}c|ccc|ccc|ccc@{}}
\toprule
\multirow{2}{*}{Rank} & \multicolumn{3}{c|}{AVE} & \multicolumn{3}{c|}{KA} & \multicolumn{3}{c}{HISC} \\ 
 & Drug ID & Target ID & Database & Drug ID & Target ID & Database & Drug ID & Target ID & Database \\ \midrule
1 & DB01215 & P48169 & DB & DB00696 & P08908 & DB,DC & DB01215 & P48169 & DB \\
2 & DB00829 & P48169 & DB & DB00363 & P21918 & DC & DB00829 & P48169 & DB \\
3 & DB00696 & P08908 & DB,DC & DB01215 & P48169 & DB & DB00696 & P08908 & DB,DC \\
4 & DB00363 & P21918 & DC & DB00829 & P48169 & DB & DB00363 & P21918 & DC \\
5 & DB00652 & P41143 & DC & DB00652 & P41143 & DC & DB00652 & P41143 & DC \\
6 & DB06216 & P28221 & DC & DB00454 & Q8TCU5 & - & DB06216 & P28221 & DC \\
7 & DB00335 & P07550 & DB,DC & DB00335 & P07550 & DB,DC & DB00335 & P07550 & DB,DC \\
8 & DB00543 & P28223 & DB,DC & DB06216 & P28221 & DC & DB00543 & P28223 & DB,DC \\
9 & DB00454 & Q8TCU5 & - & DB00543 & P28223 & DB,DC & DB00454 & Q8TCU5 & - \\
10 & DB00831 & P21728 & - & DB00799 & P19793 & - & DB00831 & P21728 & - \\
\midrule
\multirow{2}{*}{Rank} & \multicolumn{3}{c|}{LIC} & \multicolumn{3}{c|}{SNF-H} & \multicolumn{3}{c}{SNF-F} \\
 & Drug ID & Target ID & Database & Drug ID & Target ID & Database & Drug ID & Target ID & Database \\\midrule
1 & DB00363 & P21918 & DC & DB00363 & P21918 & DC & DB00363 & P21918 & DC \\
2 & DB01215 & P48169 & DB & DB00696 & P08908 & DB,DC & DB00185 & P08172 & DC \\
3 & DB00829 & P48169 & DB & DB06216 & P28221 & DC & DB00696 & P08908 & DB,DC \\
4 & DB00696 & P08908 & DB,DC & DB00734 & P28222 & DC & DB06216 & P28221 & DC \\
5 & DB00454 & Q8TCU5 & - & DB00185 & P08172 & DC & DB00829 & P48169 & DB \\
6 & DB00652 & P41143 & DC & DB00333 & P41145 & DC & DB01215 & P48169 & DB \\
7 & DB06216 & P28221 & DC & DB00799 & P19793 & - & DB01337 & P11229 & - \\
8 & DB00543 & P28223 & DB,DC & DB00829 & P48169 & DB & DB00333 & P41145 & DC \\
9 & DB00335 & P07550 & DB,DC & DB01215 & P48169 & DB & DB06216 & P21918 & DC \\
10 & DB00734 & P28222 & DC & DB01337 & P11229 & - & DB00734 & P28222 & DC \\ \bottomrule
\end{tabular}
\end{table}

\begin{table}[h]
\centering
\small
\caption{The number of verified DTIs in top 10 new DTIs discovered by different similarity integration methods with GRMF and NRLMF from the Luo dataset}
\label{tab:new_DTIs_numbers}
\begin{tabular}{@{}cccccccc@{}}
\toprule
Base Model & AVE & KA & HSIC & LIC         & SNF-H & SNF-F & FGS         \\ \midrule
GRMF       & 10  & 9  & 10   & 10          & 8     & 9     & 10          \\
NRLMF      & 8   & 8  & 8    & 9           & 8     & 9     & 8           \\ \midrule
%DLapRLS    & 8   & 6  & 8    & 8           & 1     & 7     & 9           \\ \midrule
%\textit{Sum}        & 26  & 23 & 26   & \textbf{27} & 17    & 25    & \textbf{27} \\ \bottomrule
\textit{Sum}        & 18  & 17 & 18   & 19 & 16  & 18  & 18 \\ \bottomrule
\end{tabular}
\end{table}

\begin{figure*}[h]
\centering
\subfloat[DB01273 (varenicline)-Q15822 (CHRNA2)]{\includegraphics[width=0.4\textwidth]{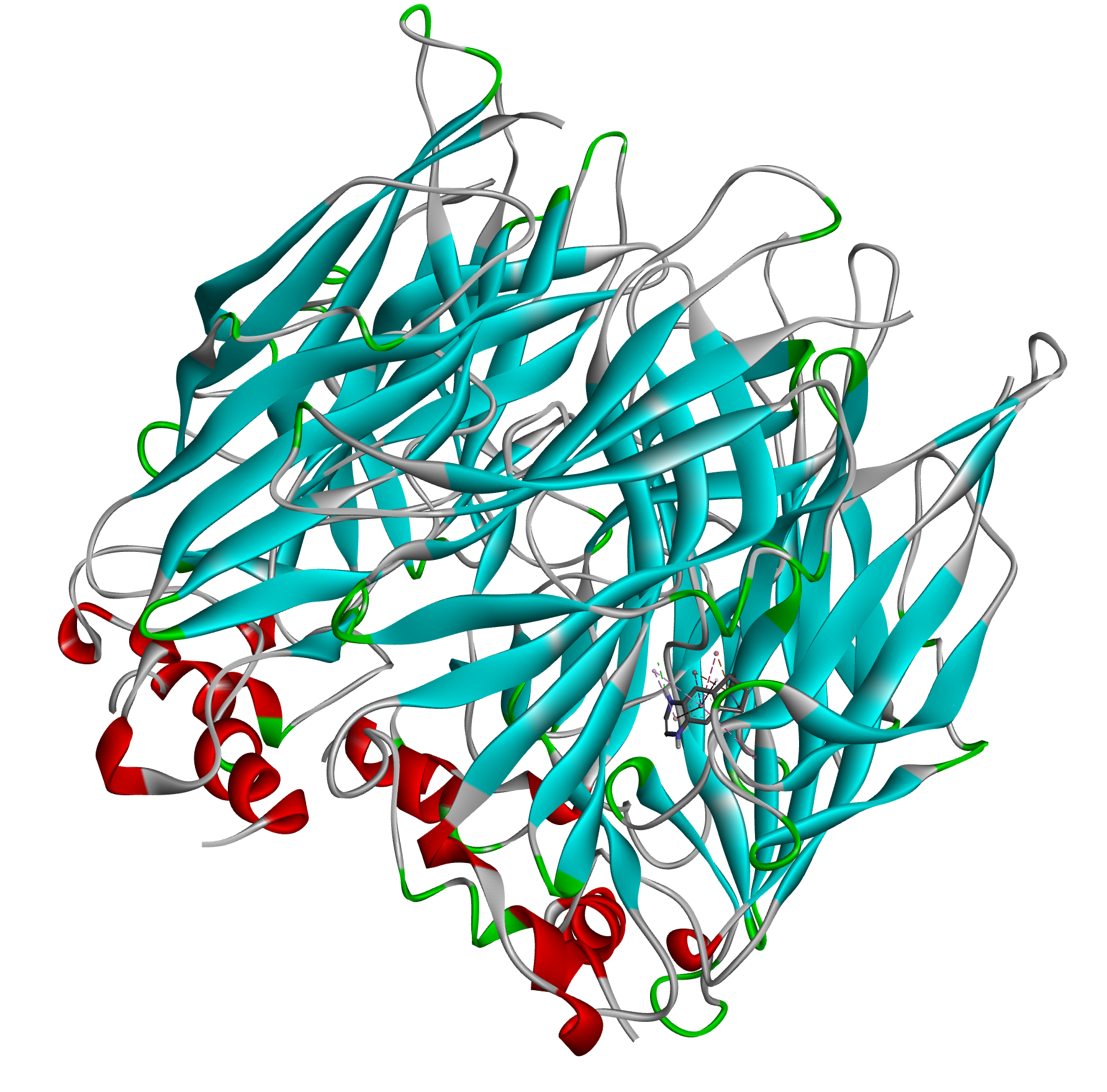}} 
%\hspace{1em}% Space between image A and B
\subfloat[DB01186 (pergolide)-P34969 (HTR7)]{\includegraphics[width=0.4\textwidth]{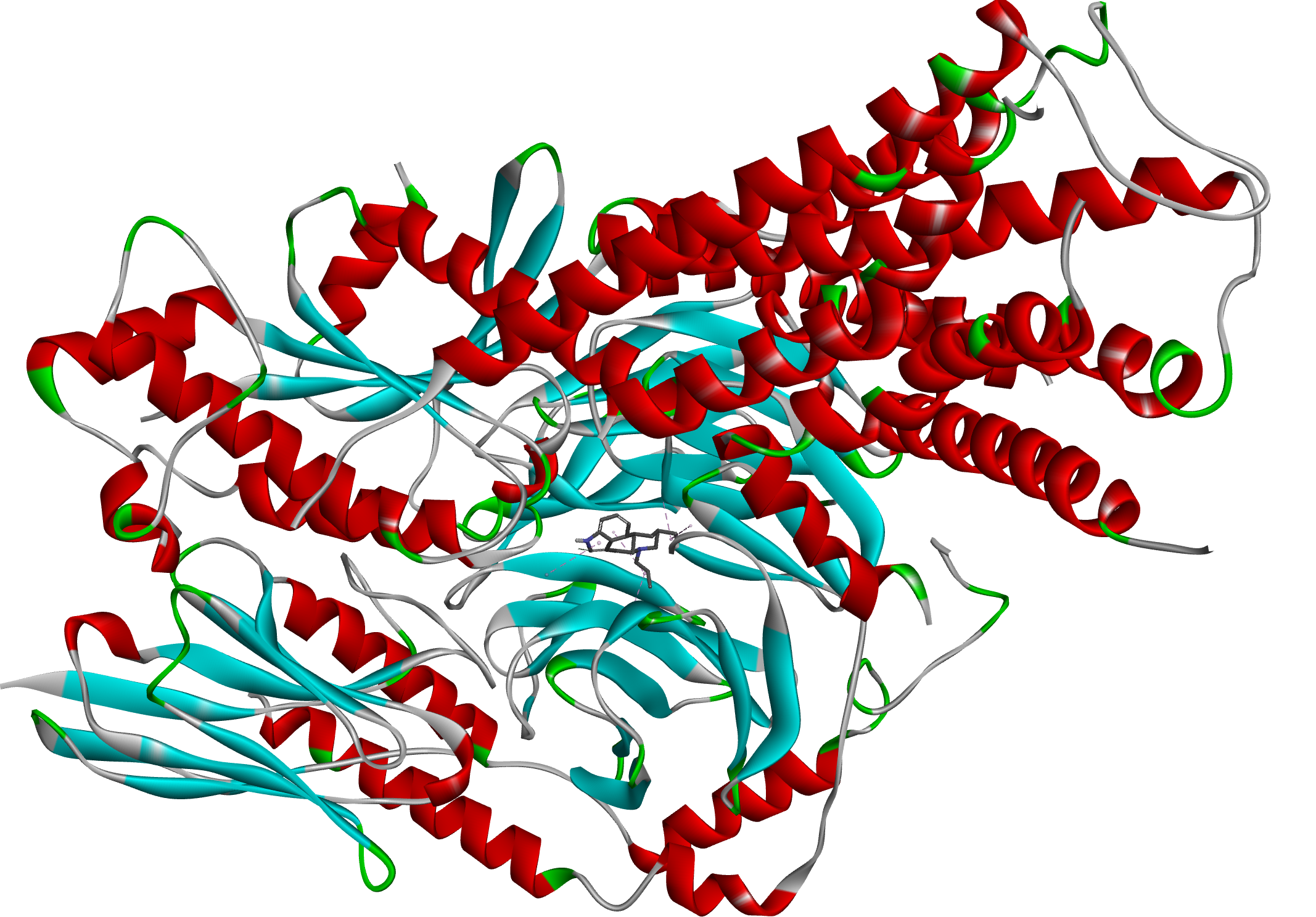}} 
\caption{3D visualization of docking results}
\label{fig:Docking_3d}
\end{figure*}

\clearpage
\bibliographystyle{unsrt}
%\bibliography{LB}
\bibliography{BL_Nov_25}